\pdfoutput=1
\documentclass[11pt]{article}
\usepackage[final]{acl}

\usepackage{times}

\usepackage{graphicx}
\usepackage{hyperref}
\usepackage{caption}
\usepackage{makecell}
\usepackage{subcaption}
\usepackage{colortbl}
\usepackage{latexsym}
\usepackage{url}
\usepackage{booktabs}
\usepackage{arydshln}
\usepackage{amsmath}
\usepackage{multirow}
\usepackage{CJKutf8}
\usepackage{array}
\usepackage{hhline}
\usepackage{enumitem} 
\usepackage[T1]{fontenc}
\usepackage{xcolor}
\usepackage[utf8]{inputenc}
\usepackage{amssymb}
\usepackage{marvosym}
\usepackage{fontawesome5}
\usepackage{pifont}

\usepackage{microtype}
\newcommand{\myCheckMark}{\textcolor{blue}{\ding{52}}}
\newcommand{\myCrossMark}{\textcolor{red}{\ding{55}}}

\definecolor{gemini_blue}{HTML}{4328e7}
\definecolor{internvl_purple}{HTML}{9654e5}
\definecolor{minicpm_pink}{HTML}{ff6283}
\definecolor{qwen_orange}{HTML}{ff8800}
\definecolor{videollama_yellow}{HTML}{ffc502}
\definecolor{gpt4_green}{HTML}{007d8e}
\definecolor{gt_yellow}{HTML}{F8E3B5}

\usepackage{inconsolata}

\usepackage{graphicx}

\title{VideoVista-CulturalLingo: 360° Horizons-Bridging Cultures, Languages, and Domains in Video Comprehension}

\author{
 \textbf{Xinyu Chen\textsuperscript{1}},
 \textbf{Yunxin Li\textsuperscript{1}},
 \textbf{Haoyuan Shi\textsuperscript{1}},
 \textbf{Baotian Hu\textsuperscript{1}\thanks{~~~Corresponding author.}},
\\
 \textbf{Wenhan Luo\textsuperscript{2}},
 \textbf{Yaowei Wang\textsuperscript{1}},
 \textbf{Min Zhang\textsuperscript{1}},
\\
 \textsuperscript{1}Harbin Institute of Technology, Shenzhen, China,
\\
 \textsuperscript{2}Hong Kong University of Science and Technology,
\\
\texttt{\{chenxinyuhitsz, liyunxin987\}}@163.com
\\
\texttt{\{hubaotian, zhangmin2021\}}@hit.edu.cn
}
\begin{document}
\maketitle
\begin{abstract}
% In recent years, large multimodal models (LMMs) have undoubtedly been a hot research topic in the field of artificial intelligence, and research in the multimodal video domain has also been developing rapidly in the past year.
Assessing the video comprehension capabilities of multimodal AI systems can effectively measure their understanding and reasoning abilities. Most video evaluation benchmarks are limited to a single language, typically English, and predominantly feature videos rooted in Western cultural contexts. In this paper, we present \textbf{VideoVista-CulturalLingo}, the first video evaluation benchmark designed to bridge cultural, linguistic, and domain divide in video comprehension. Our work differs from existing benchmarks in the following ways: 1)  \textbf{Cultural diversity}, incorporating cultures from China, North America, and Europe; 2) \textbf{Multi-linguistics}, with questions presented in Chinese and English—two of the most widely spoken languages; and 3) \textbf{Broad domain}, featuring videos sourced from hundreds of human-created domains. VideoVista-CulturalLingo contains 1,389 videos and 3,134 QA pairs, and we have evaluated 24 recent open-source or proprietary video large models. From the experiment results, we observe that: 
1) Existing models perform worse on Chinese-centric questions than Western-centric ones, particularly those related to Chinese history;
2) Current open-source models still exhibit limitations in temporal understanding, especially in the Event Localization task, achieving a maximum score of only 45.2\%;
3) Mainstream models demonstrate strong performance in general scientific questions, while open-source models demonstrate weak performance in mathematics. 

% like Calculus
\end{abstract}

%\footnote{VideoVista-CulturalLingo Repository:\url{https://www.github.com/}}
% 我们最后凝练三个有趣的发现，让审稿人买账。
% 2.10 号晚上，把数据集确定好了。补充评估。
% 文章对于数据集统计，对比；

% \textcolor{gt_yellow}{yellow}
% \colorbox{gt_yellow}{yellow.}
\section{Introduction}
\begin{figure}[!t]
    \centering
    \includegraphics[width=0.45\textwidth]{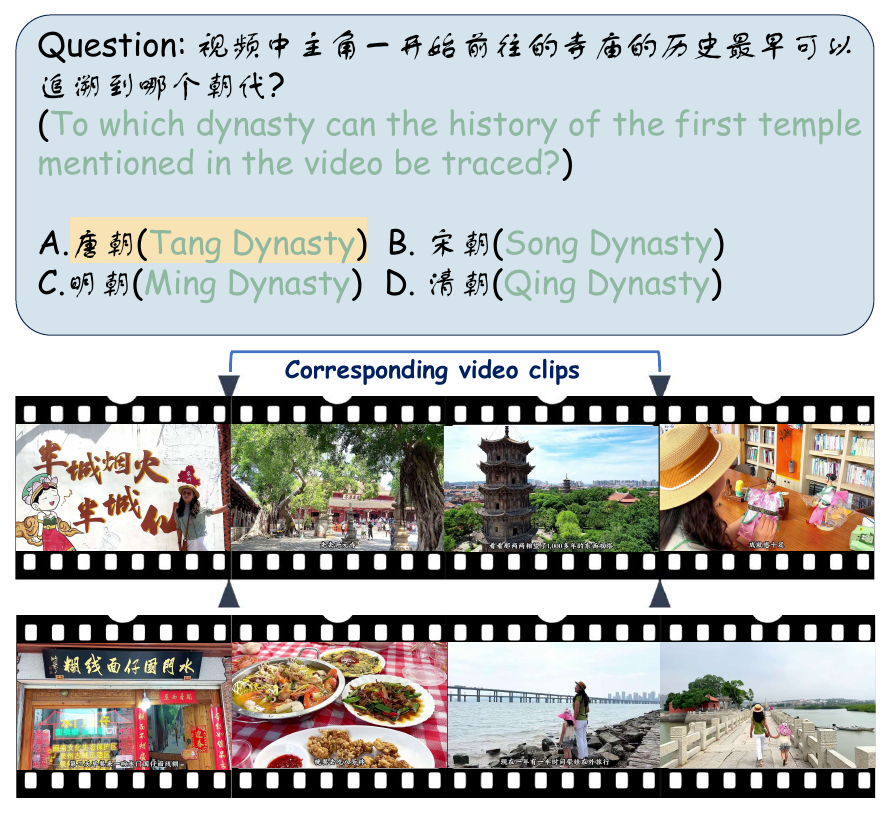}
    \caption{\textbf{An example of Chinese Culture in VideoVista-CulturalLingo.} The correct answer is highlighted in yellow.}
    \label{fig:example}
\end{figure}

\begin{figure*}[!t]
 \begin{minipage}{0.4\textwidth}
 %\centering
 \small
 \renewcommand\tabcolsep{1.7pt} % column space
 \renewcommand\arraystretch{1.10} % row space
  %\captionof{table}{Statistics and Values}
 \begin{tabular}{l r}

        \toprule
        \textbf{Category} & \textbf{Size} \\
        % \hline
        % Languages & EN \\
        \midrule
        Task Classes & 4 \\
        Subtask Classes & 14 \\
        \hline
        Video Sources & 1,389\\
        Video Clips & 2,052\\
        Max Duration & 1,877.7\\
        Average Duration & 267.5 \\
        \hline
        YouTube Video Domains & 30\\
        RedNote Video Domains & 104\\
        BiliBili Video Domains & 12\\
        \hline
        Chinese Question Number & 1,446\\
        English Question Number & 1,668\\
        \hdashline\noalign{\vskip 0.5ex}
        Chinese Culture QA Number & 231\\
        American Culture QA Number & 200\\
        European Culture QA Number & 200\\
        \midrule
        % Maximum Question Length & 67\\
        % Maximum Option Length & 108\\
        % Maximum Choice Number & 4\\
        % \hdashline\noalign{\vskip 0.5ex}
        Average Question Length &  18\\
        Average Option Length &  13\\
        Average Choice Number & 4\\
        \hdashline\noalign{\vskip 0.5ex}
        Total Samples & 3,134 \\
        Total Questions & 3,134 \\
        \bottomrule
    \end{tabular}
\end{minipage} 
\hfill
\begin{minipage}{0.60\textwidth}
 \small
 \vspace{-1mm}
\includegraphics[width=0.94\linewidth]{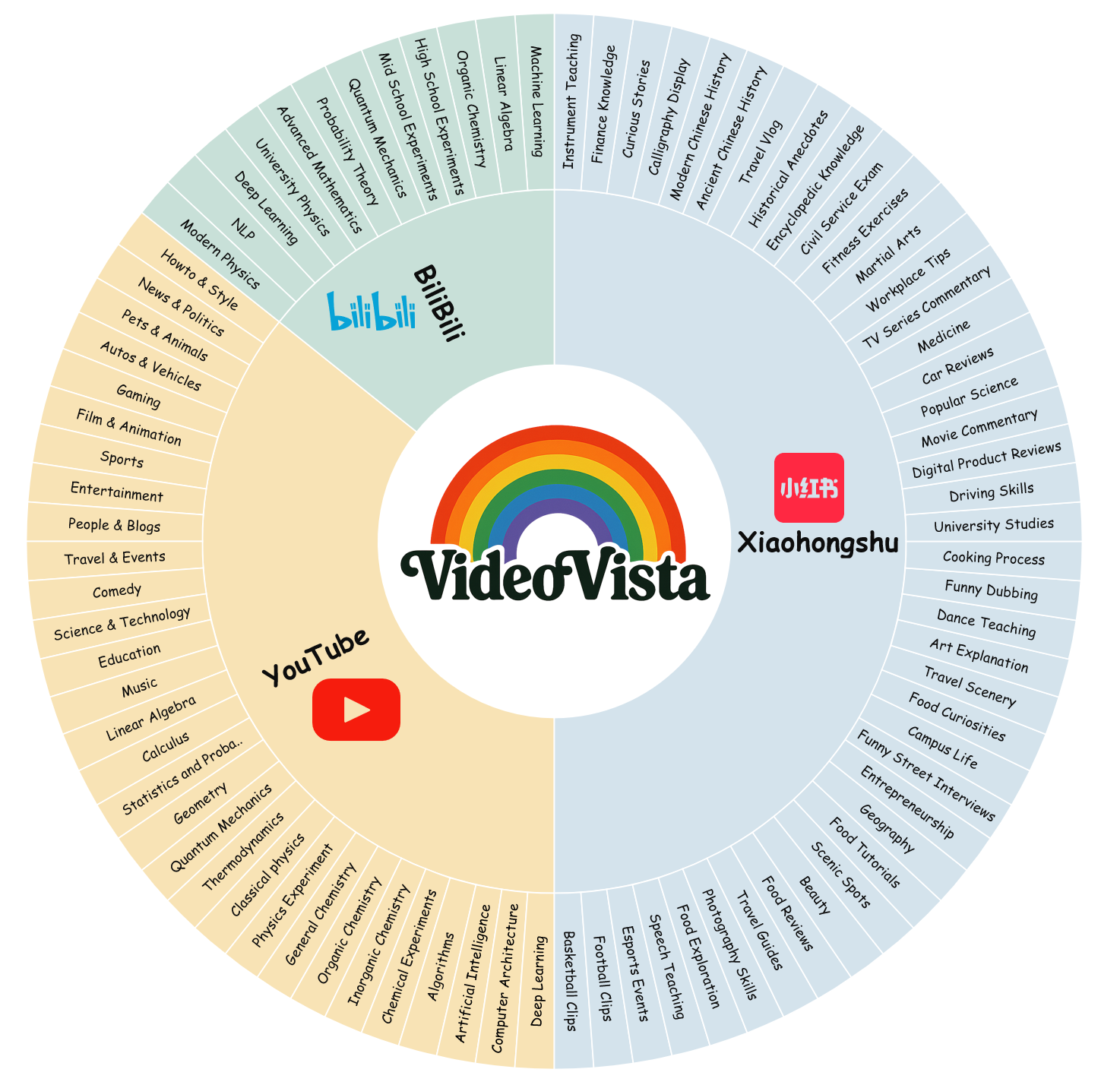}
\vspace{-1mm}
 \end{minipage}
\caption{\textbf{(Left)} Comprehensive statistics from different perspectives. The durations reported are based on the statistics from the 2,052 video clips. The question and answer length is count in tokens; \textbf{(Right)} Videos in VideoVista-CulturalLingo is sourced hundreds of domains from \textbf{3} popular video websites across the world. In the video sourced from Xiaohongshu (RedNote), we only present 42 of the all domains. }
\label{fig:comprehensive_stats}
\end{figure*}

Large Multimodal Models (LMMs) built upon Large Language Models (LLMs) have demonstrated unprecedented capabilities across various domains, including text, image, video, and audio over several years. Particularly in the past year, there has been a surge in the development of LMMs capable of processing video inputs. The dramatic expansion in the length of video frame sequences—from just a few frames to several hundred—demonstrates significant progress in video understanding capabilities. Meanwhile, video evaluation benchmarks have also emerged, evolving from early-stage basic video question answering tasks~\cite{yu2019activityqa,xu2017video} to general video evaluation benchmarks~\cite{fu2024video,MLVU,wang2024lvbench}. However, existing video evaluation benchmarks predominantly select videos from sources such as YouTube, Shutterstock, or established video datasets like Ego4D~\cite{Grauman_2022_CVPR} and Movie101~\cite{yue-etal-2023-movie101}. These datasets are primarily Western-centric, with a limited representation of Chinese-centric videos as shown in Figure~\ref{fig:example}. In addition, current video evaluation benchmarks tend to focus on specific events within the videos, neglecting the cultural context and connotations embedded in the content while overlooking the scientific principles and information that the videos are intended to convey.

To advance the development of LMMs, we introduce VideoVista-CulturalLingo, the first video evaluation benchmark designed to bridge cultures, languages, and domains in video comprehension. In Figure~\ref{fig:comprehensive_stats}, we present detailed statistics on the questions and videos in VideoVista-CulturalLingo. It comprises 3,134 questions organized into 14 tasks, spanning 2,052 video clips of varying lengths and reflecting both Western and Chinese cultures. English-language videos are sourced from YouTube, while Chinese videos are collected from Xiaohongshu and BiliBili. These videos cover hundreds of distinct domains, ranging from everyday life topics—such as news reports, travel recommendations, sports events, and vlogs—to scientific topics, including calculus, deep learning, organic chemistry, and quantum mechanics. 
% Video durations vary widely, from a few seconds to tens of minutes.

To efficiently annotate such a large-scale video dataset, we employ a hybrid annotation framework that combines the strengths of both (M)LLMs and human efforts. This framework leverages the powerful capabilities of existing large models, such as Qwen2-VL~\cite{Qwen2VL} and DeepSeek-R1~\cite{deepseekai2025deepseekr1incentivizingreasoningcapability}, to generate an initial pool of question-options-answer (QA) pairs. Human annotators then select the high-quality questions from generated QA pairs and further refine them to enhance clarity and quality.

% A comparison between VideoVista-CulturalLingo and existing video benchmarks is presented in Table~\ref{tab:comparison}.
% We believe this approach can provide valuable insights for the construction of future benchmarks.

We have evaluated 24 state-of-the-art (SOTA) LMMs, including proprietary LMMs such as GPT-4o, Gemini-2.0-Flash, as well as open-source video LMMs like Qwen2.5-VL~\cite{qwen2.5-VL} and VideoLLaMA3~\cite{damonlpsg2025videollama3}, and image LMMs such as Molmo~\cite{molmo2024} and DeepSeek2-VL~\cite{wu2024deepseekvl2mixtureofexpertsvisionlanguagemodels}. Experimental results show that Gemini-2.0-Flash demonstrates the strongest performance among all models, achieving an accuracy score of \textbf{76.3\%}. Among open-source video LMMs, Qwen2.5-VL-72B achieves the highest score of 61.3\%, with a large performance gap compared to Gemini-2.0-Flash in video location tasks. Interestingly, Qwen2.5-VL performs best on cultural understanding, yet still achieves only 65.8\% in Chinese cultural understanding. 
In summary, the main contributions are as follows:
\begin{itemize}[leftmargin=*,topsep=0.1em,itemsep=0.1em,parsep=0.1em]
    \item We present the first video evaluation benchmark that covers diverse domains, languages, and cultures in video comprehension.
    \item We introduce an autonomic video annotation framework, harnessing the strengths of (M)LLMs (including Qwen2-VL and DeepSeek-R1) and visual recognition tools (including SAM2) to improve the efficiency of video annotation.
    \item We conduct extensive experiments and in-depth analysis with VideoVista-CulturalLingo, revealing the limitations of existing LMMs in videos with different cultural or linguistic contexts.
\end{itemize}
% Experimental results reveal that: 1) 
% 1)the performance gap between open-source LMMs and proprietary LMMs has significantly narrowed compared to one year ago. 
% 2)Despite a increase in the number of frames sampled, the temporal perception capabilities of existing large multimodal models still exhibit notable limitations. 
% 3) Compared to videos in the Chinese-centric category, most LMMs perform better on videos in the Western-centric category.

% \begin{figure*}[t]
%     \centering
%     \includegraphics[width=1.0\textwidth]{figures/Example.pdf}
%     \caption{\textbf{Examples in VideoVista-CulturalLingo.} The ground-truth answer is highlighted in yellow. }
%     \label{fig:example}
% \end{figure*}

\begin{figure*}[!t]
    \centering
    \includegraphics[width=0.85\textwidth]{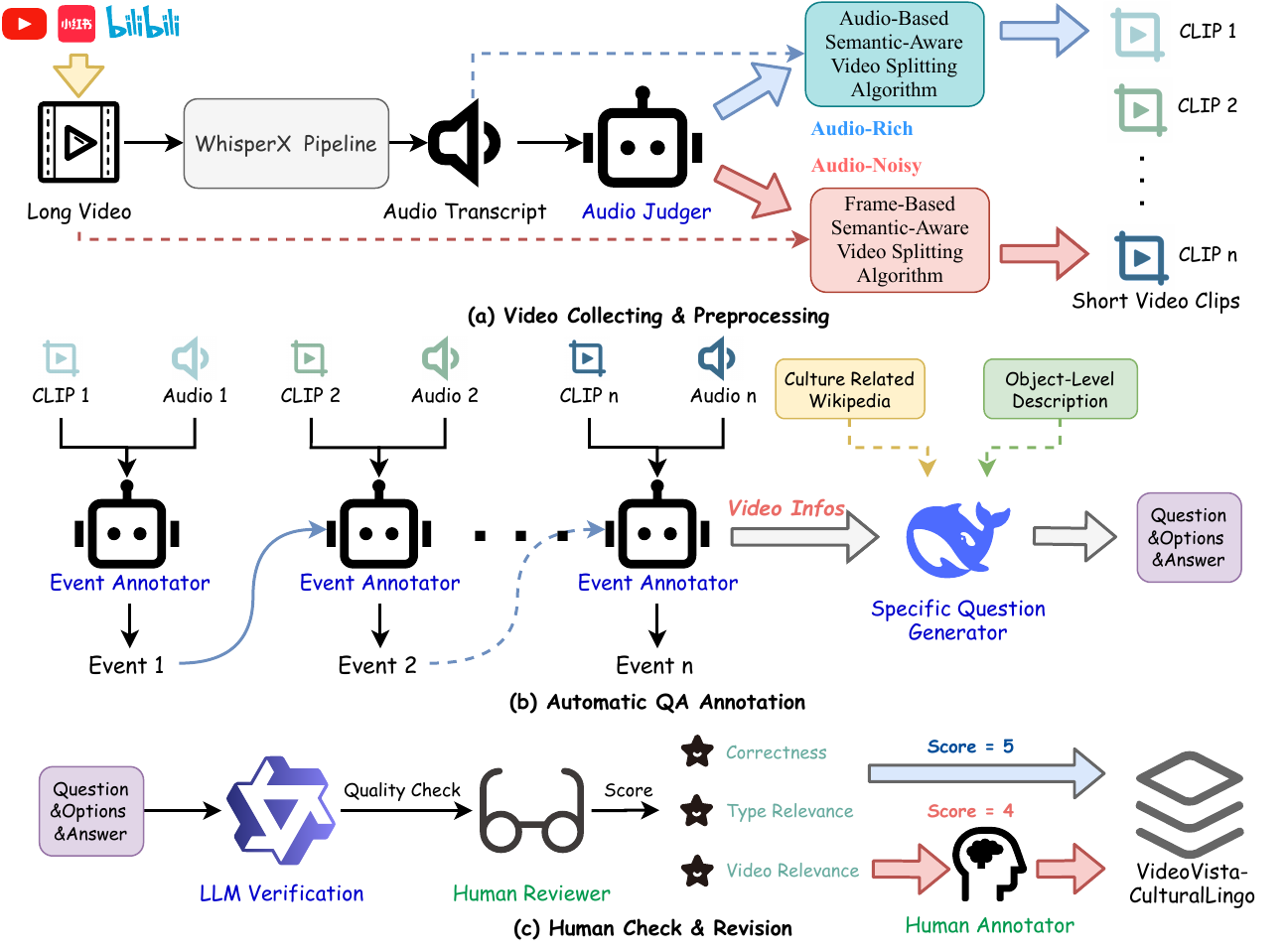}
    \caption{\textbf{The three-stage annotation process of VideoVista-CulturalLingo.} }
    \label{fig:method}
\end{figure*}

% In (b), Video Infos is represented as a list consisting of multiple video clips in the form of $(event_i, audio_i, start_i, end_i)$.

\section{Related Work}

\paragraph{Development of Video LMMs.}
 Unified encoding methods for both image and video modalities have become the mainstream approach adopted by LMMs over the past year. LongVA~\cite{zhang2024longva} utilizes a unified encoding method, Uni-Res, which allows models trained solely on image datasets to demonstrate strong potential in video evaluation tasks. Qwen2-VL~\cite{Qwen2VL} and Qwen2.5VL~\cite{qwen2.5-VL} introduce the M-RoPE positional encoding, incorporating temporal, height, and width components, enabling unified positional modeling across text, image, and video modalities. LLaVA-Video~\cite{zhang2024videoinstructiontuningsynthetic} draws inspiration from the SlowFast approach, encoding video frames at varying granularities into visual sequences of different lengths, effectively addressing the issue of excessively long sequences during video encoding. Current LMMs~\cite{chen2024expanding,yao2024minicpm,li2024llava,li2024uni,lmeye} are capable of unified encoding for image and video modalities, leveraging rich image modality data to enhance visual capabilities and demonstrate strong performance in video evaluation tasks.
 
\paragraph{Progress of Video Benchmark.}
Video evaluation benchmarks have also made significant progress. Previously, evaluation datasets~\cite{yu2019activityqa,xu2017video} typically involved posing broad questions and having the model generate a one or a few-word answer, which was then assessed for accuracy and scored by LLMs~\cite{Maaz2023VideoChatGPT}. The videos used in these datasets were often limited to just a few dozen seconds or minutes in length. Recent video benchmarks~\cite{li2024videovista} have seen considerable improvements, both in the variety of evaluation tasks and the duration of the videos. Video-MME~\cite{fu2024video} has extended the evaluation video length to an hour, while also introducing twelve distinct evaluation tasks, including Temporal Reasoning and Information Synopsis. MLVU~\cite{MLVU} includes videos of varying lengths, ranging from 3 minutes to 2 hours, covering nine different evaluation tasks, such as Needle Question-Answering. The process of video benchmarks~\cite{fang2024mmbenchvideo,wang2024lvbench,liu2024tempcompass} have undoubtedly provided a significant boost to the development of LMMs.

\section{VideoVista-CulturalLingo}

% Then the scientific video is collecting used YouTube Data API
\subsection{Video Collecting and Preprocessing}
The videos in our study can be divided into two categories: non-scientific and scientific videos. Non-scientific English videos are randomly crawled from YouTube, while their Chinese counterparts are collected from Xiaohongshu to ensure diversity within the dataset. The domains of these videos come from the original categories on the video platforms.
For scientific videos, we first identified four major disciplines: mathematics, physics, chemistry, and computer science. Within each discipline, we further defined four representative sub-disciplines, such as linear algebra in mathematics and quantum mechanics in physics. Domains of these videos are derived from search keywords.
These sub-disciplines guide the collection of English scientific videos via the YouTube Data API.  For Chinese scientific videos, human annotators manually collected videos from BiliBili.

All videos undergo audio extraction via FFmpeg, followed by transcription using Whisper-Large-v3 with sentence-level timestamp alignment. An audio quality assessment pipeline is implemented using Qwen2.5-32B~\cite{qwen2.5}, evaluating three dimensions: logical coherence, continuity, and information density. Videos are subsequently classified as either audio-rich (high-quality speech) or audio-noisy (including silent videos).
For audio-rich videos, the Qwen2.5-72B model segments transcriptions into contextually coherent paragraphs, which are synchronized with visual content through Whisper's sentence-level alignment to generate short video clips. Audio-noisy videos are processed using the semantics-aware video splitting algorithm from Panda-70M~\cite{chen2024panda70m}, which utilizes visual features to partition videos into semantically consistent segments. This process is illustrated in Figure~\ref{fig:method} (a).

To address the challenges of Chinese homophone ambiguity in transcriptions, we develop a context-aware refinement module using Qwen2.5-72B. This module performs three key operations: (1) disambiguation of homophones through semantic analysis, (2) correction of domain-specific terminology, and (3) fluency enhancement, while strictly preserving original semantic content. 

% \footnote{\url{http://www.youtube.com}}
% \footnote{\url{https://www.xiaohongshu.com}}
% \footnote{\url{https://www.bilibili.com}}

% For scientific videos, we first identified four major disciplines: mathematics, physics, chemistry, and computer science. Within each discipline, we further defined four representative sub-disciplines, such as linear algebra in mathematics, quantum mechanics in physics

% For scientific videos, we establish a hierarchical selection framework: four core disciplines (mathematics, physics, chemistry, computer science) are first identified, each containing four representative sub-disciplines (e.g., linear algebra, quantum mechanics, organic chemistry, machine learning).

\begin{table*}[!t]
\centering
\scriptsize
\scalebox{0.9}{
  \begin{tabular}{l r r r c c c c c c}
    \toprule
    \textbf{Benchmarks} & \textbf{\#Videos} & \textbf{\#Clips} & \textbf{Len.(s)} & \textbf{\#QA Pairs} & \textbf{Anno.} & \textbf{M.L.} & \textbf{M.C} & \textbf{M.D} & \textbf{Open.} \\
    \midrule
    MSRVTT-QA~\cite{xu2017video} & 2,990 & 2,990 & 15.2 & 72,821 & A & \myCrossMark & \myCrossMark & \myCrossMark & \myCheckMark \\
    MSVD-QA~\cite{xu2017video} & 504 & 504 & 9.8 & 13,157 & A & \myCrossMark &\myCrossMark & \myCrossMark & \myCheckMark \\
    TGIF-QA~\cite{li2016tgif} & 9,575 & 9,575 & 3.0 & 8,506 & A\&M & \myCrossMark & \myCrossMark& \myCrossMark & \myCheckMark  \\
    ActivityNet-QA~\cite{yu2019activityqa} & 800 & 800 & 111.4 & 8,000 & M & \myCrossMark &\myCrossMark & \myCrossMark & \myCrossMark \\
    TVQA~\cite{lei2018tvqa} & 2,179 & 15,253 & 11.2 & 15,253 & M & \myCrossMark & \myCrossMark& \myCrossMark & \myCrossMark  \\
    NExT-QA~\cite{xiao2021next} & 1,000 & 1,000 & 39.5 & 8,564 & A & \myCrossMark & \myCrossMark & \myCrossMark & \myCheckMark \\
    \midrule
    MVBench~\cite{li2023mvbench} & 3,641 & 3,641 & 16.0 & 4,000 & A & \myCrossMark & \myCrossMark & \myCrossMark & \myCheckMark \\
    EgoSchema~\cite{mangalam2024egoschema} & 5,063 & 5,063 & 180.0 & 5,063 & A\&M & \myCrossMark & \myCrossMark & \myCrossMark & \myCrossMark \\
    TempCompass~\cite{liu2024tempcompass} & 410 & 500 & 11.4 & 7,540 & A\&M & \myCrossMark & \myCrossMark & \myCrossMark & \myCheckMark  \\
    Video-MME~\cite{fu2024video} & 900 & 900 & 1024.0 & 2,700 & M  & \myCrossMark & \myCrossMark & \myCheckMark & \myCheckMark \\
    VideoVista~\cite{li2024videovista} & 894 & 3,402 & 131.0 & 24,906 & A & \myCrossMark & \myCrossMark & \myCheckMark & \myCheckMark \\
    MLVU~\cite{MLVU} & 1,323 & 1,323 & 720 & 2,593 & A\&M & \myCrossMark & \myCrossMark &  \myCheckMark  & \myCheckMark \\
    LVBench~\cite{wang2024lvbench} & 500 & 500 & 4,101.0 & 1,549 & M & \myCrossMark & \myCrossMark & \myCheckMark & \myCheckMark \\
    MMBench-Video~\cite{fang2024mmbenchvideo} & 600 & 600 & 165.4 & 1,998 & M & \myCrossMark & \myCrossMark & \myCheckMark & \myCheckMark \\
    \midrule
    \textbf{VideoVista-CulturalLingo} & 1,389 & 2,052 & 267.5 & 3,134 & A\&M & \myCheckMark & \myCheckMark &\myCheckMark & \myCheckMark \\
    \bottomrule
  \end{tabular}}
  \caption{The comparison of various benchmarks involves several key aspects: total number of videos (\textbf{\#Videos}), number of clips (\textbf{\#Clips}), average video duration (\textbf{Len.}), number of QA pairs (\textbf{\#QA Pairs}), annotation method (\textbf{Anno.}, where M/A indicates manual/automatic annotation), whether the videos span multiple language (\textbf{M.L.}),whether the videos span multiple culture background (\textbf{M.C.}) ,whether the videos span multiple duration levels (\textbf{M.D.}), and if the videos are sourced from diverse open domains (\textbf{Open.})}
  \label{tab:comparison}
\end{table*}

\subsection{Automatic QA Annotation}
The annotation framework comprises four distinct tasks: Event, Culture, Object, and Science. Our pipeline employs Qwen2-VL-72B as the primary annotator, Qwen2.5-72B for text-only annotation tasks, and paraphrase-multilingual-MiniLM-L12-v2 for embedding generation. For non-scientific tasks, DeepSeek-V3~\cite{deepseekai2024deepseekv3technicalreport} is employed as the question generator, while DeepSeek-R1~\cite{deepseekai2025deepseekr1incentivizingreasoningcapability} is used for generating scientific questions. During the annotation process, while generating questions, four options and the correct answer are also created. The process of automatic QA annotation is illustrated in Figure~\ref{fig:method} (b). \textit{The details and prompt for annotation is provided in Appendix~\ref{sec:detail_annotations}.}

\paragraph{Event.}
% We input the segmented video clips and refined audio transcriptions into the Event Annotator  to label the events occurring in each video segment. During the input process, the segments are strictly fed into the model in sequence for a complete video. When inputting the i-th sub-segment, the events generated for the previous i-1 segments are also included as input to the model, aiming to enhance the consistency of the generated events. After annotating the events in each video segment, we have the following structure for each segment $(event, audio,start,end)$, where 'start' and 'end' represent the start and end times of the current video segment within the full video. The complete video is represented as a list of such entries. This list is then input into the Question Generator to guide it in generating one or two questions of the corresponding type, along with four options for each question. Specifically, for event prediction questions, we instruct the model to select the segment from the video sequence that is logically most closely related to the preceding context as the predicted content, and we also record the index of the selected segment. In this process, each question type is associated with a specific prompt. 

We input the segmented video clips and refined audio transcriptions into the event annotator to label the events occurring in each video segment. For the $i$-th segment, the model receives historical event annotations from the previous $i-1$ segments to maintain temporal consistency. Each annotated segment follows the structure $(event, audio,start,end)$, where $start$ and $end$ denote the timestamps marking the beginning and conclusion of the current video segment within the full video. The aggregated event sequence is then fed into the question generator, which generates questions of the corresponding task, along with four options for each question and correct answer. Specifically, for event prediction questions, the model is instructed to select the segment that is most logically related to the preceding context as the predicted content. During this process, each task is associated with a specific prompt.

\paragraph{Object.}
% 

% Before performing object-level annotation, we use an Object Selector to filter the videos based on three criteria: whether the video is a Real-World Video, the richness of objects, and the dynamics of the objects.
% The filtered videos are then input into the Object Extractor to identify three to five primary objects. Next, we extract frames from the video at one frame per second, using the InternVL2-8B model to determine the presence of each object in every frame. For objects that are present, we input both the object and the corresponding video frames into a pipeline consisting of Grounding-DINO~\cite{liu2023grounding} and SAM2~\cite{ravi2024sam2segmentimages} to obtain the bounding boxes and image segmentation for each object. Above information is then fed into the Object Description Annotator to generate object-level descriptions, capturing both the temporal and spatial dimensions of each object. Finally, the object-level descriptions, along with the event data, are input into the Question Generator to generate the corresponding questions.

We feed videos into the object classifier to filter those videos that meet three criteria: real-world content, richness in objects, and motion in objects. The filtered videos are then processed by the object extractor to identify three to five primary objects 
followed by frame-wise presence detection via InternVL2-8B at 1fps sampling. The detected objects are processed through a pipeline combining Grounding-DINO~\cite{liu2023grounding} for bounding box prediction and SAM2~\cite{ravi2024sam2segmentimages} for image segmentation. The resulting information is then fed into the object description annotator to generate object-level descriptions that capture both the temporal and spatial aspects of each object. Finally, the object-level descriptions, along with the aggregated event sequence, are input into the question generator to generate the questions.

\paragraph{Culture.}
% video is input to Cultural Classifier to evaluates the video's relationship to Chinese, North American, and European cultures based on both visual and audio information. If the video is associated with any of these cultures, we use a Cultural Concept Extractor to identify the two most prominent cultural concepts from the video. Subsequently, we encode these cultural concepts into embeddings using an Embedder, and recall the most similar entries from pre-encoded Wikipedia entries. Using these entries, along with a local backup of Wikipedia, we can retrieve article corresponding to the identified cultural concepts. By combining this external knowledge with pre-annotated event data, we input it into a Question Generator to generate the final questions. 

We input videos and audio transcriptions into the cultural classifier to evaluate their relationship to Chinese, American, and European cultures individually. Culturally relevant videos are then processed by the cultural concept extractor to identify the two most prominent cultural concepts. These cultural concepts are subsequently encoded into embeddings, which are used to retrieve the entries from pre-encoded Wikipedia data. Using these entries, along with a local backup of Wikipedia, we can retrieve Wikipedia articles corresponding to the identified cultural concepts. By combining this external knowledge with the aggregated event sequence, we input the data into the specific question generator to generate the questions. 

\paragraph{Science.}
% We first input the audio information of scientific videos into a Science Classifier, which evaluates the video quality based on two dimensions: thematic relevance and knowledge density, thereby filtering out videos of lower quality. The filtered videos, along with their audio and event information, are then input into the Question Generator to generate questions for different sub-categories. Unlike the questions in the previous three categories, the options for scientific questions need to be written according to a strict set of rules: Correct Option, Video Comprehension Error Option, Domain Knowledge Error Option, and Dual Error Option. These options are designed to assess the model’s ability to understand the video and perform scientific reasoning. To ensure that the model generates the options in accordance with these rules and provides justifications for the selected options, we specifically employ the powerful DeepSeek-R1~\cite{deepseekai2025deepseekr1incentivizingreasoningcapability} model as the Question Generator.

The video is input into the science classifier to evaluate its quality based on scientific thematic relevance and knowledge density. After filtering, the aggregated event sequence of the video is fed into the question generator, DeepSeek-R1, to generate questions.
In our initial experiments, we observed two recurring issues: generated questions either relied excessively on domain knowledge—detached from the video itself and thus answerable without viewing—or exhibited distractor choices that were either too divergent or too similar, producing items that were trivial or ambiguous. To resolve these shortcomings, we impose deterministic, rule‐based constraints that (i) require every question to depend on video context for its solution and (ii) ensure a balanced, pedagogically meaningful set of answer options. Specifically, each question presents four choices: Correct Option, Video Comprehension Error, Domain Knowledge Error, and Dual Error.
This structured design rigorously evaluates a model’s ability to integrate visual comprehension with scientific reasoning.
% Unlike the questions generation in the previous three tasks, the options for scientific questions must adhere to a strict set of rules: Correct Option, Video Comprehension Error Option, Domain Knowledge Error Option, and Dual Error Option. These options are designed to assess the model’s ability to comprehend the video content and perform scientific reasoning.
% To ensure that the model generates the options in accordance with these rules and provides justifications for the selected options, we specifically employ the DeepSeek-R1 model as the Question Generator.

\subsection{Human Check and Revision}

All candidate questions are first filtered linguistically using Qwen2.5‑7B with the CircularEval strategy~\cite{MMBench} to remove any video‑agnostic items. We then establish a Gradio-based annotation platform that includes three assessment dimensions: correctness, type relevance, and video relevance. The correctness score ranges from 0 to 1, assessing whether the model-generated answer is correct; the type relevance score ranges from 0 to 2, evaluating the degree of relevance between the question and task type; and the video relevance score ranges from 0 to 2, determining the degree of relevance between the question and the video content, ensuring that questions are not unrelated to the video frames. Questions achieving maximum scores (score=5) across all dimensions are  selected. For borderline cases (score=4), we utilize differentiated handling: first, for the question with wrong answer (correctness=0), we manually correct the answers; second, for the question with suboptimal type or video relevance, we manually refine the questions, options, and answers based on the original questions. We have illustrated this process in the Figure~\ref{fig:method} (c).
Specifically for cultural questions, two annotators—one of whom is a native speaker of the relevant culture—independently assess each question to ensure cross‑validation. Overall, this hybrid automatic/manual pipeline eliminates approximately 60 \% of low‑quality questions.
% \textit{The manual checking website is presented in the Appendix~\ref{webpage annoation}}.
% % 修改
% Specifically for culturally related questions, we applied a cross-validation approach where each model-generated question was independently evaluated by two annotators—ensuring that at least one annotator was a native speaker of the relevant culture.
% This pipeline eliminates 60\% of low-quality questions through combined automatic and manual filtering.

% 

% All candidate questions are first filtered linguistically using Qwen2.5‑7B with the CircularEval strategy~\cite{MMBench} to remove any video‑agnostic items. We then deploy a Gradio‑based annotation platform to evaluate each question along three dimensions: correctness, type relevance, and video relevance. The correctness score ranges from 0 to 1, assessing whether the model-generated answer is correct; the type relevance score ranges from 0 to 2, evaluating the degree of relevance between the question and task type; and the video relevance score ranges from 0 to 2, determining the degree of relevance between the question and the video content, ensuring that questions are not unrelated to the video frames.

\begin{table*}[!t]
\renewcommand\arraystretch{1.3}
\centering
\scriptsize
  \scalebox{1.}{
  \begin{tabular}{l | l | c | c | c | c | c | c}
    \hline
    \textbf{Model} & \textbf{LLM} & \textbf{Frames} & \textbf{Overall} & \textbf{Event} & \textbf{Object} & \textbf{Culture} & \textbf{Science} \\
    \hline
    \rowcolor[HTML]{A9D0D5} \multicolumn{8}{c}{\textit{Open-source Video LMMs}} \\
    \hline
    ShareGPT4Video~\cite{chen2024sharegpt4video} & Vicuna-7B-v1.5 & 16f & 25.6 & 23.2 & 18.9 & 31.4 & 34.1 \\
    VideoChat2-Mistral~\cite{2023videochat} & Mistral-7B-Instruct-v0.2 & 16f & 29.6 & 27.5 & 25.9 & 34.7 & 33.1 \\
    Video-LLaVA~\cite{lin2023video} & Vicuna-7B-v1.5 & 8f & 38.2 & 42.2 & 34.4 & 34.5 & 41.1 \\
    VideoLLaMA2~\cite{damonlpsg2024videollama2} & Mistral-7B-Instruct-v0.2 & 32f & 31.4 & 33.6 & 23.3 & 34.9 & 36.6 \\
    LLaVA-OneVision~\cite{li2024llava} & Qwen2-7B-Instruct & 32f & 41.8 & 43.9 & 33.8 & 38.8 & 53.5 \\
    MiniCPM-V 2.6~\cite{yao2024minicpm} & Qwen2-7B-Instruct & 1fps(64) & 42.9 & 44.1 & 24.1 & 49.4 & 62.9 \\
    mPLUG-Owl3~\cite{ye2024mplugowl3longimagesequenceunderstanding} & Qwen2-7B-Instruct & 1fps(128) & 49.9 & 54.4 & 41.9 & 45.0 & 60.1 \\
    Oryx-1.5~\cite{liu2024oryx} & Qwen2.5-7B-Instruct & 128f & 41.4 & 43.8 & 32.2 & 37.6 & 55.8 \\
    LLaVA-Video~\cite{zhang2024videoinstructiontuningsynthetic} & Qwen2-7B-Instruct & 1fps(64) & 51.0 & 57.9 & 39.1 & 48.8 & 60.3 \\
    Qwen2-VL~\cite{Qwen2VL} & Qwen2-7B-Instruct & 1fps(300) & 49.7 & 50.1 & 33.8 & 54.8 & 68.0 \\
    InternVL2.5~\cite{chen2024expanding} & Internlm2.5-7b-Chat & 64f & 52.0 & 56.5 & 35.5 & 56.1 & 65.7 \\
    MiniCPM-o 2.6~\cite{yao2024minicpm} &  Qwen2.5-7B-Instruct & 1fps(64) & 49.0 & 52.9 & 28.5 & 55.9 & 67.1\\
    TPO~\cite{li2025temporalpreferenceoptimizationlongform} & Qwen2-7B-Instruct & 1fps(96) & 50.6 & 57.2 & 37.8 & 49.6 & 60.4 \\
    InternVideo2.5~\cite{wang2025internvideo} & Internlm2.5-7b-Chat & 1fps(512) & 52.0 & 52.5 & 38.1 & \underline{58.2} & 65.9 \\
    VideoLLaMA3 ~\cite{damonlpsg2025videollama3} & Qwen2.5-7B-Instruct & 1fps(180) & \underline{60.7} & \underline{58.0} & \underline{\textbf{66.4}} & 53.1 & 64.4  \\
    Qwen2.5-VL-7B~\cite{qwen2.5-VL} & Qwen2.5-7B-Instruct & 1fps(300) & 54.3 & 56.7 & 38.9 &  55.2 & \underline{73.3} \\
    Qwen2.5-VL-72B~\cite{qwen2.5-VL} & Qwen2.5-72B-Instruct & 1fps(300) & \textbf{61.3} & \textbf{61.0} & 40.5  & \textbf{71.2}  & \textbf{83.3}  \\
    \hline
    \rowcolor[HTML]{D4E3EC} \multicolumn{8}{c}{\textit{Open-source Image LMMs}} \\
    \hline
    VILA1.5-13B~\cite{lin2023vila} &  Vicuna-13B-v1.5 & 1f & 33.3 & 33.3 & 29.2 & 33.9 & 39.2\\
    VILA1.5-13B~\cite{lin2023vila} &  Vicuna-13B-v1.5 & 8f & 36.9 & 38.2 & 31.3 & 38.2 & 41.9\\
    Molmo 7B-D~\cite{molmo2024} & Qwen2-7B-Instruct & 1f & 38.3 & 44.5 & 25.3 & 39.8 & 46.5 \\
    Molmo 7B-D~\cite{molmo2024} & Qwen2-7B-Instruct & 8f & 40.3 & 44.3 & 30.1 & 41.8 & 48.0 \\
    DeepSeek2-VL~\cite{wu2024deepseekvl2mixtureofexpertsvisionlanguagemodels} & DeepSeekMoE-27B & 1f & 40.9 & 44.3 & \textbf{32.2} & 39.3 & 50.5\\
    DeepSeek2-VL~\cite{wu2024deepseekvl2mixtureofexpertsvisionlanguagemodels} & DeepSeekMoE-27B & 8f & \textbf{42.6} & \textbf{47.0} & 27.2 & \textbf{44.4} & \textbf{57.5}  \\
    \hline
    \rowcolor[HTML]{8FB8A1} \multicolumn{8}{c}{\textit{Proprietary LMMs}} \\
    \hline
    GPT-4o-2024-11-20 & GPT-4o & 1fps(128) & 56.7 & 53.4 & 38.2 & \textbf{68.0} & 78.3 \\
    Gemini-1.5-Flash & Gemini-1.5-Flash & 1fps & 69.4 & 70.0 & 65.8 & 59.0 & 84.7 \\
    Gemini-2.0-Flash-Lite & Gemini-2.0-Flash-Lite & 1fps & 70.7 & 63.1 & 71.6 & 63.1 & 82.1 \\
    Gemini-2.0-Flash & Gemini-2.0-Flash & 1fps & \textbf{76.3} & \textbf{74.0} & \textbf{77.1} & \textbf{68.0} & \textbf{87.4} \\
    \hline
  \end{tabular}}
  \caption{\textbf{Evaluation results on VideoVista-CulturalLingo benchmark.} The large language model used by LMMs (\textbf{LLM}), frames sample strategy (\textbf{Frames}), overall evaluation scores (\textbf{Overall}), evaluation scores in Event Task(\textbf{Event}), evaluation scores in Object Task (\textbf{Object}), evaluation scores in Culture Task (\textbf{Culture}), evaluation scores in Science Task (\textbf{Science}). 
  -[$N$f] indicates this LMM task $N$ frames uniformly sampled from a video as input. -[$N$fps($M$)] indicates this LMM uses $N$ frames per second uniformly sampled from a video as input, with a max frames number $M$. 
  We have highlighted the highest results in each tasks using \textbf{bold}. Meanwhile, the highest results within the 7B/8B open-source Video LMMs are highlighted with an \underline{underline}.}
  \label{tab:main_result} 
\end{table*}

\subsection{Statistic and Analysis}
% 对数据集的统计分析和比较
% 可以把table 1 放在这一页，来结合说明
% 对比之前数据集，体现我们的特点。
% 我们数据集详细统计，多语言、多文化、科学video评估数据集。
As shown in Figure~\ref{fig:comprehensive_stats}, VideoVista-CulturalLingo consists of 2,052 video clips or full videos derived from 1,389 original videos, with an average duration of 267.5 seconds. Additionally, VideoVista-CulturalLingo contains 1,446 questions in Chinese and 1,668 questions in English, with a comparable number of questions in both languages.
In Table~\ref{tab:comparison}, we compare the key characteristics of our benchmark with others. Notably, VideoVista-CulturalLingo includes the largest collection of raw videos, totaling 1,389, among benchmarks that have videos multiple duration levels. These 1,389 original videos encompass a diverse range of languages and cultural backgrounds, a feature that sets our benchmark apart from previous ones. Additionally, we employed a annotation approach that combines (M)LLM preliminary annotation with human verification and question revision.

% 文化
% 科学挑战性： 可以说一下为什么高，我们采用的其实简单一些科学常识。复杂的我们将会在下一版本去做。

\section{Experiment}
\begin{table*}[t]
  \renewcommand\arraystretch{1.3}
  \centering
  \scriptsize
    \scalebox{1.}{
    \begin{tabular}{l | c c c c| c c c| c c c|c c c c}
      \hline
      \multirow{2}{*}{\textbf{Model}} & \multicolumn{4}{c|}{\textbf{Event}} & \multicolumn{3}{c|}{\textbf{Object}} & \multicolumn{3}{c|}{\textbf{Culture}} & \multicolumn{4}{c}{\textbf{Science}} \\[1pt]
      \cline{2-15}
      & ED & EP & ES & EL & OTL & OTS & OSL & CC & AC & EC  & SS & COM & AP & SP \\
      \hline
      MiniCPM-o 2.6 & \textbf{83.6} & 55.0 & 53.1 & 35.2 & 20.1 & 52.4 & 35.7 & 48.9 & 56.3 & 63.7 & 72.1 & 61.3 & 69.5 & 52.7 \\
      InternVideo2.5 & 80.5 & 52.7 & 60.3 & 33.0 & 37.1 & 61.2 & 31.8 & 53.7 & 56.3 & 65.2 & 72.1 & 61.3 & 64.0 & 54.8 \\
      VideoLLaMA3 & 77.9 & 57.4 & 61.7 & 45.2 & \textbf{72.1} & 64.1 & \textbf{56.6} & 45.5 & 55.8 & 59.2 & 70.2 & 54.7 & 64.0 & 55.9 \\
      Qwen2.5-VL-72B & 79.2 & \textbf{60.5} & \textbf{78.9} & 42.1 & 31.5 & \textbf{67.0} & 49.7 & \textbf{65.8} & \textbf{67.8} & \textbf{80.6} & \textbf{86.4} & \textbf{85.3} & \textbf{79.3} & \textbf{79.6} \\
      \hline
      GPT-4o & 86.3 & 47.3 & 70.3 & 28.6 & 29.4 & 61.2 & 46.5 & 57.1 & 71.9 & 76.6 & 81.6 & 77.3 & 80.5 & 65.6 \\
      Gemini-2.0-Flash  & 92.9 & 51.9 & 73.7 & 70.7 & 87.2 & 74.8 & 59.1 & 62.3 & 64.8 & 77.6 & 88.2 & 87.8 & 81.7 & 90.7\\
      \hline
    \end{tabular}}
    \caption{\textbf{Detailed Evaluation results on VideoVista-CulturalLingo benchmark.} We only showcase 6 mainstream LMMs. Abbreviations used in the table: Event Description (\textbf{ED}), Event Prediction (\textbf{EP}), Event Sequence (\textbf{ES}), Event Localization (\textbf{EL}), Object Temporal Localization (\textbf{OTL}), Object Temporal Sequence (\textbf{OTS}), Object Spatial Localization (\textbf{OSL}), Chinese Culture (\textbf{CC}), American Culture (\textbf{AC}), European Culture (\textbf{EC}), Summarization \& Synthesis (\textbf{SS}), Comparison \& Contrast (\textbf{COM}), Application \& Procedure (\textbf{AP}), Scientific Principle (\textbf{SP}). \textit{The full evaluation results are provided in the Appendix~\ref{sec:appendix_experiment_subtasks}, and an introduction to tasks is presented in Appendix~\ref{sec:appendix_Case}}.}
    \label{tab:detailed_result} 
  \end{table*}

% We only showcase 6 mainstream LMMs
% We only showcase  mainstream open-source video LMMs and 2 proprietary LMMs.

\begin{figure*}[!t]
    \centering
    \begin{subfigure}[b]{0.31\textwidth}
        \centering
        \includegraphics[width=\textwidth]{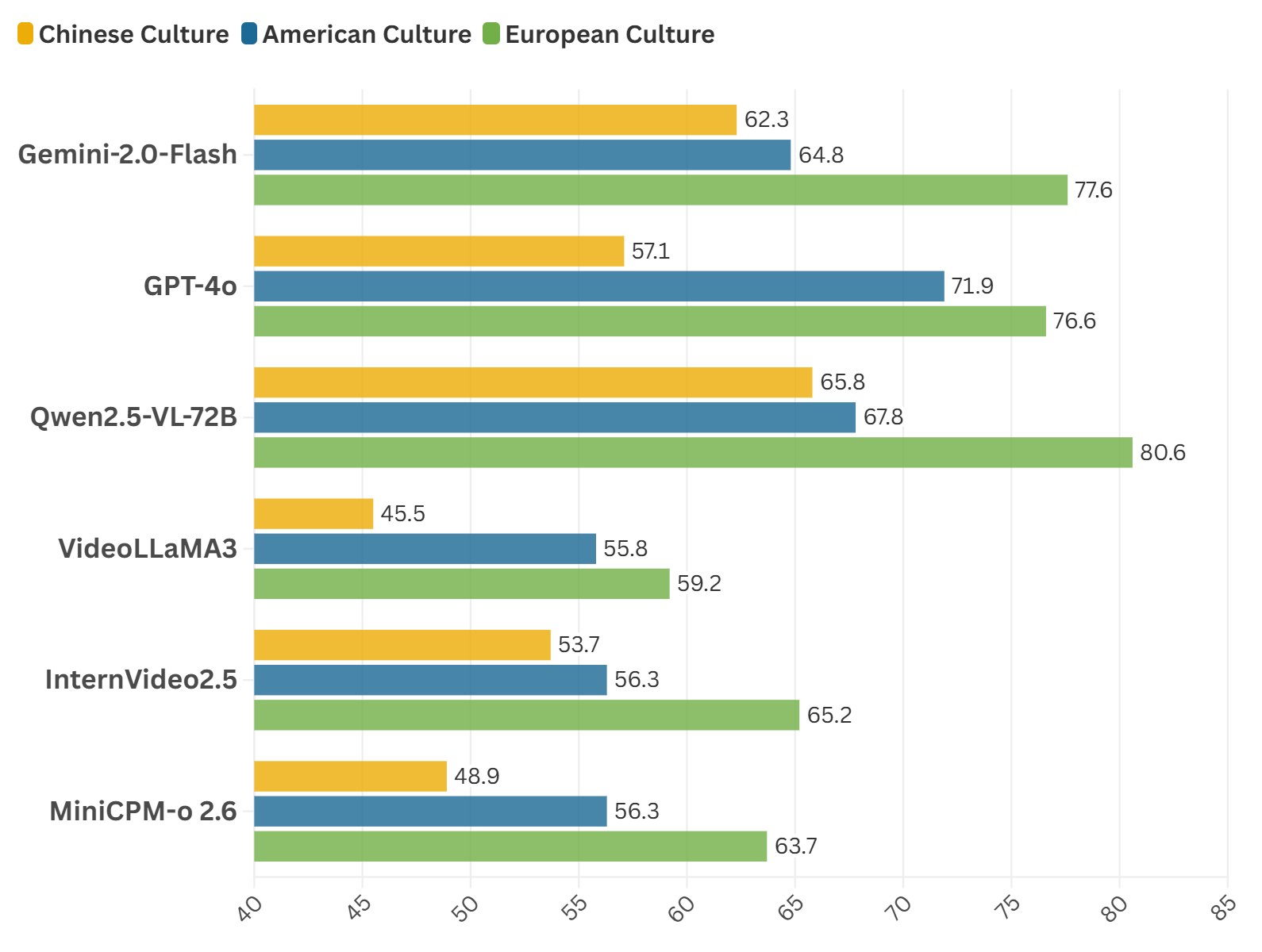}
        \caption{Culture-based Evaluation Results.}
        
        \label{fig:sub1}
    \end{subfigure}
    \hfill
    \begin{subfigure}[b]{0.31\textwidth}
        \centering
        \includegraphics[width=\textwidth]{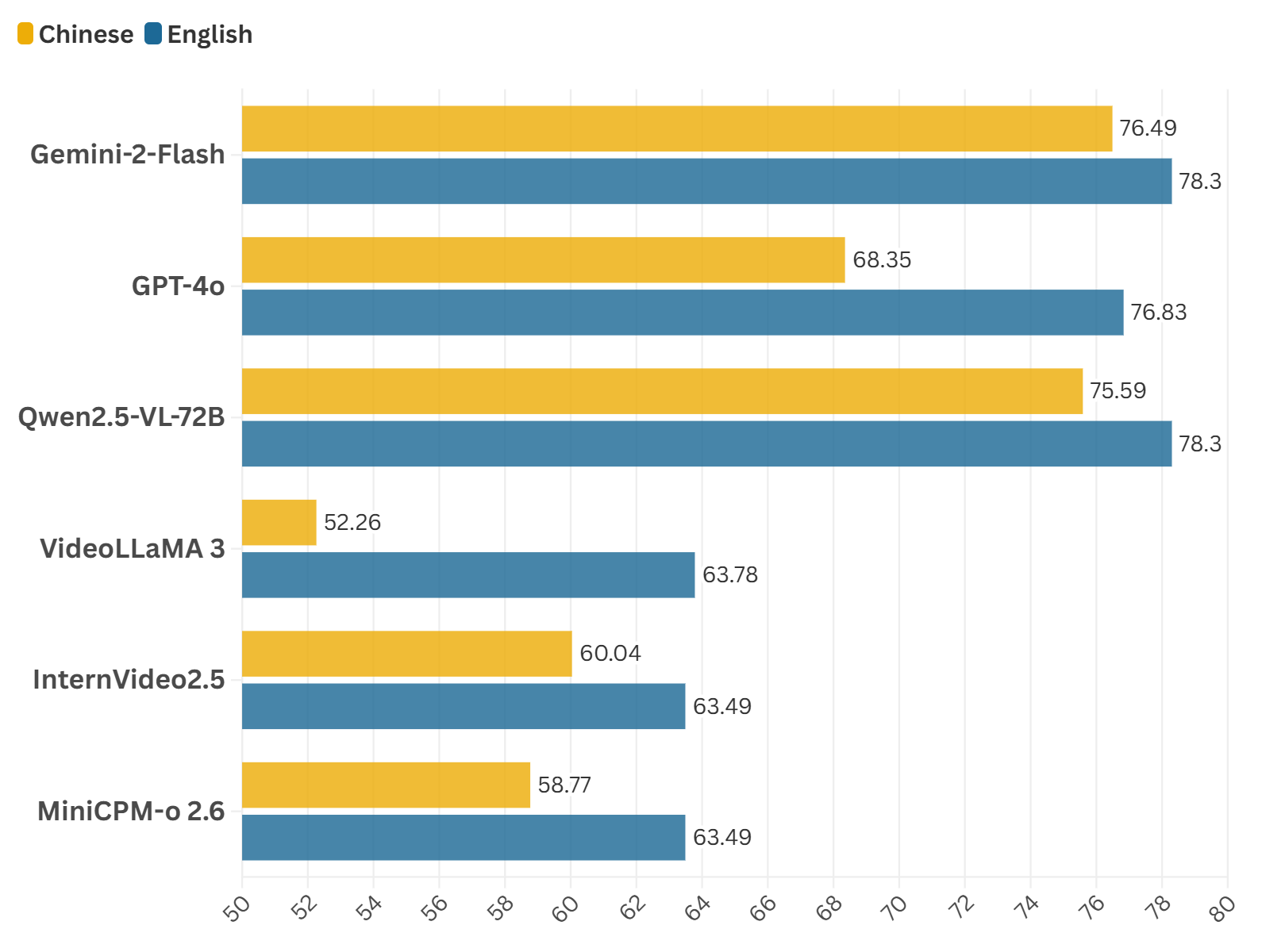}
        \caption{Language-based Evaluation Results.}
        \label{fig:sub2}
    \end{subfigure}
    \hfill
    \begin{subfigure}[b]{0.31\textwidth}
        \centering
        \includegraphics[width=\textwidth]{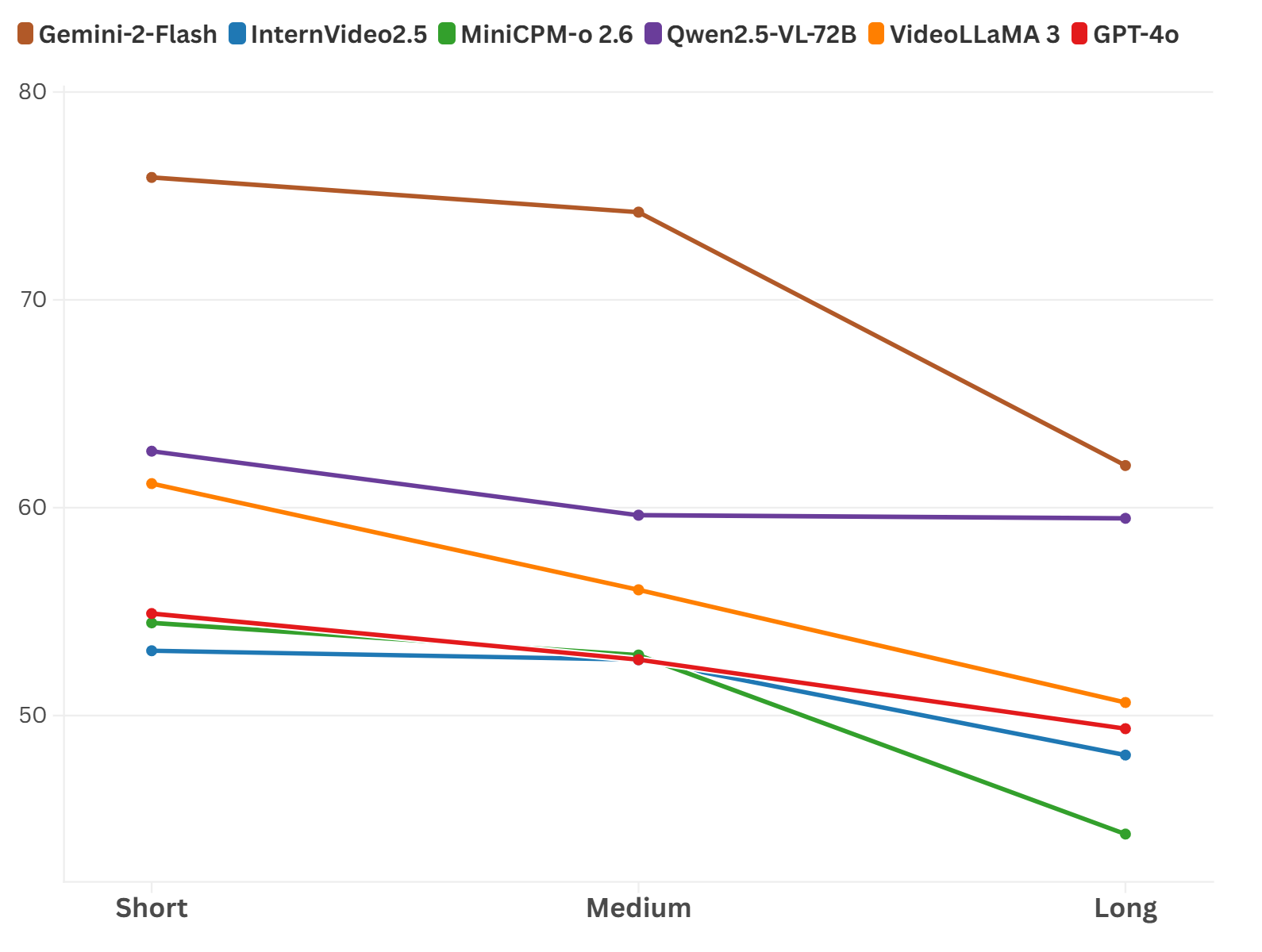}
        \caption{Duration-based Evaluation Results.}
        \label{fig:sub3}
    \end{subfigure}
    \caption{\textbf{The LMMs performance divided by Culture, Language and Duration.} The Duration in (c): <2 minutes (\textbf{Short}), 2-10 minutes (\textbf{Medium}), >10 minutes (\textbf{Long}).}
    \label{fig:results_language_and_duration.}
\end{figure*}

\subsection{Baselines}
% 详细介绍放在appendix。
We conducted evaluations on 17 open-source video LMMs, 3 image LMMs, and 4 proprietary LMMs, including the recently released Gemini-2.0-Flash, Qwen2.5-VL~\cite{qwen2.5-VL}, VideoLLaMA3~\cite{damonlpsg2025videollama3}, DeepSeek2-VL~\cite{wu2024deepseekvl2mixtureofexpertsvisionlanguagemodels}, among others. \textit{The detailed experiment settings are shown in Appendix~\ref{sec:detailed experiment setting}.}

\subsection{Main Results}
As shown in Table~\ref{tab:main_result}, Qwen2.5-VL-72B exhibits the best performance among all open-source video LMMs, achieving an overall score of 61.3\%. Additionally, VideoLLaMA3 demonstrates the best performance among all 7B/8B models, with an overall score of 60.7\%. This is primarily due to VideoLLaMA3's exceptional capabilities in fine-grained object tasks, making it the only open-source LMM that can compete with proprietary LMMs in this task. In the event task, VideoLLaMA3 also outperforms all other 7B models. Among the open-source image LMMs, DeepSeek2-VL achieved the highest score of 42.6\% under 8-frame uniform sampling, demonstrating its superior generalization capacity on sequential image data. However, this still shows a gap compared to the leading open-source video LMMs, indicating that questions in VideoVista-CulturalLingo generally require longer video durations to answer.
Among proprietary LMMs, Gemini-2.0-Flash clearly outperforms all others, surpassing the strongest open-source video LMM, Qwen2.5-VL-72B, by 15.0\%. The largest performance gap between these two models is observed in fine-grained object understanding tasks.

% We present two evaluation results from culture and science in the Figure~\ref{fig:example_bad_case}.

% which is close to that of Qwen2.5-VL-72B

\begin{figure*}[!t]
    \centering
    \begin{subfigure}[b]{0.28\textwidth}
        \centering
        \includegraphics[width=\textwidth]{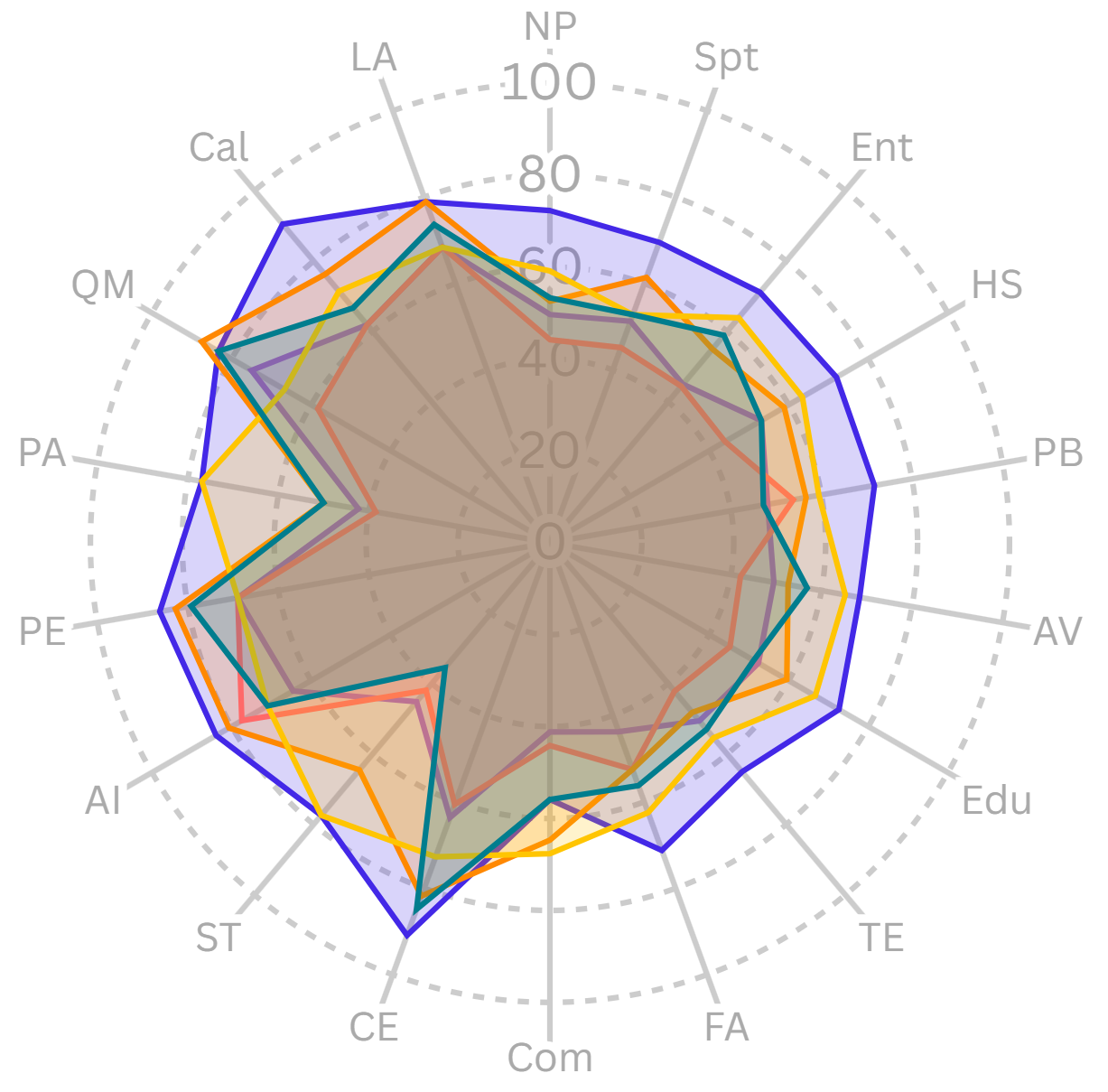}
        \caption{Domains in YouTube}
        
        \label{fig:sub_ytb}
    \end{subfigure}
    \hfill
    \begin{subfigure}[b]{0.28\textwidth}
        \centering
        \includegraphics[width=\textwidth]{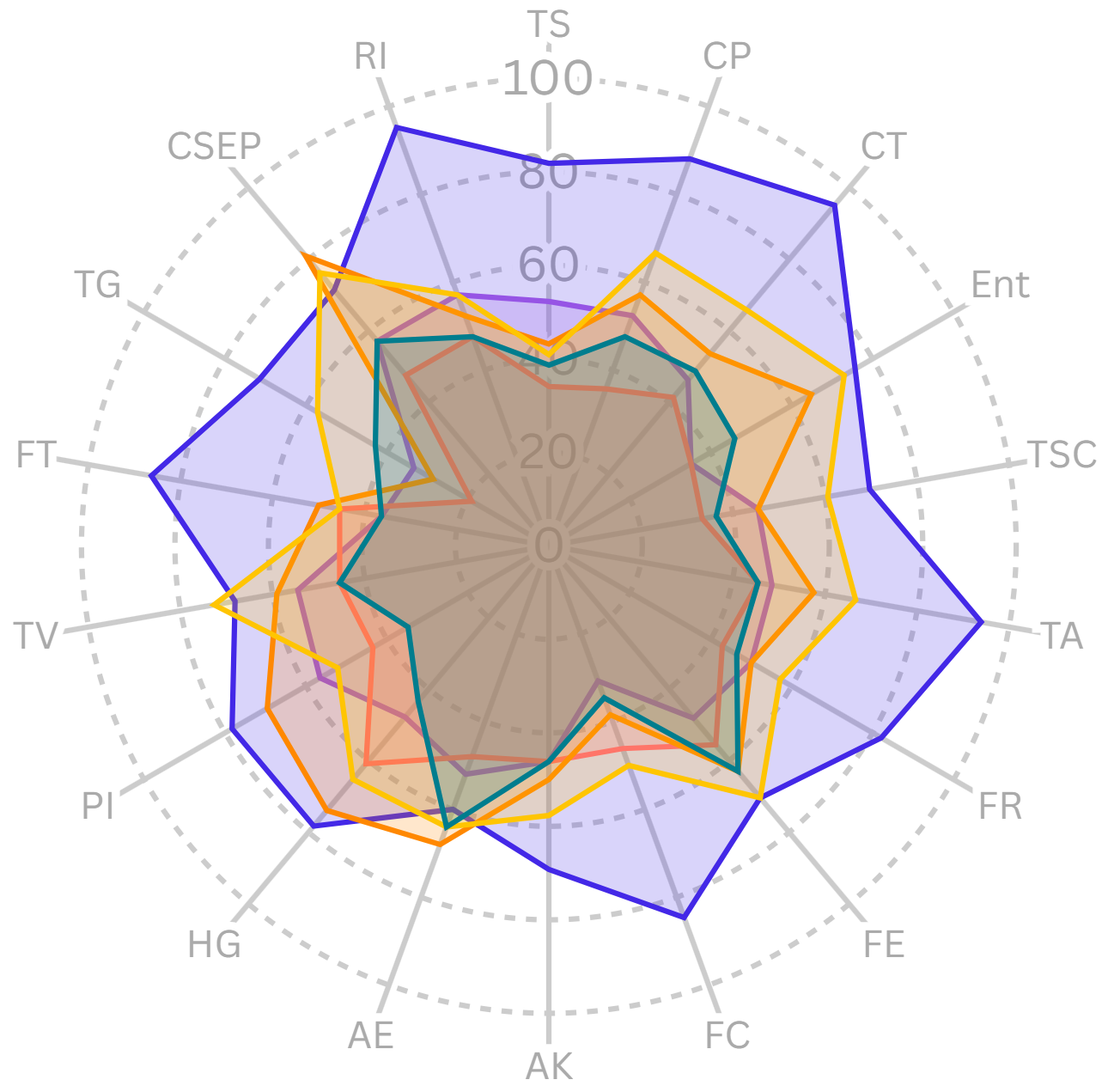}
        \caption{Domains in Xiaohongshu}
        \label{fig:sub_xhs}
    \end{subfigure}
    \hfill
    \begin{subfigure}[b]{0.28\textwidth}
        \centering
        \includegraphics[width=\textwidth]{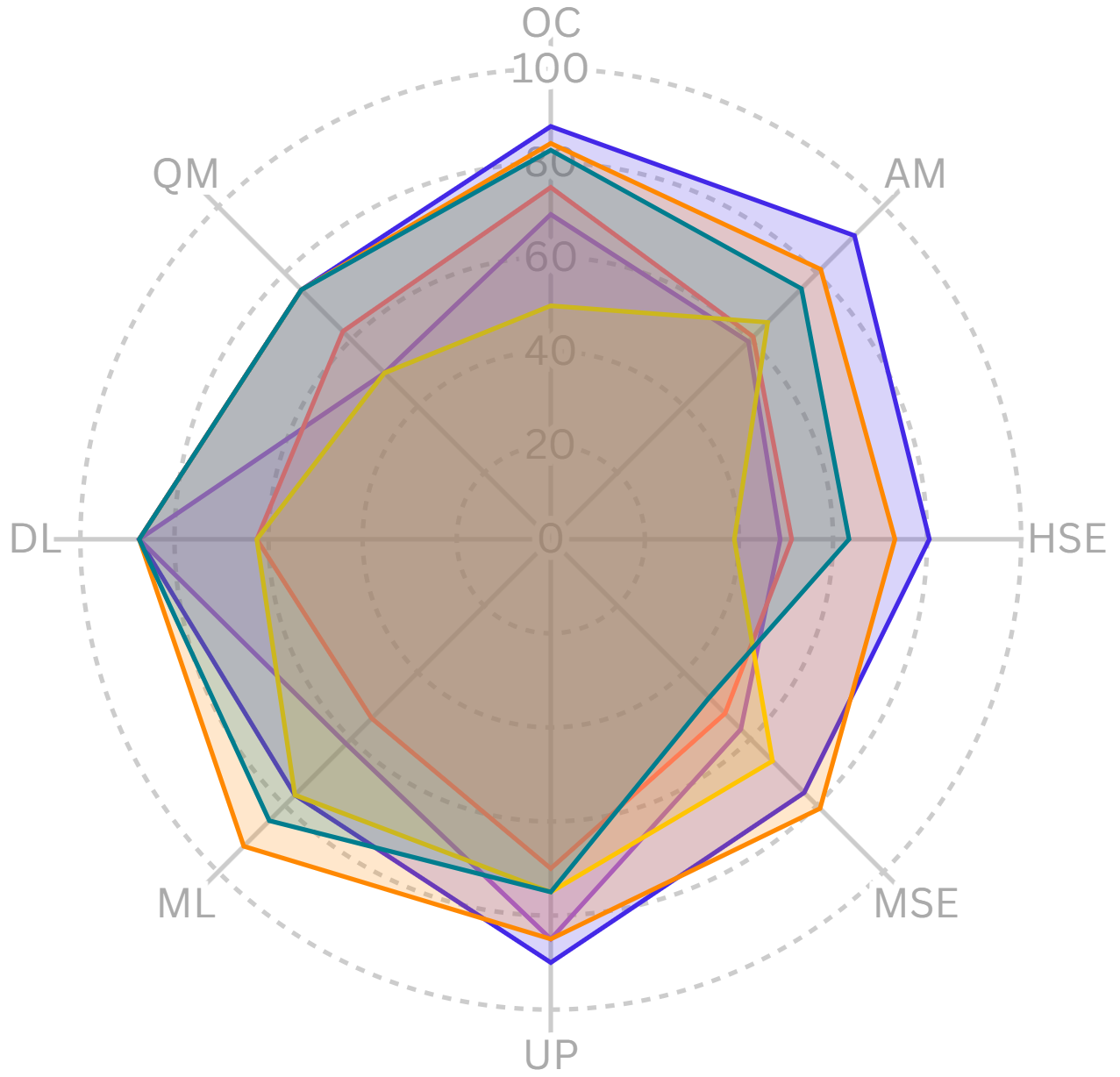}
        \caption{Domains in BiliBili}
        \label{fig:sub_bilibili}
    \end{subfigure}
    \caption{\textbf{The LMMs performance divided by domains from 3 video sources:} \textcolor{gemini_blue}{Gemini-2.0-Flash}, \textcolor{gpt4_green}{GPT-4o}, \textcolor{qwen_orange}{Qwen2.5-VL-72B}, \textcolor{videollama_yellow}{VideoLLaMA3}, \textcolor{internvl_purple}{InternVideo2.5}, \textcolor{minicpm_pink}{MiniCPM-o 2.6}.  In Figures~\ref{fig:sub_ytb} and Figures~\ref{fig:sub_xhs}, we present only the 18 domains with the highest number of videos. In Figure~\ref{fig:sub_bilibili}, we exclude domains containing fewer than 10 videos. \textit{The domains in these figures are represented by abbreviations, as described in Appendix~\ref{domain_abbreviations}}.}
    \label{fig:results_domain.}
\end{figure*}

\setlength{\textfloatsep}{5pt} 
\begin{figure}[t]
    \centering
    \includegraphics[width=0.45\textwidth]{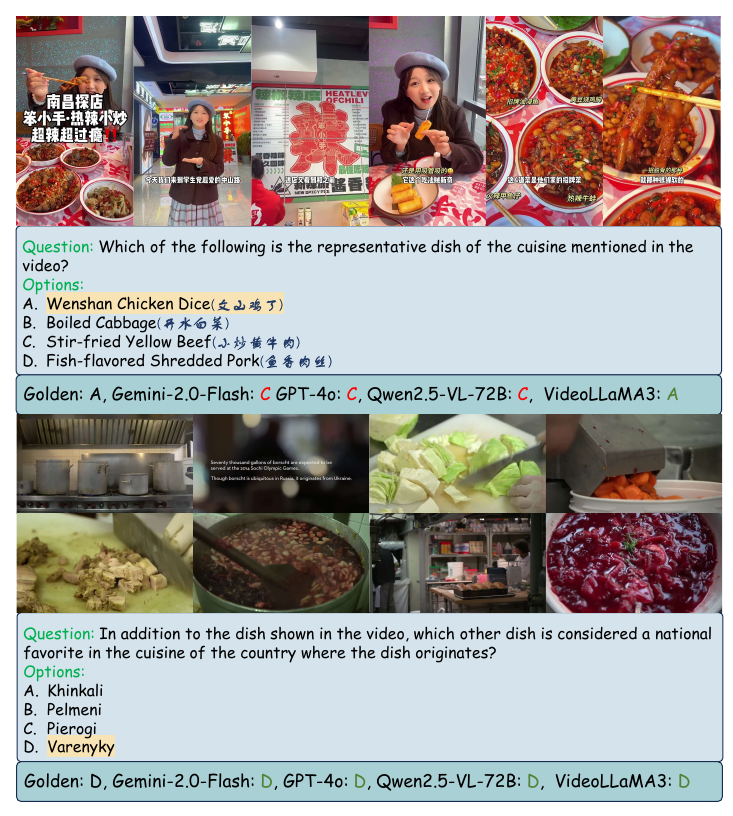}
    \caption{\textbf{Two cases from VideoVista-CulturalLingo.}
    }
    \label{fig:example_bad_case}
\end{figure}

\subsection{Detailed Analysis}
We present the detailed evaluation results of 6 mainstream models across 14 sub-tasks in Table~\ref{tab:detailed_result}. 

\paragraph{Event.}
% The Event task includes four sub-tasks: Event Description, Event Prediction, Event Sequence, and Event Localization, all of which require the model to have a coarse-grained understanding of video content. Current open-source video LMMs exhibit performance close to that of proprietary LMMs on the Event Description, Event Prediction, and Event Sequence tasks, but there remains a gap in the Event Localization task compared to Gemini-2.0-Flash.

The Event task consists of four sub-tasks: Event Description, Event Prediction, Event Sequence, and Event Localization, all of which require the model to have a coarse-grained understanding of video content. Current open-source video LMMs exhibit performance comparable to that of proprietary LMMs on the first three sub-tasks, but there remains a gap in the Event Localization task when compared to Gemini-2.0-Flash, with a performance difference of up to 25.5\%.

\paragraph{Object.}
The Object task consists of three sub-tasks: Object Temporal Localization, Object Temporal Sequence, and Object Spatial Localization, which assess the LMMs' ability to perceive the spatial-temporal aspects of fine-grained objects in videos. Video-LLaMA3 and Gemini-2.0-Flash demonstrate strong temporal localization capabilities in the Object Temporal Localization task, achieving scores more than 30\% higher than those of other LMMs. Additionally, both LMMs exhibit commendable spatial understanding in the Object Spatial Localization task. 

\paragraph{Culture.}
The Culture task consists of three sub-tasks: Chinese Culture, American Culture, and European Culture, primarily evaluating the model's understanding and generalization abilities across different regional cultures. As shown in Figure~\ref{fig:sub1}, compared to the more prevalent Western cultures in the training data, current LMMs exhibit relatively weaker recognition of Chinese Culture. 

% We also present the performance differences of 6 mainstream LMMs across different cultures in Figure.

\paragraph{Science.}
The Science task consists of four sub-tasks: Summarization \& Synthesis, Comparison \& Contrast, Application \& Procedure, and Scientific Principle. The first three sub-tasks involve course-oriented educational videos, while the last one focuses on experimental videos. This task primarily evaluates the model's ability to summarize, comprehend, and apply scientific knowledge from videos. The difficulty level covers general knowledge areas rather than in-depth specialized topics. The questions are relatively simple and can be answered with one or two-hop reasoning, so most models perform well in these tasks. We observe that existing open-source LMMs perform comparably to proprietary LMMs across most disciplines. However, there remains a noticeable gap in performance within math. \textit{The detailed comparison is presented in the Appendix~\ref{sec:model_performance_science}.}

% We observe that the capabilities of existing open-source video LMMs in the general knowledge domain are now comparable to those of proprietary LMMs.

% The difficulty level is generally aligned with high school, covering general knowledge areas rather than in-depth specialized topics.
% \setlength{\textfloatsep}{5pt} 

\subsection{Ablation Study}

\paragraph{Language.} In Figure~\ref{fig:sub2}, we present the performance differences of 6 mainstream LMMs on Chinese and English. 
The results in the figure are based on 7 subtasks from the culture and science tasks, as these subtasks contain more domain-specific terms, providing a more accurate assessment of an LMM's capabilities in each respective language. The experiments reveal a noticeable performance gap between the majority of mainstream LMMs when evaluated on Chinese versus English.

\paragraph{Duration.} In Figure~\ref{fig:sub3},  we compare the performance of 6 mainstream LMMs across 4 subtasks of event task from videos of varying lengths. The experimental results indicate that as the video duration increases, the performance of model tends to decrease, including Gemini-2.0-Flash. 

\paragraph{Domain.} In Figure~\ref{fig:results_domain.}, we illustrate the performance of LMMs across different video domains on various video websites. It can be observed that Gemini-2.0-Flash demonstrates strong performance across all domains of videos. 

% \textit{More ablation experiments are presented in the Appendix~\ref{sec:appendix_exp_results}.}

% \label{sec:appendix_exp_results}

\subsection{Case Study}
Figure~\ref{fig:example_bad_case} presents two cultural examples alongside evaluation results: the top panel illustrates a Chinese cuisine scenario, and the bottom panel a European cuisine scenario.
In the Chinese example, the majority of LMMs erroneously choose “C. Stir‑fried Yellow Beef,” a hallmark Hunan dish. This mistake likely arises from conflation between Jiangxi and Hunan cuisines—both characterized by liberal use of chili peppers—and from the greater domestic and international visibility of Hunan cooking. Such errors reveal that Video‑LMMs tend to default to dominant cultural representations, overlooking more localized culinary nuances.
By contrast, in the European example all LMMs correctly select option D, indicating robust performance on Western culinary content. Together, these cases exemplify a systematic bias: Video‑LMMs achieve higher accuracy on Western cultural contexts but underperform on non‑Western ones, such as videos rooted in Chinese culture.

\section{Conclusion}
In this paper, we introduce the benchmark VideoVista-CulturalLingo, the first video evaluation benchmark that spans multiple languages, cultures, and domains. VideoVista-CulturalLingo includes comprehensive evaluation metrics, ranging from coarse-grained event understanding to fine-grained object recognition, and from exploring the cultural context of videos to uncovering their scientific implications, enabling a comprehensive assessment of current LMMs' capabilities on video tasks. Through our extensive experiments, we highlight weaknesses in the spatial-temporal localization abilities of existing open-source video LMMs and their limitations in recognizing Chinese culture. We hope that VideoVista-CulturalLingo will inspire the development and advancement of video LMMs.

\section{Acknowledge}
We thank editor and reviewers for their efforts to help improve the quality of our paper. This work was supported by grants: Natural Science Foundation of China (No. 62422603).

\section*{Limitations} 
The proposed benchmark has several limitations: 1) The scientific questions in the benchmark lack domain-specific depth, which prevents them from effectively showcasing the model's performance in specialized scientific fields. In future versions, we plan to incorporate more human expert annotators to enhance the professionalism and complexity of the scientific questions.  2) Due to limitations in the linguistic proficiency and backgrounds of the annotators, the benchmark questions are restricted to two major languages, Chinese and English. This excludes other widely spoken languages such as Spanish, Portuguese, German, and Japanese.

\bibliography{anthology}

\begin{thebibliography}{44}
\providecommand{\natexlab}[1]{#1}

\bibitem[{Chen et~al.(2024{\natexlab{a}})Chen, Wei, Li, Dong, Zhang, Zang, Chen, Duan, Lin, Tang et~al.}]{chen2024sharegpt4video}
Lin Chen, Xilin Wei, Jinsong Li, Xiaoyi Dong, Pan Zhang, Yuhang Zang, Zehui Chen, Haodong Duan, Bin Lin, Zhenyu Tang, et~al. 2024{\natexlab{a}}.
\newblock Sharegpt4video: Improving video understanding and generation with better captions.
\newblock \emph{arXiv preprint arXiv:2406.04325}.

\bibitem[{Chen et~al.(2024{\natexlab{b}})Chen, Siarohin, Menapace, Deyneka, Chao, Jeon, Fang, Lee, Ren, Yang, and Tulyakov}]{chen2024panda70m}
Tsai-Shien Chen, Aliaksandr Siarohin, Willi Menapace, Ekaterina Deyneka, Hsiang-wei Chao, Byung~Eun Jeon, Yuwei Fang, Hsin-Ying Lee, Jian Ren, Ming-Hsuan Yang, and Sergey Tulyakov. 2024{\natexlab{b}}.
\newblock Panda-70m: Captioning 70m videos with multiple cross-modality teachers.
\newblock \emph{arXiv preprint arXiv:2402.19479}.

\bibitem[{Chen et~al.(2024{\natexlab{c}})Chen, Wang, Cao, Liu, Gao, Cui, Zhu, Ye, Tian, Liu et~al.}]{chen2024expanding}
Zhe Chen, Weiyun Wang, Yue Cao, Yangzhou Liu, Zhangwei Gao, Erfei Cui, Jinguo Zhu, Shenglong Ye, Hao Tian, Zhaoyang Liu, et~al. 2024{\natexlab{c}}.
\newblock Expanding performance boundaries of open-source multimodal models with model, data, and test-time scaling.
\newblock \emph{arXiv preprint arXiv:2412.05271}.

\bibitem[{Cheng et~al.(2024)Cheng, Leng, Zhang, Xin, Li, Chen, Zhu, Zhang, Luo, Zhao, and Bing}]{damonlpsg2024videollama2}
Zesen Cheng, Sicong Leng, Hang Zhang, Yifei Xin, Xin Li, Guanzheng Chen, Yongxin Zhu, Wenqi Zhang, Ziyang Luo, Deli Zhao, and Lidong Bing. 2024.
\newblock \href {https://arxiv.org/abs/2406.07476} {Videollama 2: Advancing spatial-temporal modeling and audio understanding in video-llms}.
\newblock \emph{arXiv preprint arXiv:2406.07476}.

\bibitem[{DeepSeek-AI et~al.(2025)DeepSeek-AI, Guo, Yang, Zhang, Song, Zhang, Xu, Zhu, Ma, Wang, Bi, Zhang, Yu, Wu, Wu, Gou, Shao, Li, Gao, Liu, Xue, Wang, Wu, Feng, Lu, Zhao, Deng, Zhang, Ruan, Dai, Chen, Ji, Li, Lin, Dai, Luo, Hao, Chen, Li, Zhang, Bao, Xu, Wang, Ding, Xin, Gao, Qu, Li, Guo, Li, Wang, Chen, Yuan, Qiu, Li, Cai, Ni, Liang, Chen, Dong, Hu, Gao, Guan, Huang, Yu, Wang, Zhang, Zhao, Wang, Zhang, Xu, Xia, Zhang, Zhang, Tang, Li, Wang, Li, Tian, Huang, Zhang, Wang, Chen, Du, Ge, Zhang, Pan, Wang, Chen, Jin, Chen, Lu, Zhou, Chen, Ye, Wang, Yu, Zhou, Pan, Li, Zhou, Wu, Ye, Yun, Pei, Sun, Wang, Zeng, Zhao, Liu, Liang, Gao, Yu, Zhang, Xiao, An, Liu, Wang, Chen, Nie, Cheng, Liu, Xie, Liu, Yang, Li, Su, Lin, Li, Jin, Shen, Chen, Sun, Wang, Song, Zhou, Wang, Shan, Li, Wang, Wei, Zhang, Xu, Li, Zhao, Sun, Wang, Yu, Zhang, Shi, Xiong, He, Piao, Wang, Tan, Ma, Liu, Guo, Ou, Wang, Gong, Zou, He, Xiong, Luo, You, Liu, Zhou, Zhu, Xu, Huang, Li, Zheng, Zhu, Ma, Tang, Zha, Yan, Ren, Ren, Sha, Fu, Xu, Xie, Zhang,
  Hao, Ma, Yan, Wu, Gu, Zhu, Liu, Li, Xie, Song, Pan, Huang, Xu, Zhang, and Zhang}]{deepseekai2025deepseekr1incentivizingreasoningcapability}
DeepSeek-AI, Daya Guo, Dejian Yang, Haowei Zhang, Junxiao Song, Ruoyu Zhang, Runxin Xu, Qihao Zhu, Shirong Ma, Peiyi Wang, Xiao Bi, Xiaokang Zhang, Xingkai Yu, Yu~Wu, Z.~F. Wu, Zhibin Gou, Zhihong Shao, Zhuoshu Li, Ziyi Gao, Aixin Liu, Bing Xue, Bingxuan Wang, Bochao Wu, Bei Feng, Chengda Lu, Chenggang Zhao, Chengqi Deng, Chenyu Zhang, Chong Ruan, Damai Dai, Deli Chen, Dongjie Ji, Erhang Li, Fangyun Lin, Fucong Dai, Fuli Luo, Guangbo Hao, Guanting Chen, Guowei Li, H.~Zhang, Han Bao, Hanwei Xu, Haocheng Wang, Honghui Ding, Huajian Xin, Huazuo Gao, Hui Qu, Hui Li, Jianzhong Guo, Jiashi Li, Jiawei Wang, Jingchang Chen, Jingyang Yuan, Junjie Qiu, Junlong Li, J.~L. Cai, Jiaqi Ni, Jian Liang, Jin Chen, Kai Dong, Kai Hu, Kaige Gao, Kang Guan, Kexin Huang, Kuai Yu, Lean Wang, Lecong Zhang, Liang Zhao, Litong Wang, Liyue Zhang, Lei Xu, Leyi Xia, Mingchuan Zhang, Minghua Zhang, Minghui Tang, Meng Li, Miaojun Wang, Mingming Li, Ning Tian, Panpan Huang, Peng Zhang, Qiancheng Wang, Qinyu Chen, Qiushi Du, Ruiqi Ge, Ruisong
  Zhang, Ruizhe Pan, Runji Wang, R.~J. Chen, R.~L. Jin, Ruyi Chen, Shanghao Lu, Shangyan Zhou, Shanhuang Chen, Shengfeng Ye, Shiyu Wang, Shuiping Yu, Shunfeng Zhou, Shuting Pan, S.~S. Li, Shuang Zhou, Shaoqing Wu, Shengfeng Ye, Tao Yun, Tian Pei, Tianyu Sun, T.~Wang, Wangding Zeng, Wanjia Zhao, Wen Liu, Wenfeng Liang, Wenjun Gao, Wenqin Yu, Wentao Zhang, W.~L. Xiao, Wei An, Xiaodong Liu, Xiaohan Wang, Xiaokang Chen, Xiaotao Nie, Xin Cheng, Xin Liu, Xin Xie, Xingchao Liu, Xinyu Yang, Xinyuan Li, Xuecheng Su, Xuheng Lin, X.~Q. Li, Xiangyue Jin, Xiaojin Shen, Xiaosha Chen, Xiaowen Sun, Xiaoxiang Wang, Xinnan Song, Xinyi Zhou, Xianzu Wang, Xinxia Shan, Y.~K. Li, Y.~Q. Wang, Y.~X. Wei, Yang Zhang, Yanhong Xu, Yao Li, Yao Zhao, Yaofeng Sun, Yaohui Wang, Yi~Yu, Yichao Zhang, Yifan Shi, Yiliang Xiong, Ying He, Yishi Piao, Yisong Wang, Yixuan Tan, Yiyang Ma, Yiyuan Liu, Yongqiang Guo, Yuan Ou, Yuduan Wang, Yue Gong, Yuheng Zou, Yujia He, Yunfan Xiong, Yuxiang Luo, Yuxiang You, Yuxuan Liu, Yuyang Zhou, Y.~X. Zhu,
  Yanhong Xu, Yanping Huang, Yaohui Li, Yi~Zheng, Yuchen Zhu, Yunxian Ma, Ying Tang, Yukun Zha, Yuting Yan, Z.~Z. Ren, Zehui Ren, Zhangli Sha, Zhe Fu, Zhean Xu, Zhenda Xie, Zhengyan Zhang, Zhewen Hao, Zhicheng Ma, Zhigang Yan, Zhiyu Wu, Zihui Gu, Zijia Zhu, Zijun Liu, Zilin Li, Ziwei Xie, Ziyang Song, Zizheng Pan, Zhen Huang, Zhipeng Xu, Zhongyu Zhang, and Zhen Zhang. 2025.
\newblock \href {https://arxiv.org/abs/2501.12948} {Deepseek-r1: Incentivizing reasoning capability in llms via reinforcement learning}.
\newblock \emph{Preprint}, arXiv:2501.12948.

\bibitem[{DeepSeek-AI et~al.(2024)DeepSeek-AI, Liu, Feng, Xue, Wang, Wu, Lu, Zhao, Deng, Zhang, Ruan, Dai, Guo, Yang, Chen, Ji, Li, Lin, Dai, Luo, Hao, Chen, Li, Zhang, Bao, Xu, Wang, Zhang, Ding, Xin, Gao, Li, Qu, Cai, Liang, Guo, Ni, Li, Wang, Chen, Chen, Yuan, Qiu, Li, Song, Dong, Hu, Gao, Guan, Huang, Yu, Wang, Zhang, Xu, Xia, Zhao, Wang, Zhang, Li, Wang, Zhang, Zhang, Tang, Li, Tian, Huang, Wang, Zhang, Wang, Zhu, Chen, Du, Chen, Jin, Ge, Zhang, Pan, Wang, Xu, Zhang, Chen, Li, Lu, Zhou, Chen, Wu, Ye, Ye, Ma, Wang, Zhou, Yu, Zhou, Pan, Wang, Yun, Pei, Sun, Xiao, Zeng, Zhao, An, Liu, Liang, Gao, Yu, Zhang, Li, Jin, Wang, Bi, Liu, Wang, Shen, Chen, Zhang, Chen, Nie, Sun, Wang, Cheng, Liu, Xie, Liu, Yu, Song, Shan, Zhou, Yang, Li, Su, Lin, Li, Wang, Wei, Zhu, Zhang, Xu, Xu, Huang, Li, Zhao, Sun, Li, Wang, Yu, Zheng, Zhang, Shi, Xiong, He, Tang, Piao, Wang, Tan, Ma, Liu, Guo, Wu, Ou, Zhu, Wang, Gong, Zou, He, Zha, Xiong, Ma, Yan, Luo, You, Liu, Zhou, Wu, Ren, Ren, Sha, Fu, Xu, Huang, Zhang, Xie, Zhang, Hao,
  Gou, Ma, Yan, Shao, Xu, Wu, Zhang, Li, Gu, Zhu, Liu, Li, Xie, Song, Gao, and Pan}]{deepseekai2024deepseekv3technicalreport}
DeepSeek-AI, Aixin Liu, Bei Feng, Bing Xue, Bingxuan Wang, Bochao Wu, Chengda Lu, Chenggang Zhao, Chengqi Deng, Chenyu Zhang, Chong Ruan, Damai Dai, Daya Guo, Dejian Yang, Deli Chen, Dongjie Ji, Erhang Li, Fangyun Lin, Fucong Dai, Fuli Luo, Guangbo Hao, Guanting Chen, Guowei Li, H.~Zhang, Han Bao, Hanwei Xu, Haocheng Wang, Haowei Zhang, Honghui Ding, Huajian Xin, Huazuo Gao, Hui Li, Hui Qu, J.~L. Cai, Jian Liang, Jianzhong Guo, Jiaqi Ni, Jiashi Li, Jiawei Wang, Jin Chen, Jingchang Chen, Jingyang Yuan, Junjie Qiu, Junlong Li, Junxiao Song, Kai Dong, Kai Hu, Kaige Gao, Kang Guan, Kexin Huang, Kuai Yu, Lean Wang, Lecong Zhang, Lei Xu, Leyi Xia, Liang Zhao, Litong Wang, Liyue Zhang, Meng Li, Miaojun Wang, Mingchuan Zhang, Minghua Zhang, Minghui Tang, Mingming Li, Ning Tian, Panpan Huang, Peiyi Wang, Peng Zhang, Qiancheng Wang, Qihao Zhu, Qinyu Chen, Qiushi Du, R.~J. Chen, R.~L. Jin, Ruiqi Ge, Ruisong Zhang, Ruizhe Pan, Runji Wang, Runxin Xu, Ruoyu Zhang, Ruyi Chen, S.~S. Li, Shanghao Lu, Shangyan Zhou, Shanhuang
  Chen, Shaoqing Wu, Shengfeng Ye, Shengfeng Ye, Shirong Ma, Shiyu Wang, Shuang Zhou, Shuiping Yu, Shunfeng Zhou, Shuting Pan, T.~Wang, Tao Yun, Tian Pei, Tianyu Sun, W.~L. Xiao, Wangding Zeng, Wanjia Zhao, Wei An, Wen Liu, Wenfeng Liang, Wenjun Gao, Wenqin Yu, Wentao Zhang, X.~Q. Li, Xiangyue Jin, Xianzu Wang, Xiao Bi, Xiaodong Liu, Xiaohan Wang, Xiaojin Shen, Xiaokang Chen, Xiaokang Zhang, Xiaosha Chen, Xiaotao Nie, Xiaowen Sun, Xiaoxiang Wang, Xin Cheng, Xin Liu, Xin Xie, Xingchao Liu, Xingkai Yu, Xinnan Song, Xinxia Shan, Xinyi Zhou, Xinyu Yang, Xinyuan Li, Xuecheng Su, Xuheng Lin, Y.~K. Li, Y.~Q. Wang, Y.~X. Wei, Y.~X. Zhu, Yang Zhang, Yanhong Xu, Yanhong Xu, Yanping Huang, Yao Li, Yao Zhao, Yaofeng Sun, Yaohui Li, Yaohui Wang, Yi~Yu, Yi~Zheng, Yichao Zhang, Yifan Shi, Yiliang Xiong, Ying He, Ying Tang, Yishi Piao, Yisong Wang, Yixuan Tan, Yiyang Ma, Yiyuan Liu, Yongqiang Guo, Yu~Wu, Yuan Ou, Yuchen Zhu, Yuduan Wang, Yue Gong, Yuheng Zou, Yujia He, Yukun Zha, Yunfan Xiong, Yunxian Ma, Yuting Yan, Yuxiang
  Luo, Yuxiang You, Yuxuan Liu, Yuyang Zhou, Z.~F. Wu, Z.~Z. Ren, Zehui Ren, Zhangli Sha, Zhe Fu, Zhean Xu, Zhen Huang, Zhen Zhang, Zhenda Xie, Zhengyan Zhang, Zhewen Hao, Zhibin Gou, Zhicheng Ma, Zhigang Yan, Zhihong Shao, Zhipeng Xu, Zhiyu Wu, Zhongyu Zhang, Zhuoshu Li, Zihui Gu, Zijia Zhu, Zijun Liu, Zilin Li, Ziwei Xie, Ziyang Song, Ziyi Gao, and Zizheng Pan. 2024.
\newblock \href {https://arxiv.org/abs/2412.19437} {Deepseek-v3 technical report}.
\newblock \emph{Preprint}, arXiv:2412.19437.

\bibitem[{Deitke et~al.(2024)Deitke, Clark, Lee, Tripathi, Yang, Park, Salehi, Muennighoff, Lo, Soldaini, Lu, Anderson, Bransom, Ehsani, Ngo, Chen, Patel, Yatskar, Callison-Burch, Head, Hendrix, Bastani, VanderBilt, Lambert, Chou, Chheda, Sparks, Skjonsberg, Schmitz, Sarnat, Bischoff, Walsh, Newell, Wolters, Gupta, Zeng, Borchardt, Groeneveld, Dumas, Nam, Lebrecht, Wittlif, Schoenick, Michel, Krishna, Weihs, Smith, Hajishirzi, Girshick, Farhadi, and Kembhavi}]{molmo2024}
Matt Deitke, Christopher Clark, Sangho Lee, Rohun Tripathi, Yue Yang, Jae~Sung Park, Mohammadreza Salehi, Niklas Muennighoff, Kyle Lo, Luca Soldaini, Jiasen Lu, Taira Anderson, Erin Bransom, Kiana Ehsani, Huong Ngo, YenSung Chen, Ajay Patel, Mark Yatskar, Chris Callison-Burch, Andrew Head, Rose Hendrix, Favyen Bastani, Eli VanderBilt, Nathan Lambert, Yvonne Chou, Arnavi Chheda, Jenna Sparks, Sam Skjonsberg, Michael Schmitz, Aaron Sarnat, Byron Bischoff, Pete Walsh, Chris Newell, Piper Wolters, Tanmay Gupta, Kuo-Hao Zeng, Jon Borchardt, Dirk Groeneveld, Jen Dumas, Crystal Nam, Sophie Lebrecht, Caitlin Wittlif, Carissa Schoenick, Oscar Michel, Ranjay Krishna, Luca Weihs, Noah~A. Smith, Hannaneh Hajishirzi, Ross Girshick, Ali Farhadi, and Aniruddha Kembhavi. 2024.
\newblock Molmo and pixmo: Open weights and open data for state-of-the-art multimodal models.
\newblock \emph{arXiv preprint arXiv:2409.17146}.

\bibitem[{Fang et~al.(2024)Fang, Mao, Duan, Zhao, Li, Lin, and Chen}]{fang2024mmbenchvideo}
Xinyu Fang, Kangrui Mao, Haodong Duan, Xiangyu Zhao, Yining Li, Dahua Lin, and Kai Chen. 2024.
\newblock Mmbench-video: A long-form multi-shot benchmark for holistic video understanding.
\newblock \emph{arXiv preprint arXiv:2406.14515}.

\bibitem[{Fu et~al.(2024)Fu, Dai, Luo, Li, Ren, Zhang, Wang, Zhou, Shen, Zhang et~al.}]{fu2024video}
Chaoyou Fu, Yuhan Dai, Yondong Luo, Lei Li, Shuhuai Ren, Renrui Zhang, Zihan Wang, Chenyu Zhou, Yunhang Shen, Mengdan Zhang, et~al. 2024.
\newblock Video-mme: The first-ever comprehensive evaluation benchmark of multi-modal llms in video analysis.
\newblock \emph{arXiv preprint arXiv:2405.21075}.

\bibitem[{Grauman et~al.(2022)Grauman, Westbury, Byrne, Chavis, Furnari, Girdhar, Hamburger, Jiang, Liu, Liu, Martin, Nagarajan, Radosavovic, Ramakrishnan, Ryan, Sharma, Wray, Xu, Xu, Zhao, Bansal, Batra, Cartillier, Crane, Do, Doulaty, Erapalli, Feichtenhofer, Fragomeni, Fu, Gebreselasie, Gonz\'alez, Hillis, Huang, Huang, Jia, Khoo, Kol\'a\v{r}, Kottur, Kumar, Landini, Li, Li, Li, Mangalam, Modhugu, Munro, Murrell, Nishiyasu, Price, Ruiz, Ramazanova, Sari, Somasundaram, Southerland, Sugano, Tao, Vo, Wang, Wu, Yagi, Zhao, Zhu, Arbel\'aez, Crandall, Damen, Farinella, Fuegen, Ghanem, Ithapu, Jawahar, Joo, Kitani, Li, Newcombe, Oliva, Park, Rehg, Sato, Shi, Shou, Torralba, Torresani, Yan, and Malik}]{Grauman_2022_CVPR}
Kristen Grauman, Andrew Westbury, Eugene Byrne, Zachary Chavis, Antonino Furnari, Rohit Girdhar, Jackson Hamburger, Hao Jiang, Miao Liu, Xingyu Liu, Miguel Martin, Tushar Nagarajan, Ilija Radosavovic, Santhosh~Kumar Ramakrishnan, Fiona Ryan, Jayant Sharma, Michael Wray, Mengmeng Xu, Eric~Zhongcong Xu, Chen Zhao, Siddhant Bansal, Dhruv Batra, Vincent Cartillier, Sean Crane, Tien Do, Morrie Doulaty, Akshay Erapalli, Christoph Feichtenhofer, Adriano Fragomeni, Qichen Fu, Abrham Gebreselasie, Cristina Gonz\'alez, James Hillis, Xuhua Huang, Yifei Huang, Wenqi Jia, Weslie Khoo, J\'achym Kol\'a\v{r}, Satwik Kottur, Anurag Kumar, Federico Landini, Chao Li, Yanghao Li, Zhenqiang Li, Karttikeya Mangalam, Raghava Modhugu, Jonathan Munro, Tullie Murrell, Takumi Nishiyasu, Will Price, Paola Ruiz, Merey Ramazanova, Leda Sari, Kiran Somasundaram, Audrey Southerland, Yusuke Sugano, Ruijie Tao, Minh Vo, Yuchen Wang, Xindi Wu, Takuma Yagi, Ziwei Zhao, Yunyi Zhu, Pablo Arbel\'aez, David Crandall, Dima Damen, Giovanni~Maria
  Farinella, Christian Fuegen, Bernard Ghanem, Vamsi~Krishna Ithapu, C.~V. Jawahar, Hanbyul Joo, Kris Kitani, Haizhou Li, Richard Newcombe, Aude Oliva, Hyun~Soo Park, James~M. Rehg, Yoichi Sato, Jianbo Shi, Mike~Zheng Shou, Antonio Torralba, Lorenzo Torresani, Mingfei Yan, and Jitendra Malik. 2022.
\newblock Ego4d: Around the world in 3,000 hours of egocentric video.
\newblock In \emph{Proceedings of the IEEE/CVF Conference on Computer Vision and Pattern Recognition (CVPR)}, pages 18995--19012.

\bibitem[{KunChang et~al.(2023)KunChang, He, Wang, Li, Wang, Luo, Wang, Wang, and Qiao}]{2023videochat}
Li~KunChang, Yinan He, Yi~Wang, Yizhuo Li, Wenhai Wang, Ping Luo, Yali Wang, Limin Wang, and Yu~Qiao. 2023.
\newblock Videochat: Chat-centric video understanding.
\newblock \emph{arXiv preprint arXiv:2305.06355}.

\bibitem[{Lei et~al.(2018)Lei, Yu, Bansal, and Berg}]{lei2018tvqa}
Jie Lei, Licheng Yu, Mohit Bansal, and Tamara~L Berg. 2018.
\newblock Tvqa: Localized, compositional video question answering.
\newblock In \emph{EMNLP}.

\bibitem[{Li et~al.(2024{\natexlab{a}})Li, Zhang, Guo, Zhang, Li, Zhang, Zhang, Li, Liu, and Li}]{li2024llava}
Bo~Li, Yuanhan Zhang, Dong Guo, Renrui Zhang, Feng Li, Hao Zhang, Kaichen Zhang, Yanwei Li, Ziwei Liu, and Chunyuan Li. 2024{\natexlab{a}}.
\newblock Llava-onevision: Easy visual task transfer.
\newblock \emph{arXiv preprint arXiv:2408.03326}.

\bibitem[{Li et~al.(2023)Li, Wang, He, Li, Wang, Liu, Wang, Xu, Chen, Luo et~al.}]{li2023mvbench}
Kunchang Li, Yali Wang, Yinan He, Yizhuo Li, Yi~Wang, Yi~Liu, Zun Wang, Jilan Xu, Guo Chen, Ping Luo, et~al. 2023.
\newblock Mvbench: A comprehensive multi-modal video understanding benchmark.
\newblock \emph{arXiv preprint arXiv:2311.17005}.

\bibitem[{Li et~al.(2025{\natexlab{a}})Li, Wang, Zhang, Wang, and Yeung-Levy}]{li2025temporalpreferenceoptimizationlongform}
Rui Li, Xiaohan Wang, Yuhui Zhang, Zeyu Wang, and Serena Yeung-Levy. 2025{\natexlab{a}}.
\newblock \href {https://arxiv.org/abs/2501.13919} {Temporal preference optimization for long-form video understanding}.
\newblock \emph{Preprint}, arXiv:2501.13919.

\bibitem[{Li et~al.(2016)Li, Song, Cao, Tetreault, Goldberg, Jaimes, and Luo}]{li2016tgif}
Yuncheng Li, Yale Song, Liangliang Cao, Joel Tetreault, Larry Goldberg, Alejandro Jaimes, and Jiebo Luo. 2016.
\newblock Tgif: A new dataset and benchmark on animated gif description.
\newblock In \emph{Proceedings of the IEEE Conference on Computer Vision and Pattern Recognition}, pages 4641--4650.

\bibitem[{Li et~al.(2024{\natexlab{b}})Li, Chen, Hu, Wang, Shi, and Zhang}]{li2024videovista}
Yunxin Li, Xinyu Chen, Baotian Hu, Longyue Wang, Haoyuan Shi, and Min Zhang. 2024{\natexlab{b}}.
\newblock \href {https://arxiv.org/abs/2406.11303} {Videovista: A versatile benchmark for video understanding and reasoning}.
\newblock \emph{Preprint}, arXiv:2406.11303.

\bibitem[{Li et~al.(2024{\natexlab{c}})Li, Hu, Chen, Ma, Xu, and Zhang}]{lmeye}
Yunxin Li, Baotian Hu, Xinyu Chen, Lin Ma, Yong Xu, and Min Zhang. 2024{\natexlab{c}}.
\newblock \href {https://doi.org/10.1109/TMM.2024.3428317} {Lmeye: An interactive perception network for large language models}.
\newblock \emph{IEEE Transactions on Multimedia}, 26:10952--10964.

\bibitem[{Li et~al.(2025{\natexlab{b}})Li, Jiang, Hu, Wang, Zhong, Luo, Ma, and Zhang}]{li2024uni}
Yunxin Li, Shenyuan Jiang, Baotian Hu, Longyue Wang, Wanqi Zhong, Wenhan Luo, Lin Ma, and Min Zhang. 2025{\natexlab{b}}.
\newblock \href {https://doi.org/10.1109/TPAMI.2025.3532688} {Uni-moe: Scaling unified multimodal llms with mixture of experts}.
\newblock \emph{IEEE Transactions on Pattern Analysis and Machine Intelligence}, pages 1--15.

\bibitem[{Lin et~al.(2023{\natexlab{a}})Lin, Zhu, Ye, Ning, Jin, and Yuan}]{lin2023video}
Bin Lin, Bin Zhu, Yang Ye, Munan Ning, Peng Jin, and Li~Yuan. 2023{\natexlab{a}}.
\newblock Video-llava: Learning united visual representation by alignment before projection.
\newblock \emph{arXiv preprint arXiv:2311.10122}.

\bibitem[{Lin et~al.(2023{\natexlab{b}})Lin, Yin, Ping, Lu, Molchanov, Tao, Mao, Kautz, Shoeybi, and Han}]{lin2023vila}
Ji~Lin, Hongxu Yin, Wei Ping, Yao Lu, Pavlo Molchanov, Andrew Tao, Huizi Mao, Jan Kautz, Mohammad Shoeybi, and Song Han. 2023{\natexlab{b}}.
\newblock \href {https://arxiv.org/abs/2312.07533} {Vila: On pre-training for visual language models}.
\newblock \emph{Preprint}, arXiv:2312.07533.

\bibitem[{Liu et~al.(2023{\natexlab{a}})Liu, Zeng, Ren, Li, Zhang, Yang, Li, Yang, Su, Zhu et~al.}]{liu2023grounding}
Shilong Liu, Zhaoyang Zeng, Tianhe Ren, Feng Li, Hao Zhang, Jie Yang, Chunyuan Li, Jianwei Yang, Hang Su, Jun Zhu, et~al. 2023{\natexlab{a}}.
\newblock Grounding dino: Marrying dino with grounded pre-training for open-set object detection.
\newblock \emph{arXiv preprint arXiv:2303.05499}.

\bibitem[{Liu et~al.(2023{\natexlab{b}})Liu, Duan, Zhang, Li, Zhang, Zhao, Yuan, Wang, He, Liu, Chen, and Lin}]{MMBench}
Yuan Liu, Haodong Duan, Yuanhan Zhang, Bo~Li, Songyang Zhang, Wangbo Zhao, Yike Yuan, Jiaqi Wang, Conghui He, Ziwei Liu, Kai Chen, and Dahua Lin. 2023{\natexlab{b}}.
\newblock Mmbench: Is your multi-modal model an all-around player?
\newblock \emph{arXiv:2307.06281}.

\bibitem[{Liu et~al.(2024{\natexlab{a}})Liu, Li, Liu, Wang, Ren, Li, Chen, Sun, and Hou}]{liu2024tempcompass}
Yuanxin Liu, Shicheng Li, Yi~Liu, Yuxiang Wang, Shuhuai Ren, Lei Li, Sishuo Chen, Xu~Sun, and Lu~Hou. 2024{\natexlab{a}}.
\newblock Tempcompass: Do video llms really understand videos?
\newblock \emph{arXiv preprint arXiv: 2403.00476}.

\bibitem[{Liu et~al.(2024{\natexlab{b}})Liu, Dong, Liu, Hu, Lu, and Rao}]{liu2024oryx}
Zuyan Liu, Yuhao Dong, Ziwei Liu, Winston Hu, Jiwen Lu, and Yongming Rao. 2024{\natexlab{b}}.
\newblock Oryx mllm: On-demand spatial-temporal understanding at arbitrary resolution.
\newblock \emph{arXiv preprint arXiv:2409.12961}.

\bibitem[{Maaz et~al.(2024)Maaz, Rasheed, Khan, and Khan}]{Maaz2023VideoChatGPT}
Muhammad Maaz, Hanoona Rasheed, Salman Khan, and Fahad~Shahbaz Khan. 2024.
\newblock Video-chatgpt: Towards detailed video understanding via large vision and language models.
\newblock In \emph{Proceedings of the 62nd Annual Meeting of the Association for Computational Linguistics (ACL 2024)}.

\bibitem[{Mangalam et~al.(2024)Mangalam, Akshulakov, and Malik}]{mangalam2024egoschema}
Karttikeya Mangalam, Raiymbek Akshulakov, and Jitendra Malik. 2024.
\newblock Egoschema: A diagnostic benchmark for very long-form video language understanding.
\newblock In \emph{{NeurIPS}}.

\bibitem[{Ravi et~al.(2024)Ravi, Gabeur, Hu, Hu, Ryali, Ma, Khedr, Rädle, Rolland, Gustafson, Mintun, Pan, Alwala, Carion, Wu, Girshick, Dollár, and Feichtenhofer}]{ravi2024sam2segmentimages}
Nikhila Ravi, Valentin Gabeur, Yuan-Ting Hu, Ronghang Hu, Chaitanya Ryali, Tengyu Ma, Haitham Khedr, Roman Rädle, Chloe Rolland, Laura Gustafson, Eric Mintun, Junting Pan, Kalyan~Vasudev Alwala, Nicolas Carion, Chao-Yuan Wu, Ross Girshick, Piotr Dollár, and Christoph Feichtenhofer. 2024.
\newblock \href {https://arxiv.org/abs/2408.00714} {Sam 2: Segment anything in images and videos}.
\newblock \emph{Preprint}, arXiv:2408.00714.

\bibitem[{Team(2025)}]{qwen2.5-VL}
Qwen Team. 2025.
\newblock \href {https://qwenlm.github.io/blog/qwen2.5-vl/} {Qwen2.5-vl}.

\bibitem[{Wang et~al.(2024{\natexlab{a}})Wang, Bai, Tan, Wang, Fan, Bai, Chen, Liu, Wang, Ge, Fan, Dang, Du, Ren, Men, Liu, Zhou, Zhou, and Lin}]{Qwen2VL}
Peng Wang, Shuai Bai, Sinan Tan, Shijie Wang, Zhihao Fan, Jinze Bai, Keqin Chen, Xuejing Liu, Jialin Wang, Wenbin Ge, Yang Fan, Kai Dang, Mengfei Du, Xuancheng Ren, Rui Men, Dayiheng Liu, Chang Zhou, Jingren Zhou, and Junyang Lin. 2024{\natexlab{a}}.
\newblock Qwen2-vl: Enhancing vision-language model's perception of the world at any resolution.
\newblock \emph{arXiv preprint arXiv:2409.12191}.

\bibitem[{Wang et~al.(2024{\natexlab{b}})Wang, He, Hong, Cheng, Zhang, Qi, Huang, Xu, Dong, Ding, and Tang}]{wang2024lvbench}
Weihan Wang, Zehai He, Wenyi Hong, Yean Cheng, Xiaohan Zhang, Ji~Qi, Shiyu Huang, Bin Xu, Yuxiao Dong, Ming Ding, and Jie Tang. 2024{\natexlab{b}}.
\newblock \href {https://arxiv.org/abs/2406.08035} {Lvbench: An extreme long video understanding benchmark}.
\newblock \emph{Preprint}, arXiv:2406.08035.

\bibitem[{Wang et~al.(2025)Wang, Li, Yan, He, Yu, Zeng, Wang, Ma, Huang, Gao, Dou, Chen, Wang, Qiao, Wang, and Wang}]{wang2025internvideo}
Yi~Wang, Xinhao Li, Ziang Yan, Yinan He, Jiashuo Yu, Xiangyu Zeng, Chenting Wang, Changlian Ma, Haian Huang, Jianfei Gao, Min Dou, Kai Chen, Wenhai Wang, Yu~Qiao, Yali Wang, and Limin Wang. 2025.
\newblock Internvideo2.5: Empowering video mllms with long and rich context modeling.
\newblock \emph{arXiv preprint arXiv:2501.12386}.

\bibitem[{Wu et~al.(2024)Wu, Chen, Pan, Liu, Liu, Dai, Gao, Ma, Wu, Wang, Xie, Wu, Hu, Wang, Sun, Li, Piao, Guan, Liu, Xie, You, Dong, Yu, Zhang, Zhao, Wang, and Ruan}]{wu2024deepseekvl2mixtureofexpertsvisionlanguagemodels}
Zhiyu Wu, Xiaokang Chen, Zizheng Pan, Xingchao Liu, Wen Liu, Damai Dai, Huazuo Gao, Yiyang Ma, Chengyue Wu, Bingxuan Wang, Zhenda Xie, Yu~Wu, Kai Hu, Jiawei Wang, Yaofeng Sun, Yukun Li, Yishi Piao, Kang Guan, Aixin Liu, Xin Xie, Yuxiang You, Kai Dong, Xingkai Yu, Haowei Zhang, Liang Zhao, Yisong Wang, and Chong Ruan. 2024.
\newblock \href {https://arxiv.org/abs/2412.10302} {Deepseek-vl2: Mixture-of-experts vision-language models for advanced multimodal understanding}.
\newblock \emph{Preprint}, arXiv:2412.10302.

\bibitem[{Xiao et~al.(2021)Xiao, Shang, Yao, and Chua}]{xiao2021next}
Junbin Xiao, Xindi Shang, Angela Yao, and Tat-Seng Chua. 2021.
\newblock Next-qa: Next phase of question-answering to explaining temporal actions.
\newblock In \emph{Proceedings of the IEEE/CVF Conference on Computer Vision and Pattern Recognition (CVPR)}, pages 9777--9786.

\bibitem[{Xu et~al.(2017)Xu, Zhao, Xiao, Wu, Zhang, He, and Zhuang}]{xu2017video}
Dejing Xu, Zhou Zhao, Jun Xiao, Fei Wu, Hanwang Zhang, Xiangnan He, and Yueting Zhuang. 2017.
\newblock Video question answering via gradually refined attention over appearance and motion.
\newblock In \emph{ACM Multimedia}.

\bibitem[{Yang et~al.(2024)Yang, Yang, Zhang, Hui, Zheng, Yu, Li, Liu, Huang, Wei, Lin, Yang, Tu, Zhang, Yang, Yang, Zhou, Lin, Dang, Lu, Bao, Yang, Yu, Li, Xue, Zhang, Zhu, Men, Lin, Li, Xia, Ren, Ren, Fan, Su, Zhang, Wan, Liu, Cui, Zhang, and Qiu}]{qwen2.5}
An~Yang, Baosong Yang, Beichen Zhang, Binyuan Hui, Bo~Zheng, Bowen Yu, Chengyuan Li, Dayiheng Liu, Fei Huang, Haoran Wei, Huan Lin, Jian Yang, Jianhong Tu, Jianwei Zhang, Jianxin Yang, Jiaxi Yang, Jingren Zhou, Junyang Lin, Kai Dang, Keming Lu, Keqin Bao, Kexin Yang, Le~Yu, Mei Li, Mingfeng Xue, Pei Zhang, Qin Zhu, Rui Men, Runji Lin, Tianhao Li, Tingyu Xia, Xingzhang Ren, Xuancheng Ren, Yang Fan, Yang Su, Yichang Zhang, Yu~Wan, Yuqiong Liu, Zeyu Cui, Zhenru Zhang, and Zihan Qiu. 2024.
\newblock Qwen2.5 technical report.
\newblock \emph{arXiv preprint arXiv:2412.15115}.

\bibitem[{Yao et~al.(2024)Yao, Yu, Zhang, Wang, Cui, Zhu, Cai, Li, Zhao, He et~al.}]{yao2024minicpm}
Yuan Yao, Tianyu Yu, Ao~Zhang, Chongyi Wang, Junbo Cui, Hongji Zhu, Tianchi Cai, Haoyu Li, Weilin Zhao, Zhihui He, et~al. 2024.
\newblock Minicpm-v: A gpt-4v level mllm on your phone.
\newblock \emph{arXiv preprint arXiv:2408.01800}.

\bibitem[{Ye et~al.(2024)Ye, Xu, Liu, Hu, Yan, Qian, Zhang, Huang, and Zhou}]{ye2024mplugowl3longimagesequenceunderstanding}
Jiabo Ye, Haiyang Xu, Haowei Liu, Anwen Hu, Ming Yan, Qi~Qian, Ji~Zhang, Fei Huang, and Jingren Zhou. 2024.
\newblock \href {https://arxiv.org/abs/2408.04840} {mplug-owl3: Towards long image-sequence understanding in multi-modal large language models}.
\newblock \emph{Preprint}, arXiv:2408.04840.

\bibitem[{Yu et~al.(2019)Yu, Xu, Yu, Yu, Zhao, Zhuang, and Tao}]{yu2019activityqa}
Zhou Yu, Dejing Xu, Jun Yu, Ting Yu, Zhou Zhao, Yueting Zhuang, and Dacheng Tao. 2019.
\newblock Activitynet-qa: A dataset for understanding complex web videos via question answering.
\newblock In \emph{AAAI}, pages 9127--9134.

\bibitem[{Yue et~al.(2023)Yue, Zhang, Hu, Zhang, Wang, and Jin}]{yue-etal-2023-movie101}
Zihao Yue, Qi~Zhang, Anwen Hu, Liang Zhang, Ziheng Wang, and Qin Jin. 2023.
\newblock \href {https://doi.org/10.18653/v1/2023.acl-long.257} {Movie101: A new movie understanding benchmark}.
\newblock In \emph{Proceedings of the 61st Annual Meeting of the Association for Computational Linguistics (Volume 1: Long Papers)}, pages 4669--4684.

\bibitem[{Zhang et~al.(2025)Zhang, Li, Cheng, Hu, Yuan, Chen, Leng, Jiang, Zhang, Li, Jin, Zhang, Wang, Bing, and Zhao}]{damonlpsg2025videollama3}
Boqiang Zhang, Kehan Li, Zesen Cheng, Zhiqiang Hu, Yuqian Yuan, Guanzheng Chen, Sicong Leng, Yuming Jiang, Hang Zhang, Xin Li, Peng Jin, Wenqi Zhang, Fan Wang, Lidong Bing, and Deli Zhao. 2025.
\newblock \href {https://arxiv.org/abs/2501.13106} {Videollama 3: Frontier multimodal foundation models for image and video understanding}.
\newblock \emph{arXiv preprint arXiv:2501.13106}.

\bibitem[{Zhang et~al.(2024{\natexlab{a}})Zhang, Zhang, Li, Zeng, Yang, Zhang, Wang, Tan, Li, and Liu}]{zhang2024longva}
Peiyuan Zhang, Kaichen Zhang, Bo~Li, Guangtao Zeng, Jingkang Yang, Yuanhan Zhang, Ziyue Wang, Haoran Tan, Chunyuan Li, and Ziwei Liu. 2024{\natexlab{a}}.
\newblock \href {https://arxiv.org/abs/2406.16852} {Long context transfer from language to vision}.
\newblock \emph{arXiv preprint arXiv:2406.16852}.

\bibitem[{Zhang et~al.(2024{\natexlab{b}})Zhang, Wu, Li, Li, Ma, Liu, and Li}]{zhang2024videoinstructiontuningsynthetic}
Yuanhan Zhang, Jinming Wu, Wei Li, Bo~Li, Zejun Ma, Ziwei Liu, and Chunyuan Li. 2024{\natexlab{b}}.
\newblock \href {https://arxiv.org/abs/2410.02713} {Video instruction tuning with synthetic data}.
\newblock \emph{Preprint}, arXiv:2410.02713.

\bibitem[{Zhou et~al.(2024)Zhou, Shu, Zhao, Wu, Xiao, Yang, Xiong, Zhang, Huang, and Liu}]{MLVU}
Junjie Zhou, Yan Shu, Bo~Zhao, Boya Wu, Shitao Xiao, Xi~Yang, Yongping Xiong, Bo~Zhang, Tiejun Huang, and Zheng Liu. 2024.
\newblock Mlvu: A comprehensive benchmark for multi-task long video understanding.
\newblock \emph{arXiv preprint arXiv:2406.04264}.

\end{thebibliography}
% Custom bibliography entries only
% \bibliography{custom}

\appendix

\begin{figure*}[t]
    \centering
    \begin{subfigure}[b]{0.44\textwidth}
        \centering
        \includegraphics[width=\textwidth]{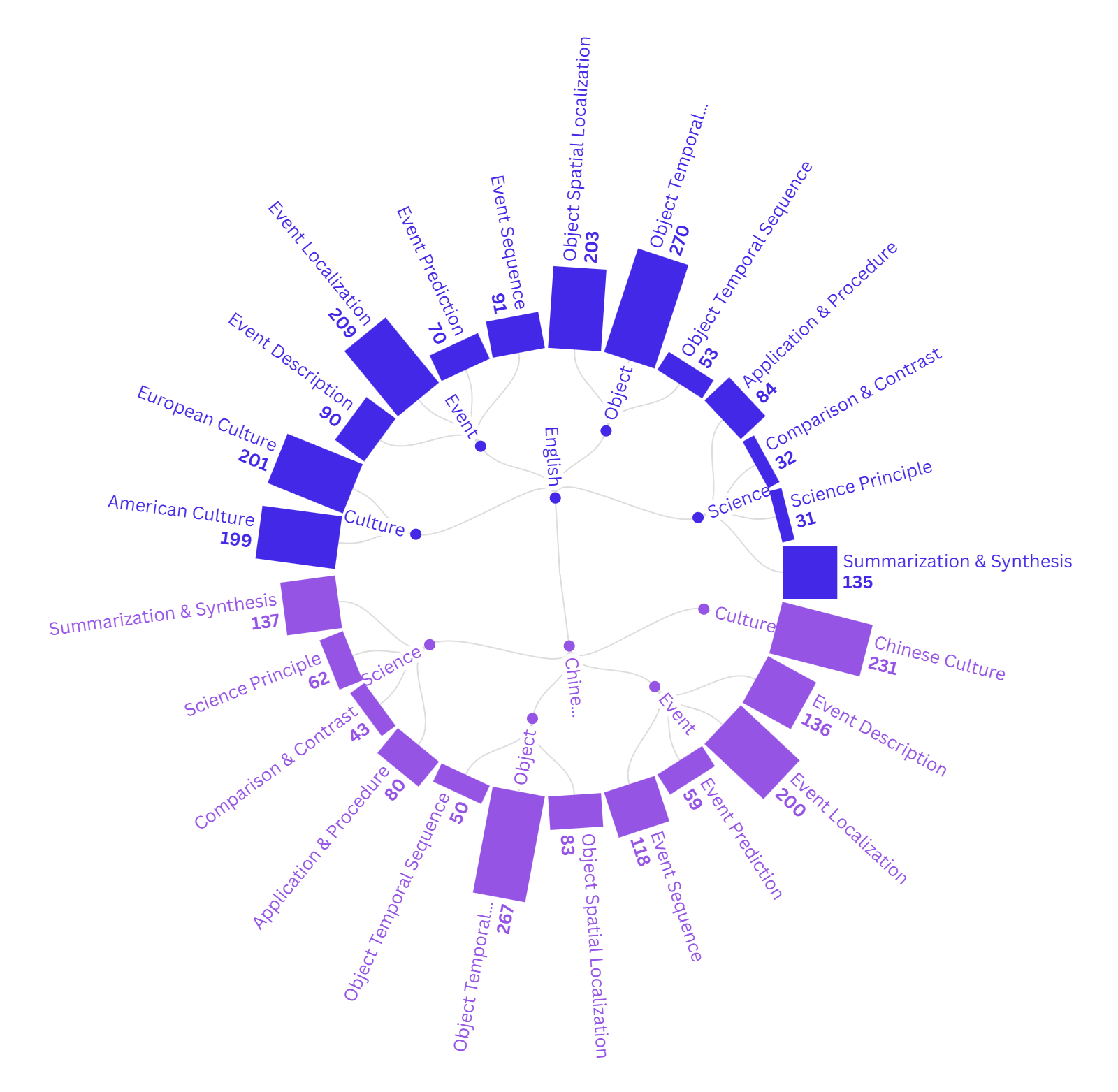}
        \caption{The statistics of 14 subtasks divided by languages.}
        
        \label{fig:sub4}
    \end{subfigure}
    \hfill
    \begin{subfigure}[b]{0.44\textwidth}
        \centering
        \includegraphics[width=\textwidth]{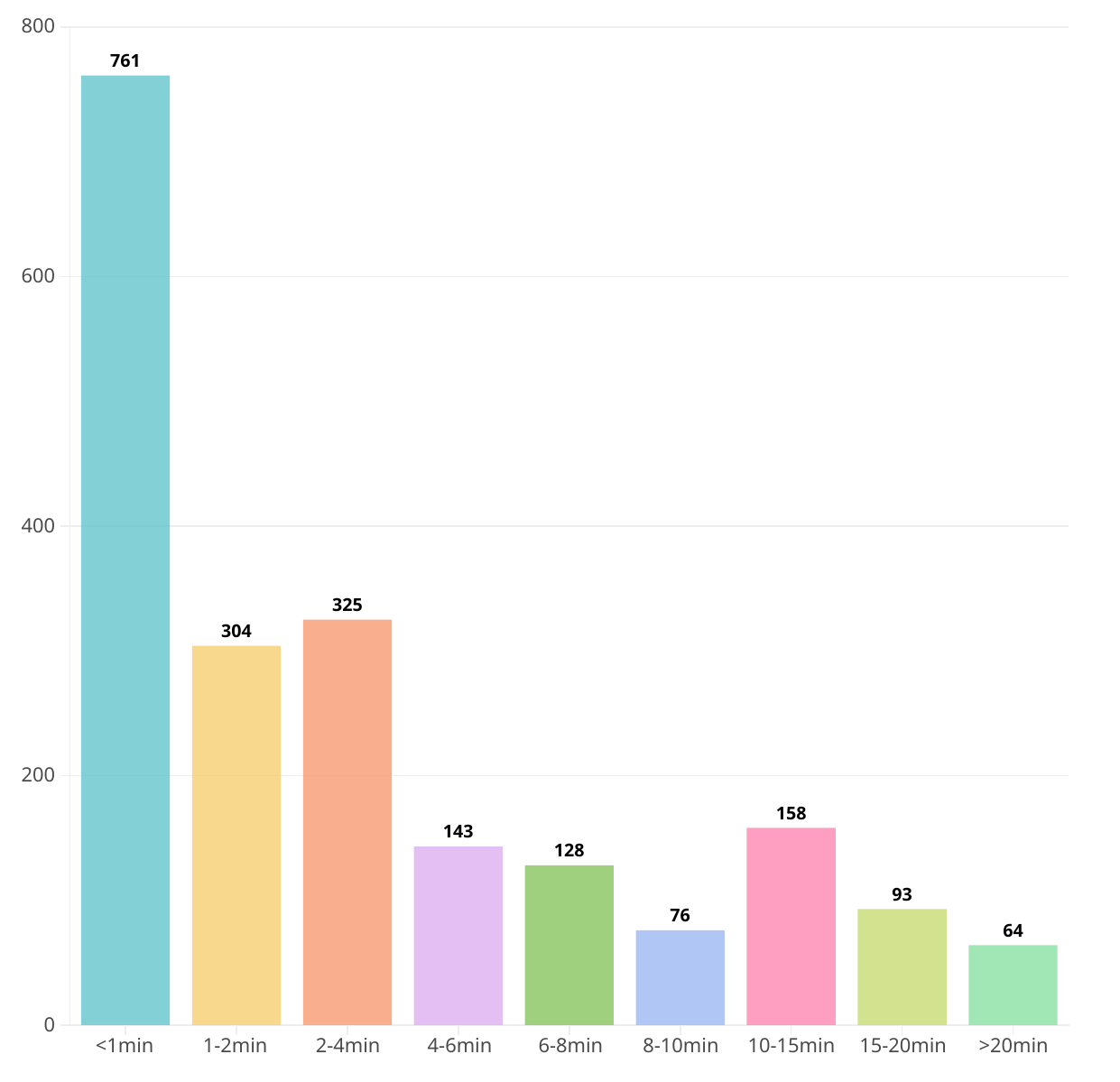}
        \caption{The statistics of duration of videos in VideoVista-CulturalLingo.}
        \label{fig:sub5}
    \end{subfigure}
    \caption{(a) shows the quantity statistics for the 14 task categories under both Chinese and English languages. (b) presents the duration statistics of all video clips in VideoVista-CulturalLingo, measured in minutes.}
    \label{fig:category_and_duration}
\end{figure*}

\section{Additional Dataset Statistics}
% \subsection{Number of each task}
\subsection{Further Statistics}
In Figure~\ref{fig:sub4}, we present the statistics for all task categories in VideoVista-CulturalLingo. In VideoVista-CulturalLingo, the number of English questions is slightly higher than that of Chinese questions, with an additional 222 English questions. The task type with the fewest questions in the dataset is "Comparison \& Contrast", with a total of only 75 questions, while the task type with the most questions is "Object Temporal Localization," with a total of 537 questions. Figure~\ref{fig:sub5} (b) shows the temporal distribution of video clips. Due to the fine-grained object recognition task, the selected videos are often short segments of longer videos, resulting in a larger proportion of videos that are under one minute in length in the dataset. However, VideoVista-CulturalLingo still contains 315 videos longer than 10 minutes, with these long videos primarily concentrated in the Event and Science task categories.

\begin{table}[!t]
\renewcommand\arraystretch{1.1}
\centering
\small
\begin{tabular}{l|c}
\hline
\textbf{Full Name} & \textbf{Abbreviation} \\
\hline
\rowcolor[HTML]{F8E3B5} \multicolumn{2}{c}{\textit{YouTube Domains}} \\
\hline
News \& Politics & \textbf{NP} \\
Sports & \textbf{Spt} \\
Entertainment & \textbf{Ent} \\
Howto \& Style & \textbf{HS} \\
People \& Blogs & \textbf{PB} \\
Autos \& Vehicles & \textbf{AV} \\
Education & \textbf{Edu} \\
Travel \& Events & \textbf{TE} \\
Film \& Animation & \textbf{FA} \\
Comedy & \textbf{Com} \\
Chemical Experiments & \textbf{CE} \\
Science \& Technology & \textbf{ST} \\
Artificial Intelligence & \textbf{AI} \\
Physics Experiment & \textbf{PE} \\
Pets \& Animals & \textbf{PA} \\
Quantum Mechanics & \textbf{QM} \\
Calculus & \textbf{Cal} \\
Linear Algebra & \textbf{LA} \\
\hline
\rowcolor[HTML]{A9D0D5} \multicolumn{2}{c}{\textit{Xiaohongshu Domains}} \\
\hline
Travel Scenery & \textbf{TS} \\
Cooking Process & \textbf{CP} \\
Cooking Tutorial & \textbf{CT} \\
Entrepreneurship & \textbf{Ent} \\
TV Series Commentary & \textbf{TSC} \\
Tourist Attractions & \textbf{TA} \\
Food Review & \textbf{FR} \\
Food Exploration & \textbf{FE} \\
Food Curiosities & \textbf{FC} \\
Astronomy Knowledge & \textbf{AK} \\
Art Explanation & \textbf{AE} \\
Historical Gossip & \textbf{HG} \\
Product Information & \textbf{PI} \\
Travel Vlog & \textbf{TV} \\
Fashion Trends & \textbf{FT} \\
Travel Guide & \textbf{TG} \\
Civil Service Exam Preparation & \textbf{CSEP} \\
Relationship Issues & \textbf{RI} \\
\hline
\rowcolor[HTML]{C8E0D8} \multicolumn{2}{c}{\textit{BiliBili Domains}} \\
\hline
Organic Chemistry & \textbf{OC} \\
Advanced Mathematics & \textbf{AM} \\
High School Experiments & \textbf{HSE} \\
Mid School Experiments & \textbf{MSE} \\
University Physics & \textbf{UP} \\
Machine Learning & \textbf{ML} \\
Deep Learning & \textbf{DL} \\
Quantum Mechanics & \textbf{QM} \\
\hline
\end{tabular}
\caption{\textbf{Abbreviations of domains from different video websites in Figure~\ref{fig:results_domain.}.} The Chinese domains have been translated into English using GPT-4o.}
\label{tab:domain_abbreviations}
\end{table}

\subsection{Abbreviations of Domains}
\label{domain_abbreviations}
We provided the abbreviations of domains in Figure~\ref{fig:results_domain.} in Table~\ref{tab:domain_abbreviations}.

% \paragraph{Open-source Video LMMs.} 

%  We evaluated the newly released Qwen2.5-VL~\cite{qwen2.5-VL}, VideoLLaMA3~\cite{damonlpsg2025videollama3}, InternVideo2.5~\cite{wang2025internvideo}, and TPO~\cite{li2025temporalpreferenceoptimizationlongform} from 2025. Additionally, we evaluated several popular video-capable LMMs introduced in the past two years, including InternVL2.5~\cite{chen2024expanding}, LLaVA-Video~\cite{zhang2024videoinstructiontuningsynthetic},  mPLUG-Owl3~\cite{ye2024mplugowl3longimagesequenceunderstanding}, and others. 

% \paragraph{Open-source Image LMMs.} 

% We also evaluated 3 open-source image LMMs on our benchmarks, including VILA 1.5~\cite{lin2023vila}, DeepSeek2-VL~\cite{wu2024deepseekvl2mixtureofexpertsvisionlanguagemodels}, and Molmo~\cite{molmo2024}. For open-source image LMMs, we employed two video input methods: uniform sampling of 1 frame and uniform sampling of 8 frames.

% \paragraph{Proprietary LMMs.}
% For proprietary LMMs, we evaluated the newly released Gemini 2.0-Flash and Gemini 2.0-Flash-Lite in February, which are currently the workhorse models of the Google Gemini series. Additionally, we conducted evaluations on other prominent proprietary LMMs, including GPT-4o and Gemini 1.5-Flash.

\section{Detailed Experiment Setting}
\label{sec:detailed experiment setting}
\subsection{Open-source Video LMMs}
We evaluated the newly released Qwen2.5-VL~\cite{qwen2.5-VL}, VideoLLaMA3~\cite{damonlpsg2025videollama3}, InternVideo2.5~\cite{wang2025internvideo}, and TPO~\cite{li2025temporalpreferenceoptimizationlongform} from 2025. Additionally, we evaluated several popular video-capable LMMs introduced in the past two years, including InternVL2.5~\cite{chen2024expanding}, LLaVA-Video~\cite{zhang2024videoinstructiontuningsynthetic},  mPLUG-Owl3~\cite{ye2024mplugowl3longimagesequenceunderstanding}, and others. 

In evaluating open-source video LMMs, we use the default hyperparameters specified in their respective open-source implementations for inference. The temperature is generally set to 0 or 0.2, num\_beamsis set to 1, do\_sampleis set to False, and top\_pis set to 1.0. The frame sampling methods for different video models are provided in the Table~\ref{tab:main_result}. Specifically, for the Qwen2.5-VL and Qwen2-VL models, we set the maximum resolution per frame to 224x224 to avoid excessively long sequence lengths.

\subsection{Open-source Image LMMs}

\begin{figure*}[t]
    \centering
    \includegraphics[width=0.8\textwidth]{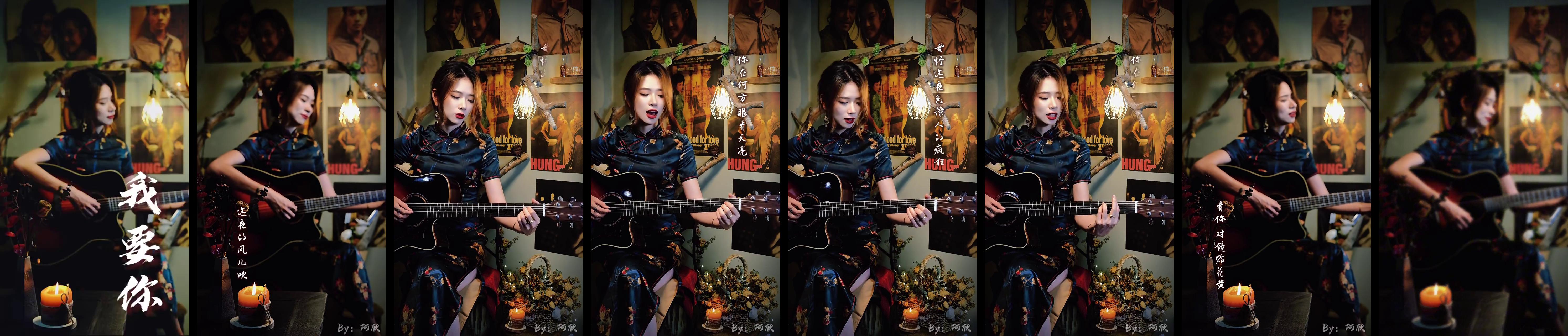}
    \caption{\textbf{An example of eight images combined in a horizontal layout.}
    }
    \label{fig:example_merge}
\end{figure*}

We also evaluated three open-source image LMMs on our benchmarks, including VILA 1.5~\cite{lin2023vila}, DeepSeek2-VL~\cite{wu2024deepseekvl2mixtureofexpertsvisionlanguagemodels}, and Molmo~\cite{molmo2024}. For open-source image LMMs, we employed two video input methods: uniform sampling of 1 frame and uniform sampling of 8 frames.

In evaluating these open-source image LMMs, we also adopted the hyperparameter settings provided in the implementations for inference. Regardless of whether single-frame or eight-frame input is used for evaluation, all images are presented at their original resolution without compression. Specifically, due to an error in the official code of the Molmo model when inputting eight images simultaneously, we concatenated the eight images horizontally into a single image and noted this in the prompt. An example of this image is Figure~\ref{fig:example_merge}.

\subsection{Proprietary LMMs}
For proprietary LMMs, we evaluated the newly released Gemini 2.0-Flash and Gemini 2.0-Flash-Lite in February, which are currently the workhorse models of the Google Gemini series. Additionally, we conducted evaluations on other prominent proprietary LMMs, including GPT-4o and Gemini 1.5-Flash.

In evaluating proprietary LMMs, we optimize API resource usage and accelerate the evaluation process by input multiple questions for each video. Thanks to the powerful instruction-following capability of Proprietary LMMs, they are able to return a dictionary in the format of \{"question id": "prediction"\} accurately. Although this may introduce some evaluation bias, Proprietary LMMs still demonstrated exceptional performance on our benchmark. Additionally, when evaluating the GPT-4 model, we compressed all video frames to a resolution of 512x512 for input.

\section{Further Experiments}
\label{sec:appendix_exp_results}

\begin{figure*}[!t]
    \centering
    \begin{subfigure}[b]{0.45\textwidth}
        \centering
        \includegraphics[width=\textwidth]{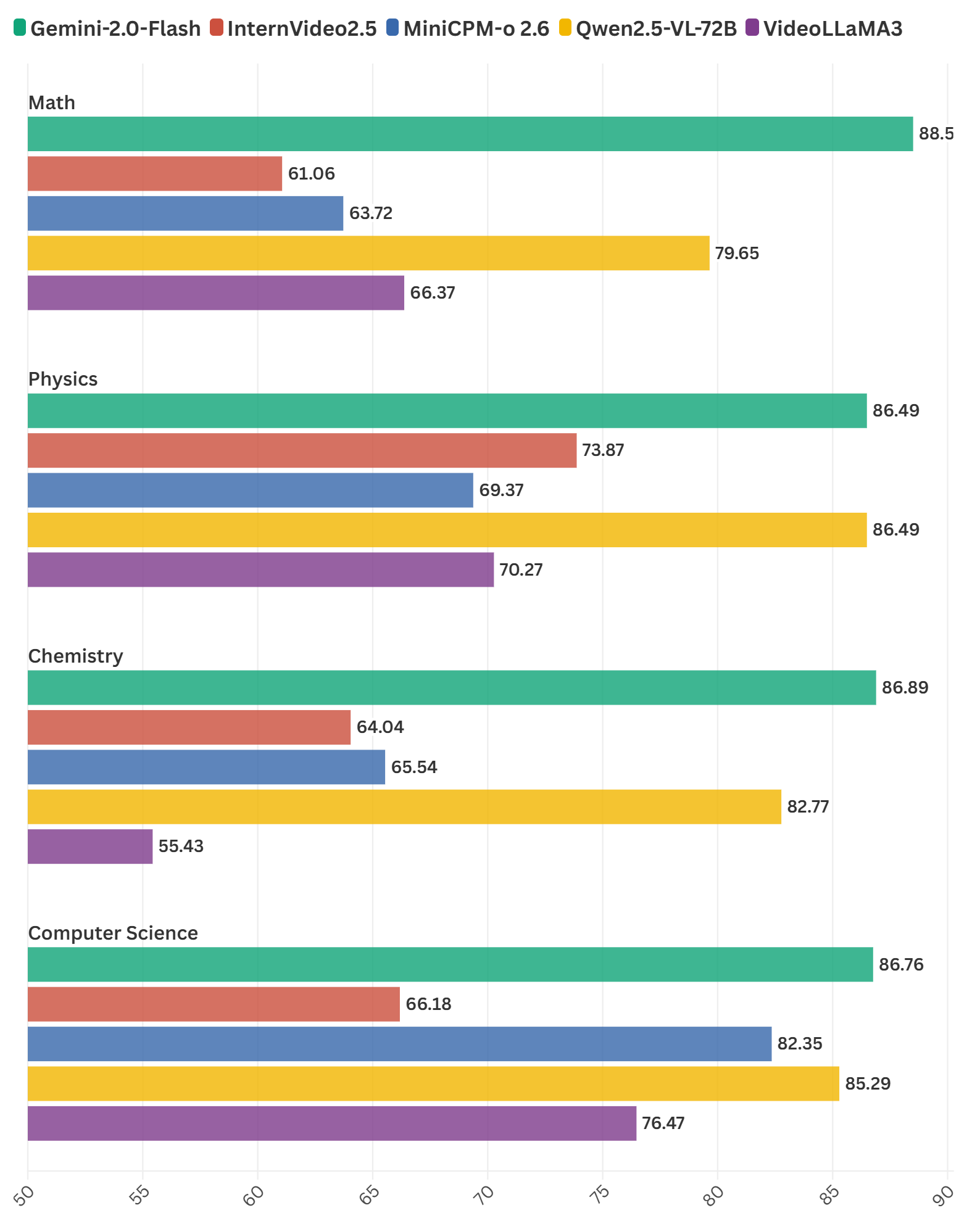}
        \caption{Science-based Evaluation Results.}
        \label{fig:science}
    \end{subfigure}
    \hfill
    \begin{subfigure}[b]{0.45\textwidth}
        \centering
        \includegraphics[width=\textwidth]{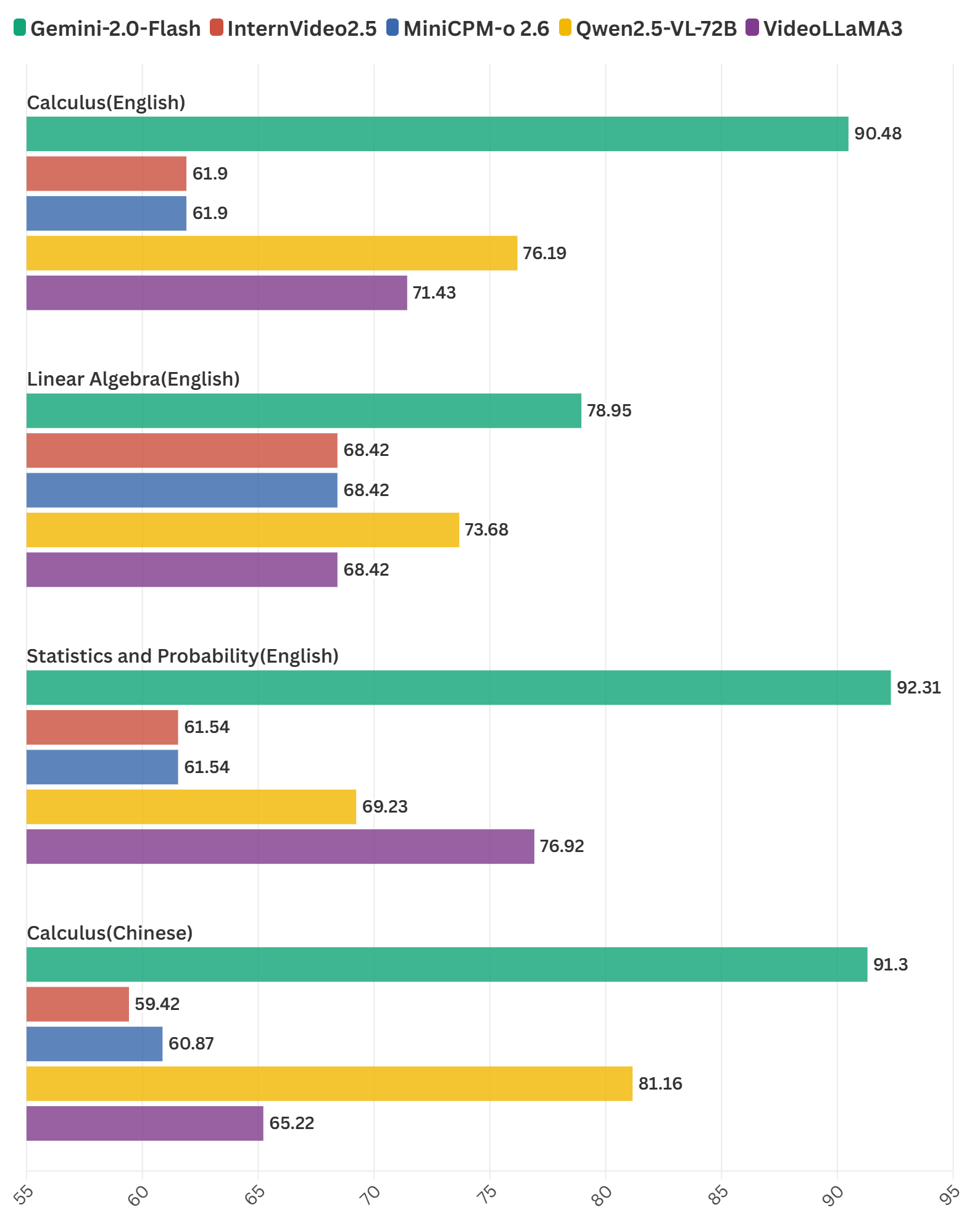}
        \caption{Math-based Evaluation Results.}
        \label{fig:math}
    \end{subfigure}
    \caption{\textbf{The Evaluation results in 4 disciplines and 4 math sub-disciplines.} The experimental results in the figure represent the average values of the four scientific sub-tasks. In (a),we have list the four disciplines covered by the scientific videos in VideoVista-CulturalLingo: Math, Physics, Chemistry, and Computer Science ; In (b), we have listed four math sub-disciplines with a larger number of questions: Calculus (English), Linear Algebra (English), Statistics and Probability (English), and Calculus (Chinese)/Advanced Mathematics. }
    \label{fig:science_and_math}
\end{figure*}

\subsection{Model Performance in Science}
\label{sec:model_performance_science}
For the third finding discussed in the abstract, we present detailed experimental results in Figure~\ref{fig:science_and_math}. We present a performance comparison between the four best-performing open-source video LMMs and the strongest proprietary model, Gemini-2.0-Flash. As shown in Figure~\ref{fig:science}, the primary performance gap between open-source Video LMMs and proprietary LMMs in scientific tasks is observed in the Mathematics disciplines. Specifically, for Physics, Chemistry, and Computer Science questions, the top-performing open-source Video LMM, Qwen2.5-VL-72B, exhibits a performance gap of less than 5\% compared to Gemini-2.0-Flash. However, for Math questions, the gap between the two models increases to nearly 10\%.

In Figure~\ref{fig:math}, we further compare the performance differences of various models across specific math sub-disciplines. It is evident that, regardless of whether the questions are in Chinese or English, existing open-source video LMMs still exhibit a performance gap when compared to the proprietary LMM Gemini-2.0-Flash.  The largest gaps are observed in the Calculus (English) and Statistics and Probability (English) categories, where the leading open-source video LMMs show a performance difference exceeding 10\% compared to Gemini-2.0-Flash.

\begin{figure}[t]
    \centering
    \includegraphics[width=0.5\textwidth]{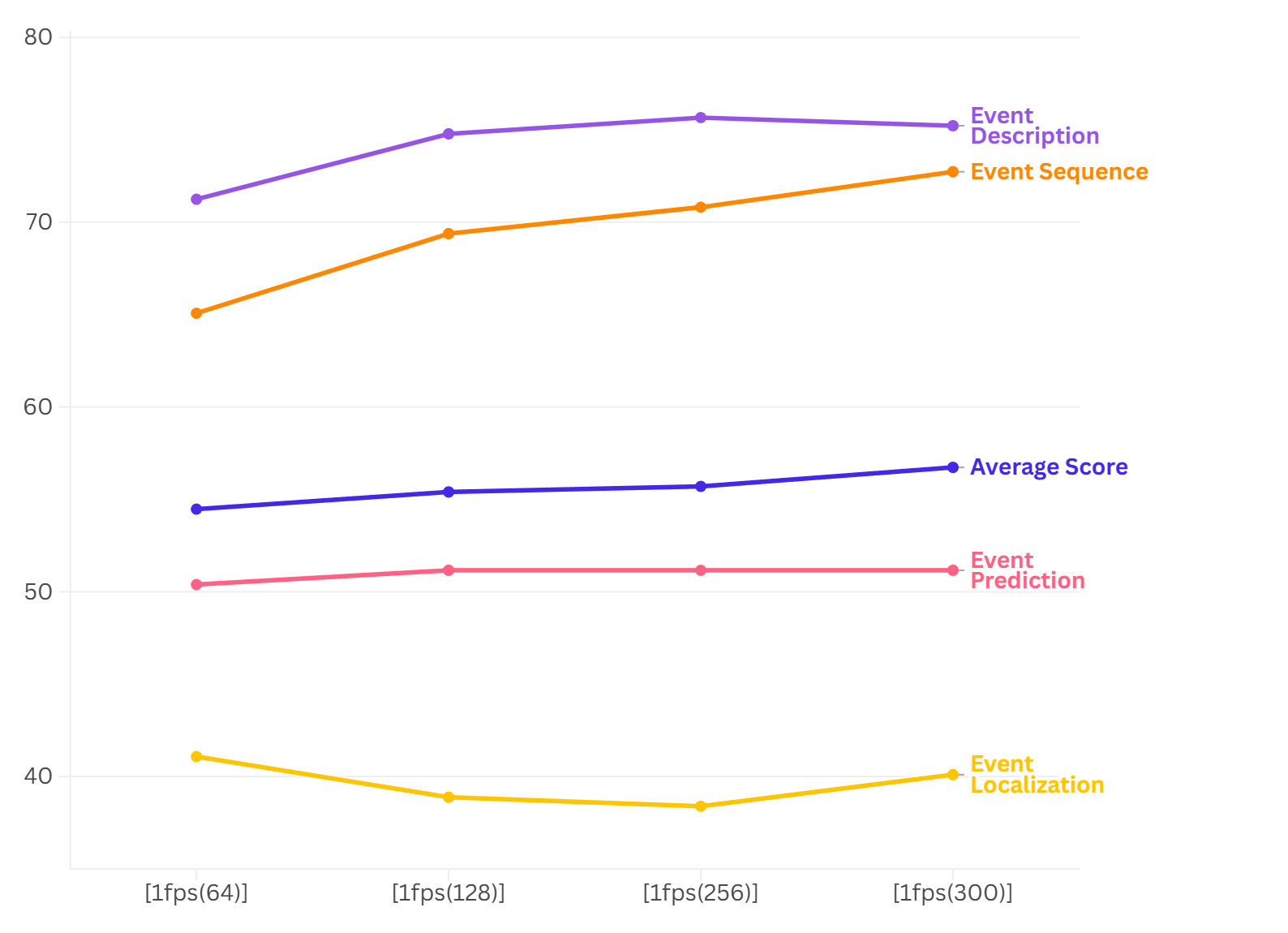}
    \caption{\textbf{The Evaluation results divided by frames upper bound of Qwen2.5-VL-7B.} We conducted experiments with four sampling methods at frame upper bound of 64, 128, 256, and 300 frames.
    }
    \label{fig:evaluation_frames}
\end{figure}

\subsection{Impact of Frame Sampling}
 We conduct an experiment to evaluate the frame sampling upper bound for event task questions using the Qwen2.5-VL-7B model, and the results are shown in the Figure~\ref{fig:evaluation_frames}. It can be observed that as the frame sampling upper upper bound increases, the overall evaluation performance of the model gradually improves. However, there is no significant leap in performance, which could be due to the fact that our final frame sampling upper limit of 300 is still not high enough.

\begin{figure}[t]
    \centering
    \includegraphics[width=0.5\textwidth]{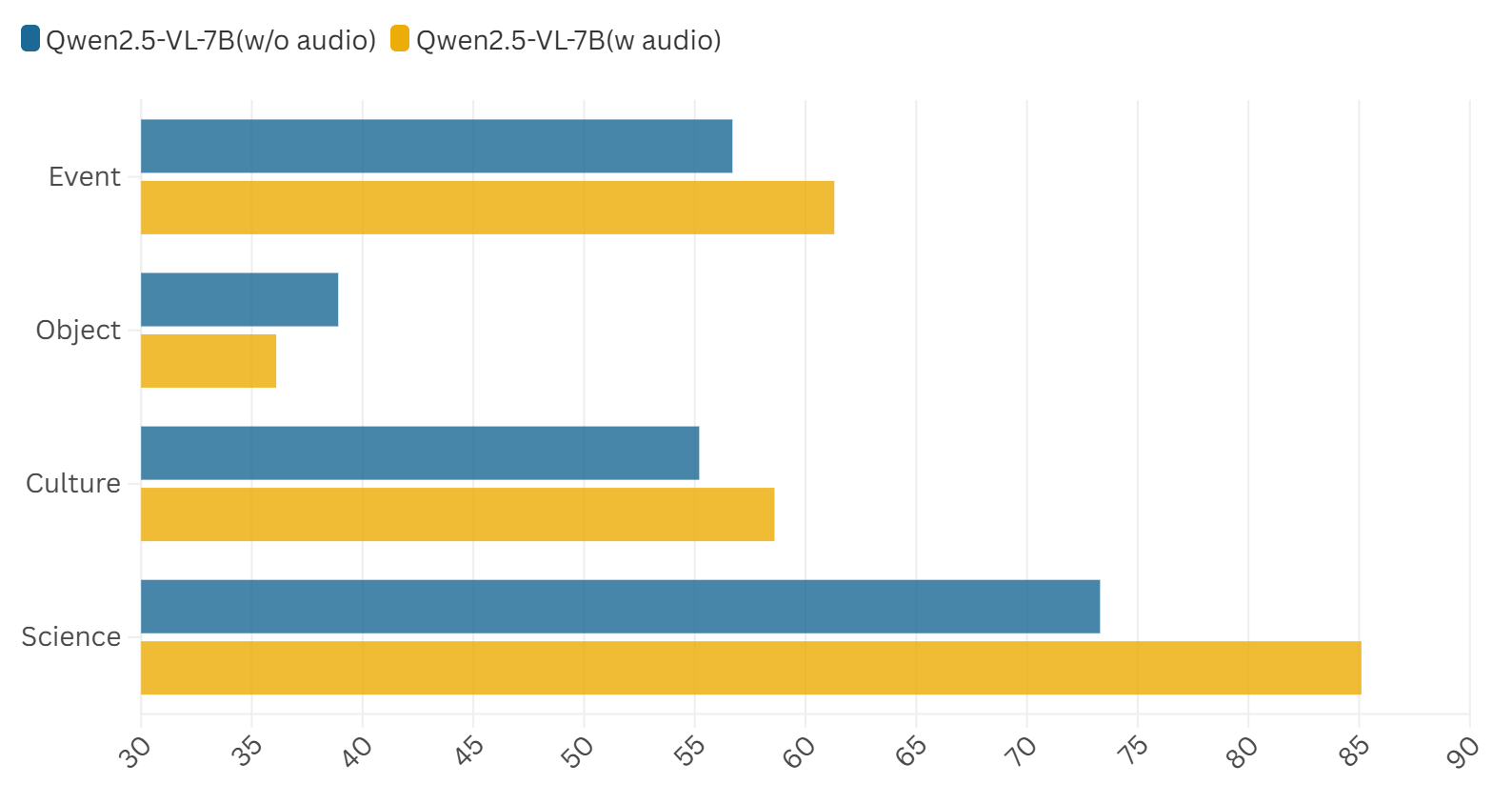}
    \caption{\textbf{The Evaluation results divided by whether input audio transcript into Qwen2.5-VL-7B.} The audio transcript is extracted using Whisper-Large-V3.
    }
    \label{fig:evaluation_audio}
\end{figure}

\subsection{Impact of Audio Information}
We also conduct experiments using the Qwen2.5-VL-7B model to investigate the impact of adding audio transcripts in VideoVista-CulturalLingo, with the experimental results are shown in the Figure~\ref{fig:evaluation_audio}. The input audio transcript is the unrefined version extracted from Whisper-Large-V3. It can be observed that incorporating additional information from the audio modality, the model's performance improves in the tasks of Event, Culture, and Science. In the Science task, the improvement in model performance is most significant. This is likely because the audio in the science videos we selected is generally clear and explicit, covering the experimental and course-related information. However, in the Event and Culture tasks, the inclusion of audio transcripts only resulted in a small improvement. We encourage LMMs to process both audio and video frames simultaneously, and therefore, we did not include the audio information in our model evaluation.

\begin{figure}[t]
    \centering
    \includegraphics[width=0.5\textwidth]{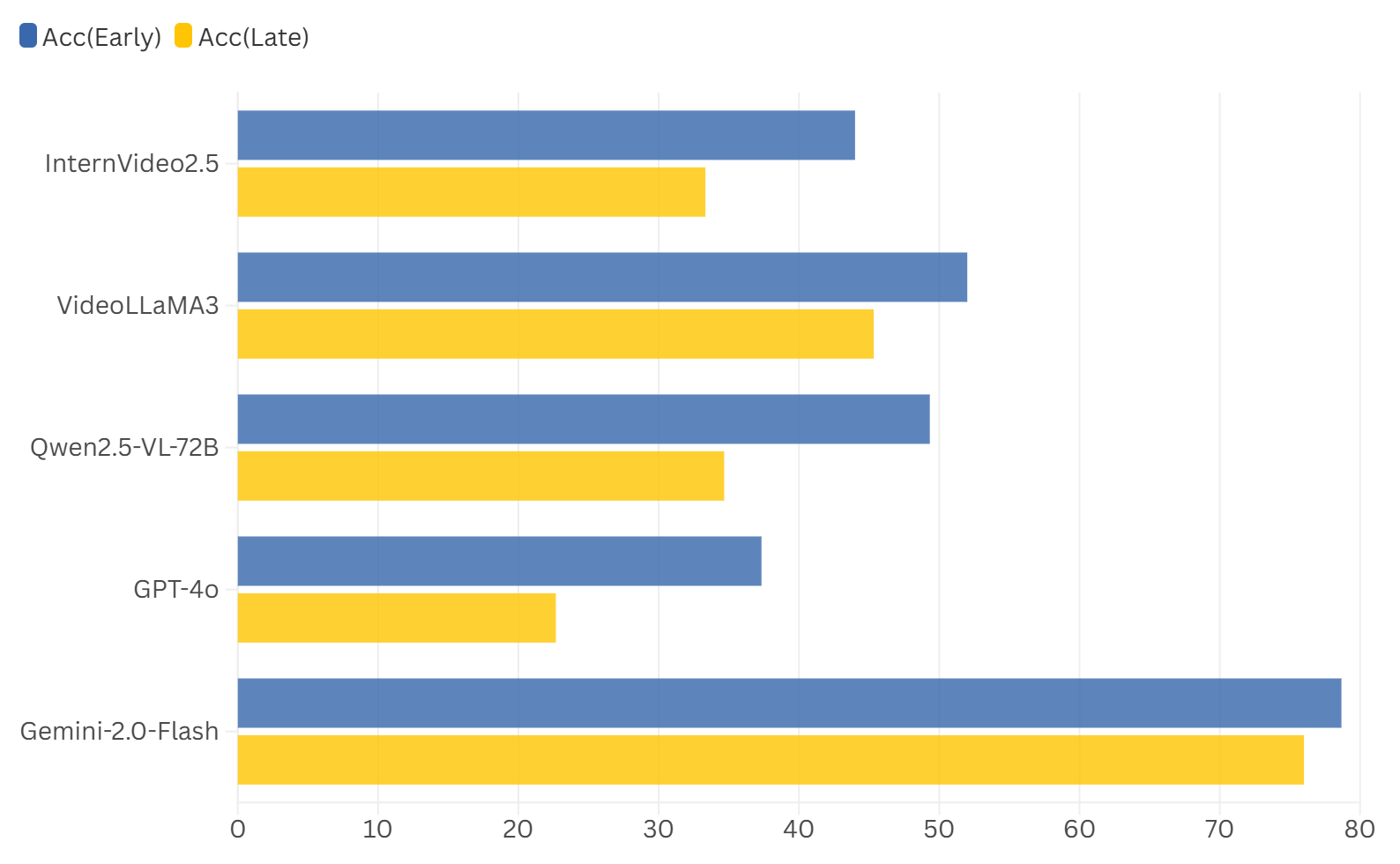}
    \caption{\textbf{The Evaluation results of the Temporal Relationship Between the Event of Interest and the Corresponding Video}
    }
    \label{fig:evaluation_temporal}
\end{figure}

\subsection{Impact of Temporal Relation}
We conduct experiments to investigate the temporal relationship between the event of interest posed in the question and the corresponding video, with the experimental results are shown in the Figure~\ref{fig:evaluation_temporal}. In the Event Localization task, we selected 75 video pairs, where each pair contained videos of similar durations, and the events in the questions occurred in the early (front half) and late (back half) parts of the video. The evaluation results of five major models on early vs. late questions are summarized in the table below. As shown, except for the powerful Gemini-2.0-Flash model, the accuracy for early questions is significantly higher than for late questions across the remaining Video-LMMs. This suggests that most existing Video-LMMs have a stronger understanding of events occurring early in the video but tend to struggle with those that happen later. We guess that the phenomenon is caused by the bias in the model's training data or the model's ability to handle long-contexts.

\begin{table*}[t]
  \renewcommand\arraystretch{1.4}
  \centering
  \scriptsize
    \scalebox{1.}{
    \begin{tabular}{l | c c c c| c c c| c c c|c c c c}
      \hline
      \multirow{2}{*}{\textbf{Model}} & \multicolumn{4}{c|}{Event} & \multicolumn{3}{c|}{Object} & \multicolumn{3}{c|}{Culture} & \multicolumn{4}{c}{Science} \\[1pt]
      \cline{2-15}
      & ED & EP & ES & EL & OTL & OTS & OSL & CC & AC & EC  & SS & COM & AP & SP \\
    \hline
    \rowcolor[HTML]{A9D0D5} \multicolumn{15}{c}{\textit{Open-source Video LMMs}} \\
    \hline
      ShareGPT4Video & 29.2 & 20.2 & 17.7 & 23.7 & 10.4 & 27.2 & 30.8 & 19.0 & 34.7 & 42.3 & 32.4 & 48.0 & 32.9 & 30.1 \\
      VideoChat2-Mistral & 38.5 & 28.7 & 31.1 & 19.3 & 25.1 & 26.2 & 27.6 & 25.1 & 40.2 & 40.3 & 36.4 & 44.0 & 23.8 & 31.2 \\
      Video-LLaVA & 51.3 & 46.5 & 31.1 & 41.6 & 32.2 & 24.3 & 42.3 & 27.7 & 35.2 & 41.8 & 42.6 & 38.7 & 40.9 & 38.7 \\
      VideoLLaMA2 & 36.3 & 28.7 & 41.6 & 29.6 & 17.9 & 15.5 & 36.4 & 25.1 & 38.7 & 42.3 & 36.4 & 42.7 & 35.4 & 34.4 \\
      LLaVA-OneVision & 47.8 & 34.9 & 44.0 & 44.5 & 30.7 & 35.0 & 39.2 & 36.4 & 41.7 & 38.8 & 55.1 & 44.0 & 57.9 & 48.4 \\
      MiniCPM-V 2.6 & 74.3 & 38.0 & 41.1 & 30.8 & 18.8 & 35.0 & 30.1 & 44.6 & 48.7 & 55.7 & 70.6 & 53.3 & 60.4 & 52.7 \\
      mPLUG-Owl3 & 66.4 & 56.6 & 52.2 & 48.2 & 35.3 & 61.2 & 41.7 & 37.7 & 45.7 & 52.7 & 62.1 & 58.7 & 60.4 & 54.8 \\
      Oryx-1.5 & 54.4 & 40.3 & 45.9 & 37.7 & 33.1 & 24.3 & 33.2 & 35.5 & 39.2 & 38.3 & 58.5 & 46.7 & 57.9 & 51.6 \\
      LLaVA-Video & 75.7 & 57.4 & 48.3 & 53.1 & 33.7 & 67.0 & 39.2 & 41.6 & 51.3 & 54.7 & 63.6 & 53.3 & 61.0 & 52.7 \\
      Qwen2-VL & 72.6 & 51.2 & 56.9 & 33.3 & 30.0 & 47.6 & 36.0 & 48.5 & 54.8 & 62.2 & 72.1 & 60.0 & 66.5 & 65.6 \\
      InternVL2.5 & 81.4 & 57.4 & 59.3 & 41.1 & 35.9 & 47.8 & 30.4 & 55.4 & 47.7 & 65.2 & 69.8 & 56.0 & 65.2 & 62.4 \\
      MiniCPM-o 2.6 & 83.6 & 55.0 & 53.1 & 35.2 & 20.1 & 52.4 & 35.7 & 48.9 & 56.3 & 63.7 & 72.1 & 61.3 & 69.5 & 52.7 \\
      TPO & 75.2 & 56.6 & 49.8 & 48.2 & 31.2 & 67.0 & 38.8 & 43.7 & 50.8 & 55.2 & 63.2 & 50.7 & 62.8 & 55.9 \\
      InternVideo2.5 & 80.5 & 52.7 & 60.3 & 33.0 & 37.1 & 61.2 & 31.8 & 53.7 & 56.3 & 65.2 & 72.1 & 61.3 & 64.0 & 54.8 \\
      VideoLLaMA3 & 77.9 & 57.4 & 61.7 & 45.2 & 72.1 & 64.1 & 56.6 & 45.5 & 55.8 & 59.2 & 70.2 & 54.7 & 64.0 & 55.9 \\
      Qwen2.5-VL-7B & 75.2 & 51.2 & 72.7 & 40.1 & 39.3 & 56.3 & 31.8 & 51.9 & 50.8 & 63.2 & 80.5 & 65.3 & 72.6 & 60.2 \\
      Qwen2.5-VL-72B & 79.2 & 60.5 & 78.9 & 42.1 & 31.5 & 67.0 & 49.7 & 65.8 & 67.8 & 80.6 & 86.4 & 85.3 & 79.3 & 79.6 \\
    \hline
    \rowcolor[HTML]{D4E3EC} \multicolumn{15}{c}{\textit{Open-source Image LMMs}} \\
    \hline
    VILA1.5-13B[1f] & 33.3 & 33.3 & 29.2 & 33.9 & 26.8 & 26.2 & 34.6 & 31.6 & 30.7 & 39.8 & 36.8 & 46.7 & 39.6 & 39.8\\
    VILA1.5-13B[8f] & 36.9 & 38.2 & 31.3 & 38.2 & 23.1 & 35.9 & 45.1 & 23.4 & 42.2 & 51.2 & 43.4 & 41.3 & 40.9 & 39.8\\
    Molmo 7B-D[1f] & 38.3 & 44.5 & 25.3 & 39.8 & 26.6 & 34.0 & 19.6 & 39.0 & 40.7 & 39.8 & 46.3 & 41.3 & 50.0 & 45.2\\
    Molmo 7B-D[8f] & 40.3 & 44.3 & 30.1 & 41.8 & 29.6 & 45.6 & 25.5 & 37.7 & 44.2 & 44.3 & 50.0 & 42.7 & 49.4 & 44.1\\
    DeepSeek2-VL[1f] & 40.9 & 44.3 & 32.2 & 39.3 & 32.4 & 33.0 & 31.5 & 37.7 & 38.2 & 42.3 & 52.2 & 44.0 & 49.4 & 52.7\\
    DeepSeek2-VL[8f] & 42.6 & 47.0 & 27.2 & 44.4 & 25.0 & 33.0 & 29.4 & 37.2 & 40.7 & 56.2 & 62.9 & 50.7 & 53.0 & 54.8\\
    \hline
    \rowcolor[HTML]{8FB8A1} \multicolumn{15}{c}{\textit{Proprietary LMMs}} \\
    \hline
      GPT-4o & 86.3 & 47.3 & 70.3 & 28.6 & 29.4 & 61.2 & 46.5 & 57.1 & 71.9 & 76.6 & 81.6 & 77.3 & 80.5 & 65.6 \\
      Gemini-1.5-Flash & 92.5 & 42.6 & 63.6 & 69.4 & 87.3 & 69.9 & 23.7  & 49.4 & 61.3 & 67.7 & 87.9 & 87.7 & 82.9 & 77.4 \\
      Gemini-2.0-Flash-Lite & 87.2 & 44.1 & 68.4 & 63.8 & 87.5 & 63.1 & 44.8  & 58.4 & 61.3 & 70.1 & 83.1 & 81.3 & 80.5 & 82.8 \\
      Gemini-2.0-Flash  & 92.9 & 51.9 & 73.7 & 70.7 & 87.2 & 74.8 & 59.1 & 62.3 & 64.8 & 77.6 & 88.2 & 87.8 & 81.7 & 90.7\\
      \hline
    \end{tabular}}
    \caption{\textbf{Detailed Evaluation results on VideoVista-CulturalLingo benchmark.} Abbreviations used in the table:Event Description (\textbf{ED}), Event Prediction (\textbf{EP}), Event Sequence (\textbf{ES}), Event Localization (\textbf{EL}), Object Temporal Localization (\textbf{OTL}), Object Temporal Sequence (\textbf{OTS}), Object Spatial Localization (\textbf{OSL}), Chinese Culture (\textbf{CC}), American Culture (\textbf{AC}), European Culture (\textbf{EC}), Summarization \& Synthesis (\textbf{SS}), Comparison \& Contrast (\textbf{COM}), Application \& Procedure (\textbf{AP}), Scientific Principle (\textbf{SP}).}
    \label{tab:detailed_result_all} 
  \end{table*}

\begin{table}[t]
\renewcommand\arraystretch{1.3}
\centering
\scriptsize
\scalebox{1.3}{
\begin{tabular}{l|c|c}
\hline
\textbf{Model} & \textbf{Chinese} & \textbf{English} \\
\hline
MiniCPM-o 2.6 & 58.77 &  63.49\\
InternVideo2.5 & 60.04 & 63.49\\
VideoLLaMA3 & 52.26 &  63.78  \\
Qwen2.5-VL-72B & 75.59 & 78.30\\
\hline
GPT-4o & 68.35 & 76.83 \\
Gemini-2.0-Flash &  76.49 & 78.30 \\
\hline
\end{tabular}}
\caption{\textbf{Model Performance by Video Language.}}
\label{tab:detailed_result_language} 
\end{table}

\begin{table}[t]
\renewcommand\arraystretch{1.3}
\centering
\scriptsize
\scalebox{1.3}{
\begin{tabular}{l|c|c|c}
\hline
\textbf{Model} & \textbf{Short} & \textbf{Medium} & \textbf{Long} \\
\hline
MiniCPM-o 2.6 & 54.46 & 52.91 & 44.30 \\
InternVideo2.5 & 53.12 & 52.69 & 48.10 \\
VideoLLaMA3 & 61.16 & 56.05 & 50.63 \\
Qwen2.5-VL-72B & 62.72 & 59.64 & 59.49 \\
\hline
GPT-4o & 54.91 & 52.69 & 49.37 \\
Gemini-2.0-Flash & 75.89 & 74.22 & 62.03 \\
\hline
\end{tabular}}
\caption{\textbf{Model Performance by Video Duration.} The Duration: <2 minutes (\textbf{Short}), 2-10 minutes (\textbf{Medium}), >10 minutes (\textbf{Long}).}
\label{tab:detailed_result_duration} 
\end{table}

\subsection{Detailed Experiment Results}
\label{sec:appendix_experiment_subtasks}
In Table~\ref{tab:detailed_result_all}, we provide a detailed presentation of the performance of all evaluated models across 14 sub-tasks. In Tables~\ref{tab:detailed_result_language} and ~\ref{tab:detailed_result_duration}, we present the detailed evaluation results used to plot Figures~\ref{fig:sub2} and ~\ref{fig:sub3}. These evaluation results effectively demonstrate the models' performance across different languages and video durations.

\begin{figure}[t]
    \centering
    \includegraphics[width=0.5\textwidth]{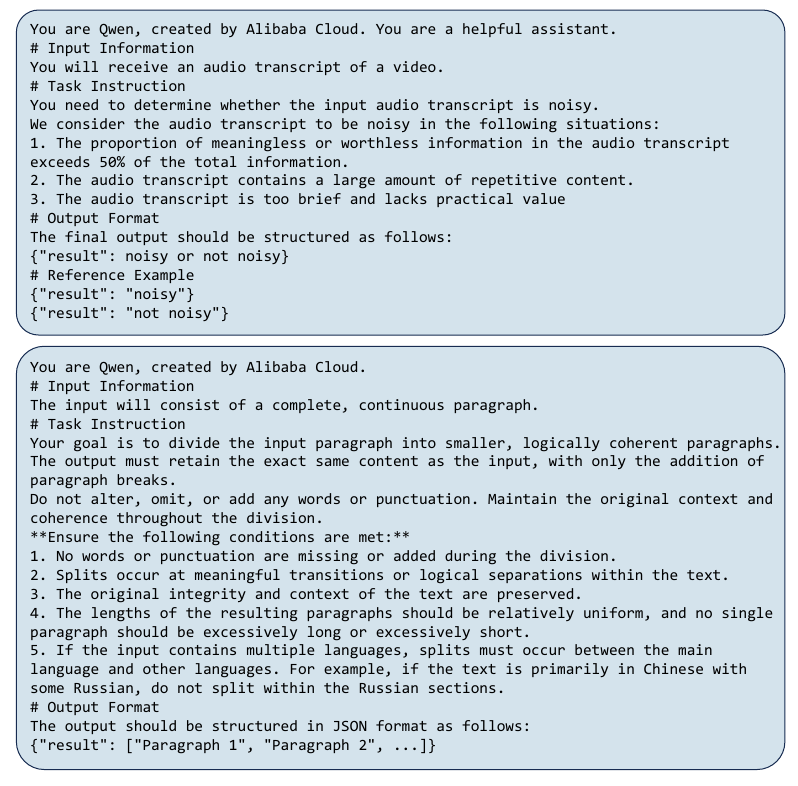}
    \caption{\textbf{Prompt for Video Processing.}
    }
    \label{fig:prompt_video_processing}
\end{figure}

\begin{figure}[t]
    \centering
    \includegraphics[width=0.5\textwidth]{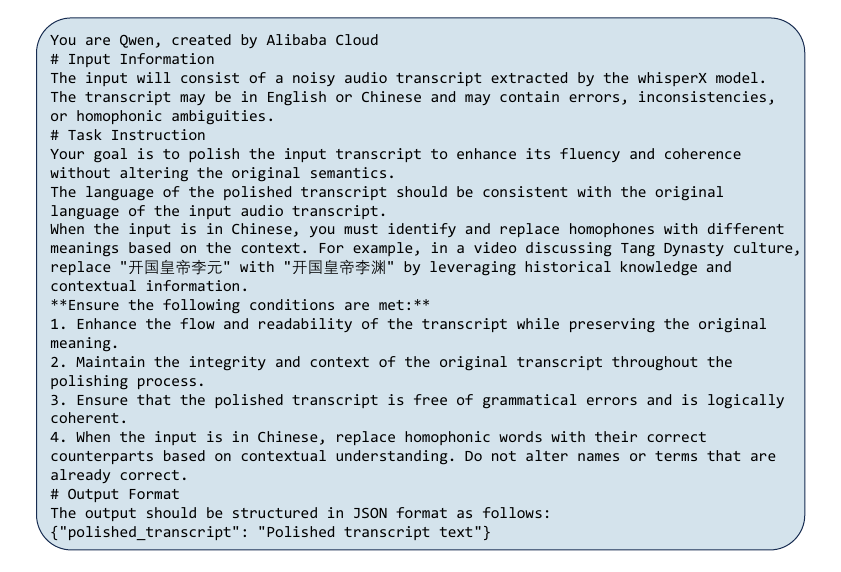}
    \caption{\textbf{Prompt for Audio Refine.}
    }
    \label{fig:prompt_video_processing_2}
\end{figure}

\section{Detailed Annotations Pipeline}
\label{sec:detail_annotations}

\subsection{Prompt for Video Preprocessing}
We introduce the prompt to determine whether the audio of video is noisy above Figure~\ref{fig:prompt_video_processing} and the prompt to split the video based on audio in below of Figure~\ref{fig:prompt_video_processing}. Both two prompt are input to Qwen2.5-72B language model during the video preprocessing stage.

In the Figure~\ref{fig:prompt_video_processing_2}, we present the prompt used to refine the audio transcripts recognized by WhisperX, primarily aimed at eliminating homophones in Chinese, reducing ambiguity, and enhancing fluency. This process is also carried out using the Qwen2.5-72B language model.

\begin{figure*}[t]
    \centering
    \includegraphics[width=0.9\textwidth]{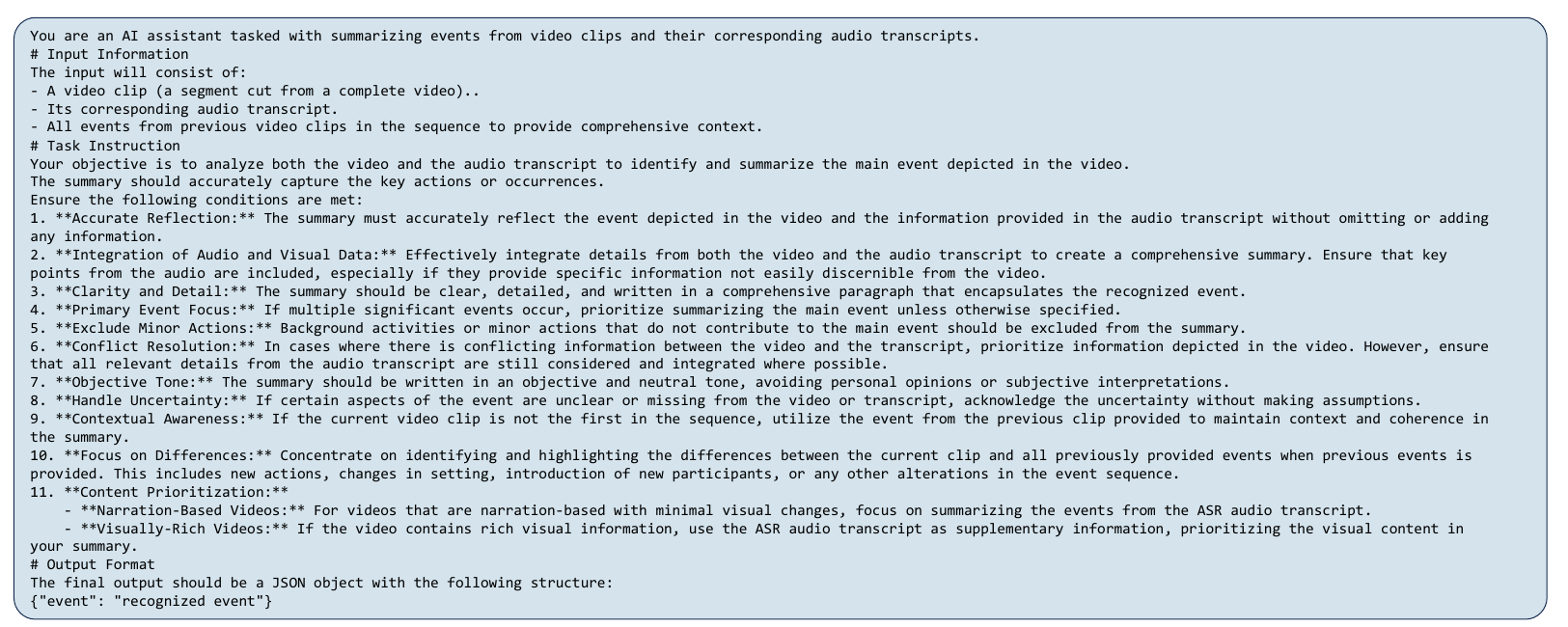}
    \caption{\textbf{Prompt for Event Annotation.}
    }
    \label{fig:prompt_event_annotation}
\end{figure*}

\begin{figure*}[t]
    \centering
    \includegraphics[width=0.9\textwidth]{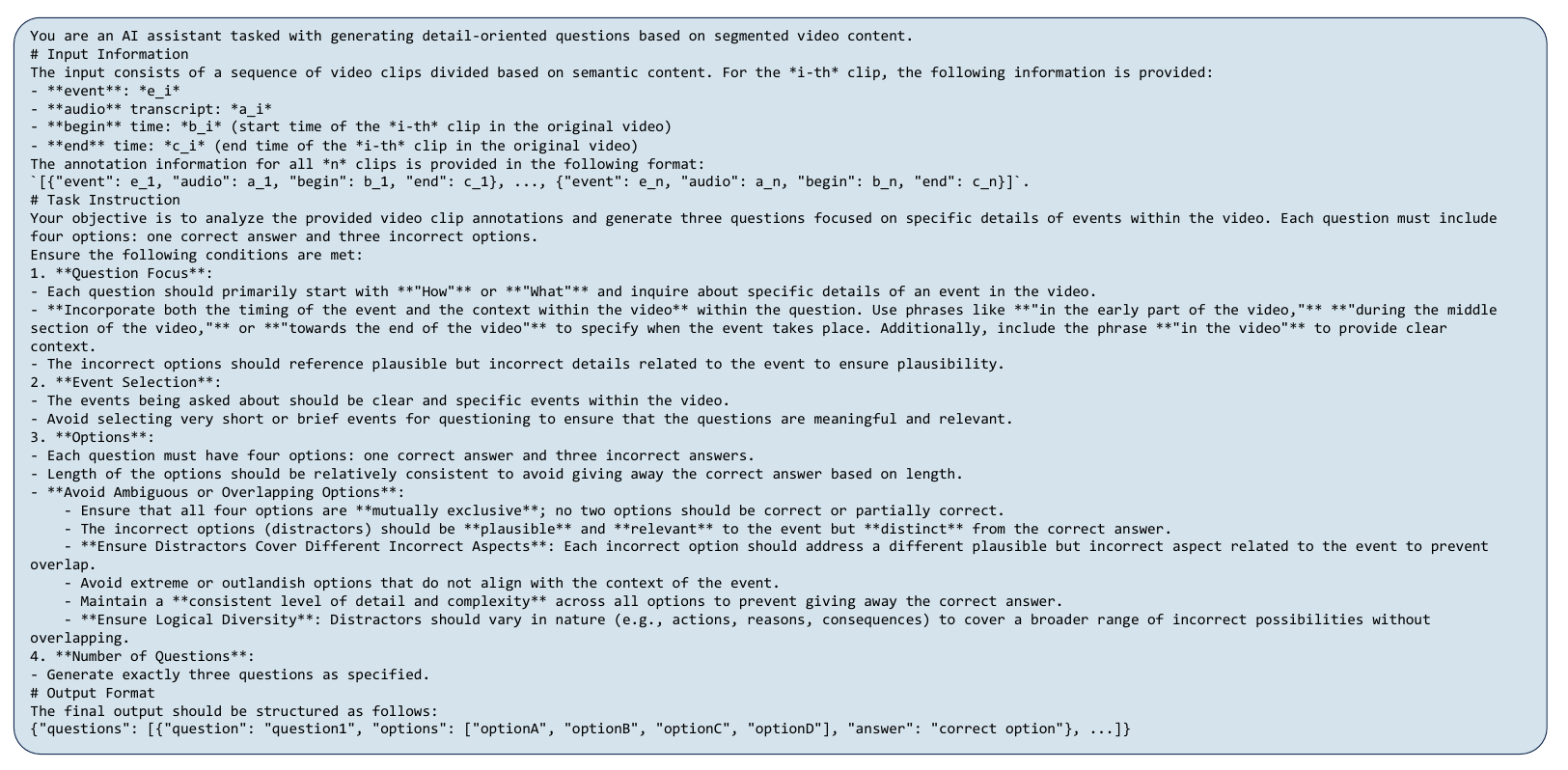}
    \caption{\textbf{Prompt for Event Description Quetions, Options and Answer Generation.}
    }
    \label{fig:prompt_qa_generation_1}
\end{figure*}

\begin{figure*}[t]
    \centering
    \includegraphics[width=0.9\textwidth]{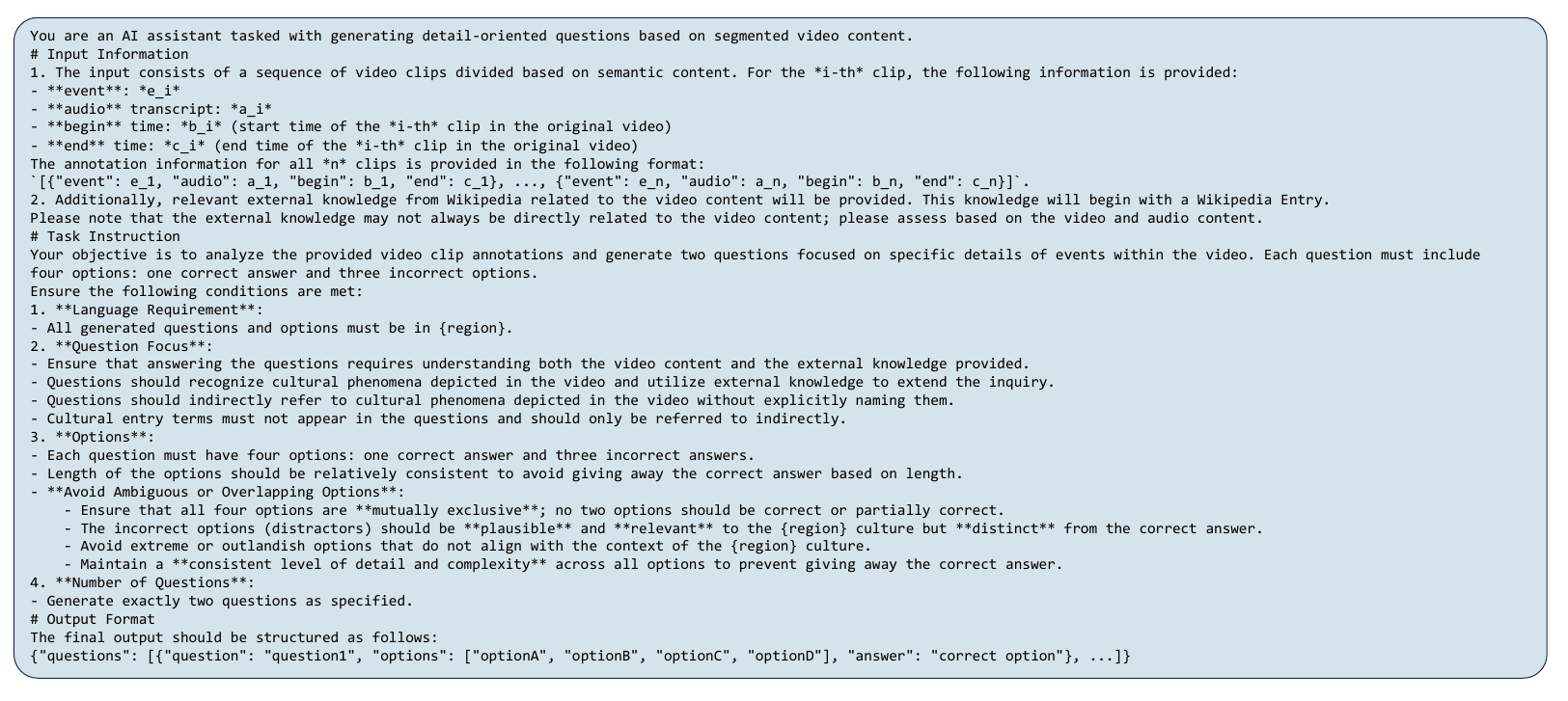}
    \caption{\textbf{Prompt for Chinese Culture Quetions, Options and Answer Generation.}
    }
    \label{fig:prompt_qa_generation_2}
\end{figure*}

\begin{figure*}[t]
    \centering
    \includegraphics[width=0.9\textwidth]{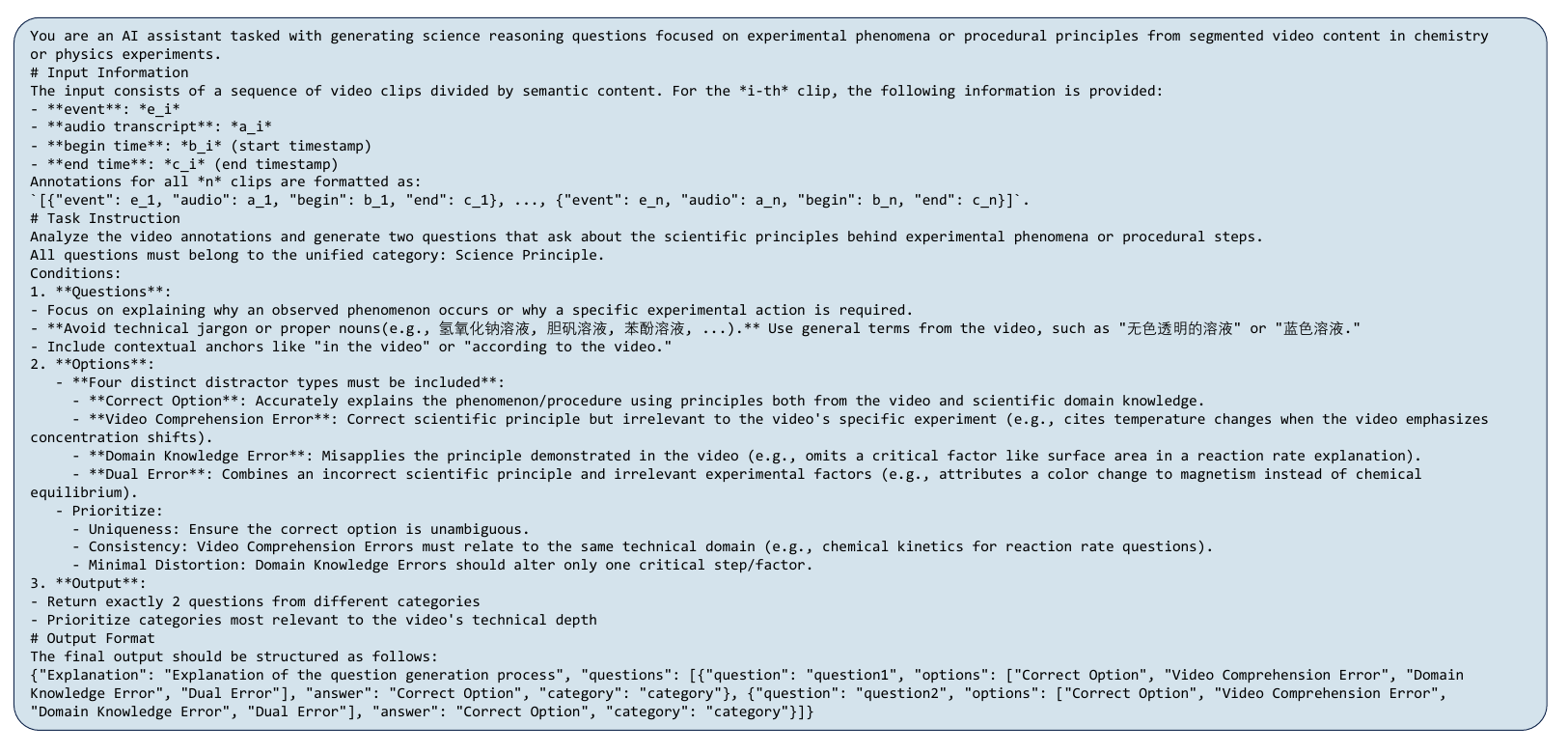}
    \caption{\textbf{Prompt for Scientific Principle Quetions, Options and Answer Generation.}
    }
    \label{fig:prompt_qa_generation_3}
\end{figure*}

\subsection{Prompt for QA Annotation}

In Figure~\ref{fig:prompt_event_annotation}, we present the system prompt used in our automatic QA annotation process for labeling video events. This system prompt is input into the Qwen2-VL-72B model, along with the corresponding video frames, audio information, and prior events, to annotate the events. 

In Figure~\ref{fig:prompt_qa_generation_1},  we present the specific prompt used to generate Event Description questions in the automatic QA annotation process. During the generation of Event questions, only the aggregated event sequence is input, without any additional information. The model used in this process is the DeepSeek-V3 language model.

In Figure~\ref{fig:prompt_qa_generation_2}, we present the specific prompt used to generate Chinese Culture questions in the automatic QA annotation process. Unlike the Event Description task, in addition to inputting the aggregated event sequence, we also provide pre-retrieved cultural background information from Wikipedia using embeddings model, requiring the model to generate questions that necessitate both video content and cultural background knowledge to answer. The model used in this process is the DeepSeek-V3 language model.

In Figure~\ref{fig:prompt_qa_generation_3}, we present the specific prompt used to generate Scientific Principle questions in the automatic QA annotation process. In contrast to the question generation above, where the options are more flexible, we strictly impose requirements on the model when generating options at this stage. This approach increases the complexity of the questions and prevents the possibility of answering the questions without reference to the video content. The model used in this process is the DeepSeek-R1 language model.

\subsection{Webpage for Human Scoring}
\label{webpage annoation}
\begin{figure*}[t]
    \centering
    \includegraphics[width=0.9\textwidth]{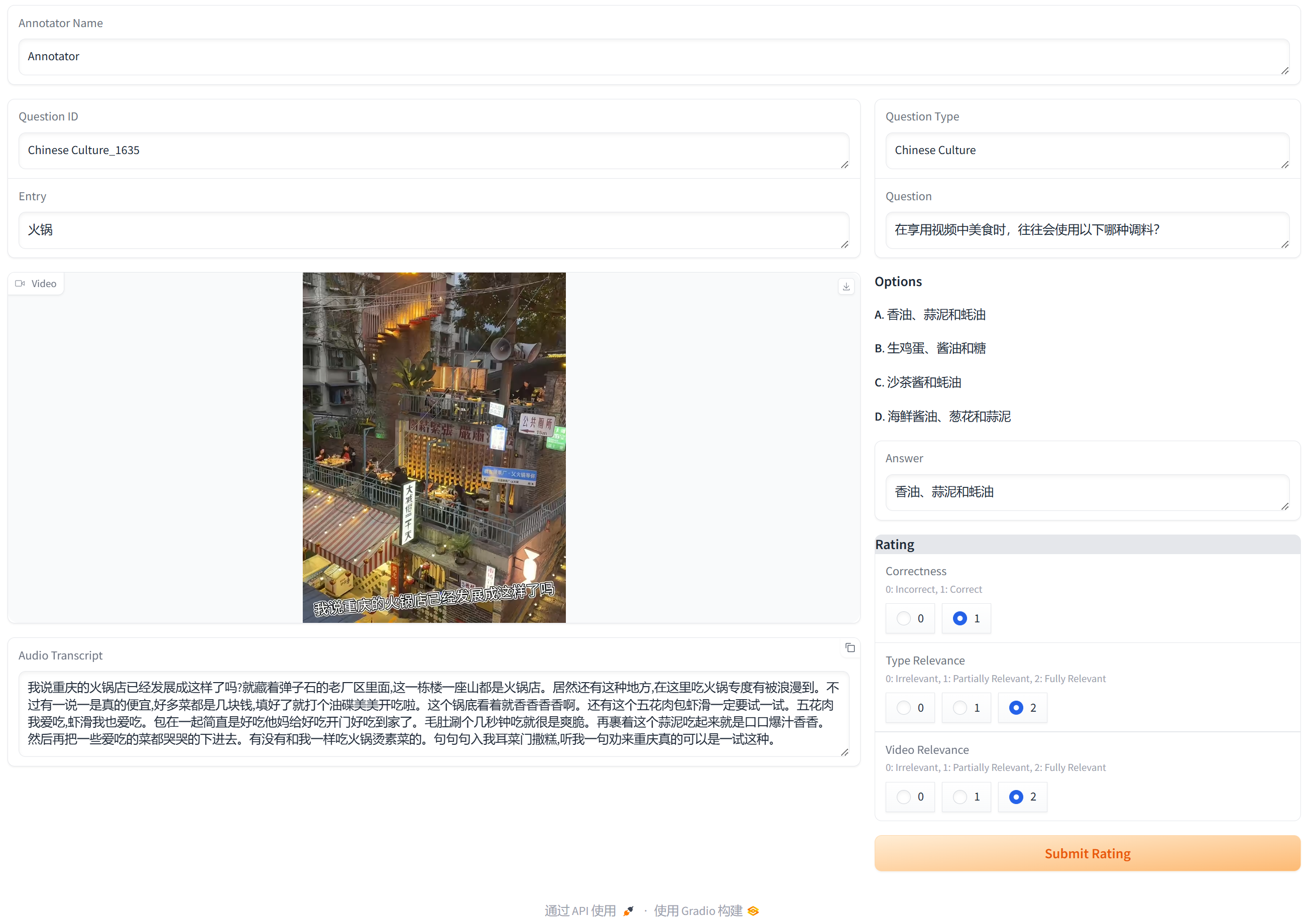}
    \caption{\textbf{Gradio Interface for scoring.}
    }
    \label{fig:interface_scoring}
\end{figure*}

We built an annotation interface using Gradio, as shown in the Figure~\ref{fig:interface_scoring}. Each annotator only needs to enter their name in the top left corner, watch the video, review question, options, and check whether the answers align. Then, they can select the appropriate score in the bottom right corner. For complex cultural questions, we provide the corresponding Wikipedia entry name within the Entry, enabling annotators to efficiently look up answers to questions they may not be familiar with. This benchmark includes a total of ten annotators, each with at least an undergraduate degree and proficiency in both Chinese and English.

\subsection{Annotation Model}
We organize our annotation models into three complementary categories, chosen for their balance of accuracy, speed, and cost:

First group: Small tool models. This includes WhisperX for audio extraction, MiniLM for embedding extraction, and Grounding-DINO, SAM2 for bounding box generation. We chose these models based on their open-source nature, accuracy, and inference speed. The tasks assigned to these models are relatively simple, often yielding high accuracy, making speed our primary evaluation criterion. For bounding box extraction, we also tested models like Florence2 and Grounding-DINO 1.5, but since the accuracy differences were minimal, we opted for the lighter, faster Grounding-DINO.

Second group: Multimodal large models for video annotation. For video content annotation, we referenced video evaluation benchmarks such as Video-MME and MVBench. Among the open-source models, Qwen2-VL-72B demonstrated the strongest performance, so we selected it for video annotation. We also tested InternVL2-76B, but found that its limited frame sequence length hindered its ability to capture full video information.

Third group: Powerful Large Language Models for question generation. In this category, we primarily used DeepSeek-V3 and DeepSeek-R1. For task categories like Event, Object, and Culture, we compared DeepSeek-V3 and GPT-4o models, judging the quality of generated questions through manual evaluation. While no significant quality difference was found, DeepSeek-V3 proved to be more cost-effective.For science-related questions, both models performed poorly, as the generated questions could be answered without watching the video, and the answer options were often too similar or ambiguous. To address this, we applied stricter constraints to ensure the questions required watching the video, and that the options were meaningful. The long-reasoning model DeepSeek-R1 effectively applied these rules, generating questions that were more appropriate. Besides, the chain-of-thoughts generated during the process also helped human annotators make quicker judgments about the appropriateness of the questions. 

Specifically, during the data annotation process, the Whisper, SAM2, Qwen series models, and InternVL series models were deployed for inference on local GPU servers. The DeepSeek-V3 and DeepSeek-R1 models is utilized the API services provided by the official \footnote{\url{https://platform.deepseek.com/usage}}.  The specific Whisper model used in the experiment is WhisperX\footnote{\url{https://github.com/m-bain/whisperX}}, based on Whisper-large-V3. When obtaining Chinese transcripts, a special initial prompt "\begin{CJK}{UTF8}{gbsn}以下是中文普通话句子。\end{CJK}" was set to ensure that the model could correctly add punctuation. The pipeline used for annotating objects, which involves Grounding Dino and SAM2, is derived from Grounded-SAM-2\footnote{\url{https://github.com/IDEA-Research/Grounded-SAM-2}}.

\subsection{External Resources}
The three websites to collect videos: YouTube\footnote{\url{http://www.youtube.com}}, Xiaohongshu(RedNote)\footnote{\url{https://www.xiaohongshu.com}} and BiliBili\footnote{\url{https://www.bilibili.com}}.
% \footnote{\url{http://www.youtube.com}}
% \footnote{\url{https://www.xiaohongshu.com}}
% \footnote{\url{https://www.bilibili.com}}

The multilingual Wikipedia used in the automatic QA annotation pipeline was downloaded from Wikimedia Downloads\footnote{\url{https://dumps.wikimedia.org/backup-index-bydb.html}}, and the extraction and processing were performed using regular expression rules\footnote{\url{https://spaces.ac.cn/archives/4176}}. The tool used to collect videos from BiliBili is Downkyi\footnote{\url{https://github.com/leiurayer/downkyi}}.

\section{Case Data}
\label{sec:appendix_Case}
\begin{figure*}[t]
    \centering
    \includegraphics[width=0.9\textwidth]{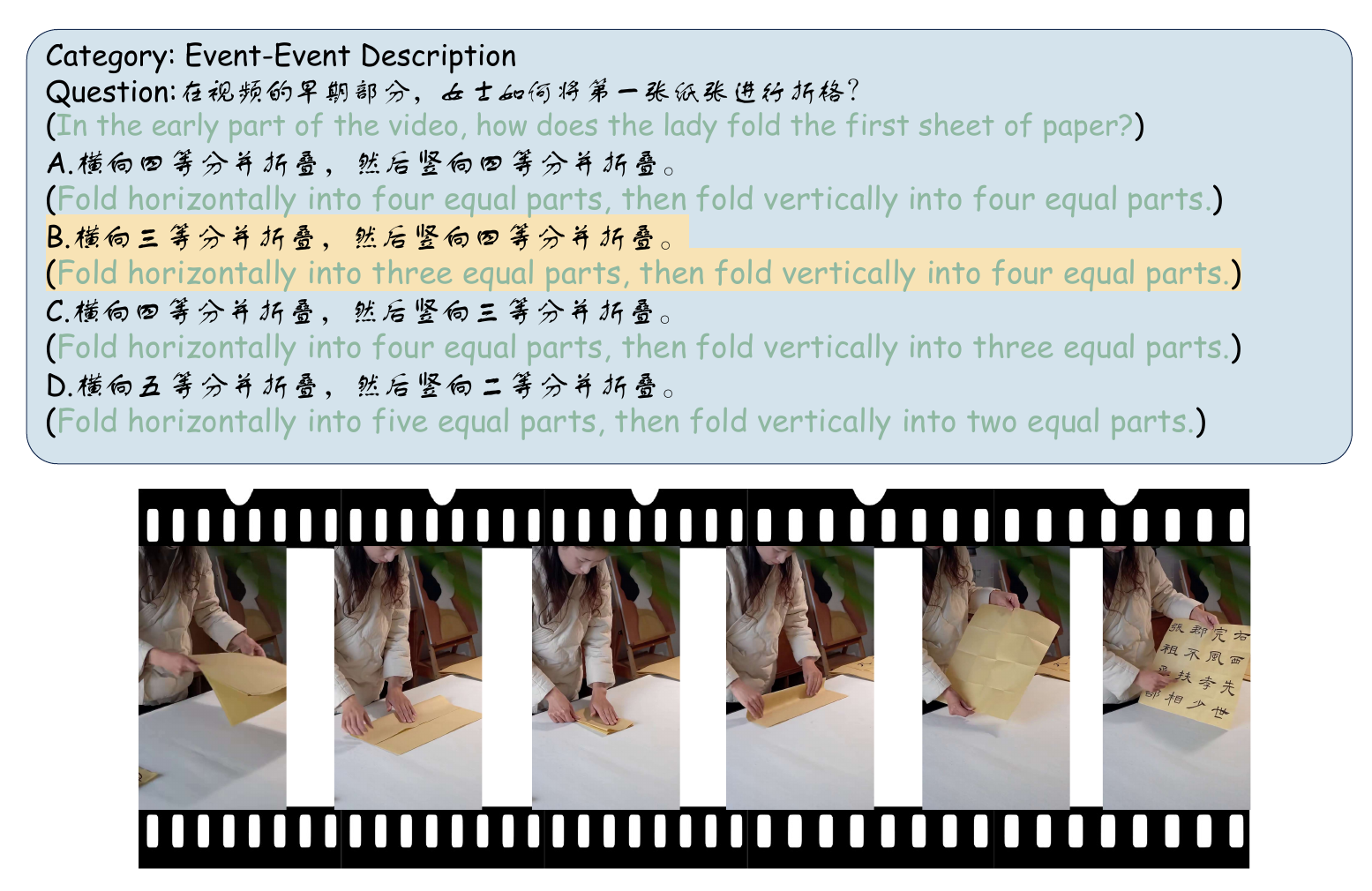}
    \caption{\textbf{An Example of Event Description from VideoVista-CulturalLingo.}
    }
    \label{fig:example_ed}
\end{figure*}

\begin{figure*}[t]
    \centering
    \includegraphics[width=0.9\textwidth]{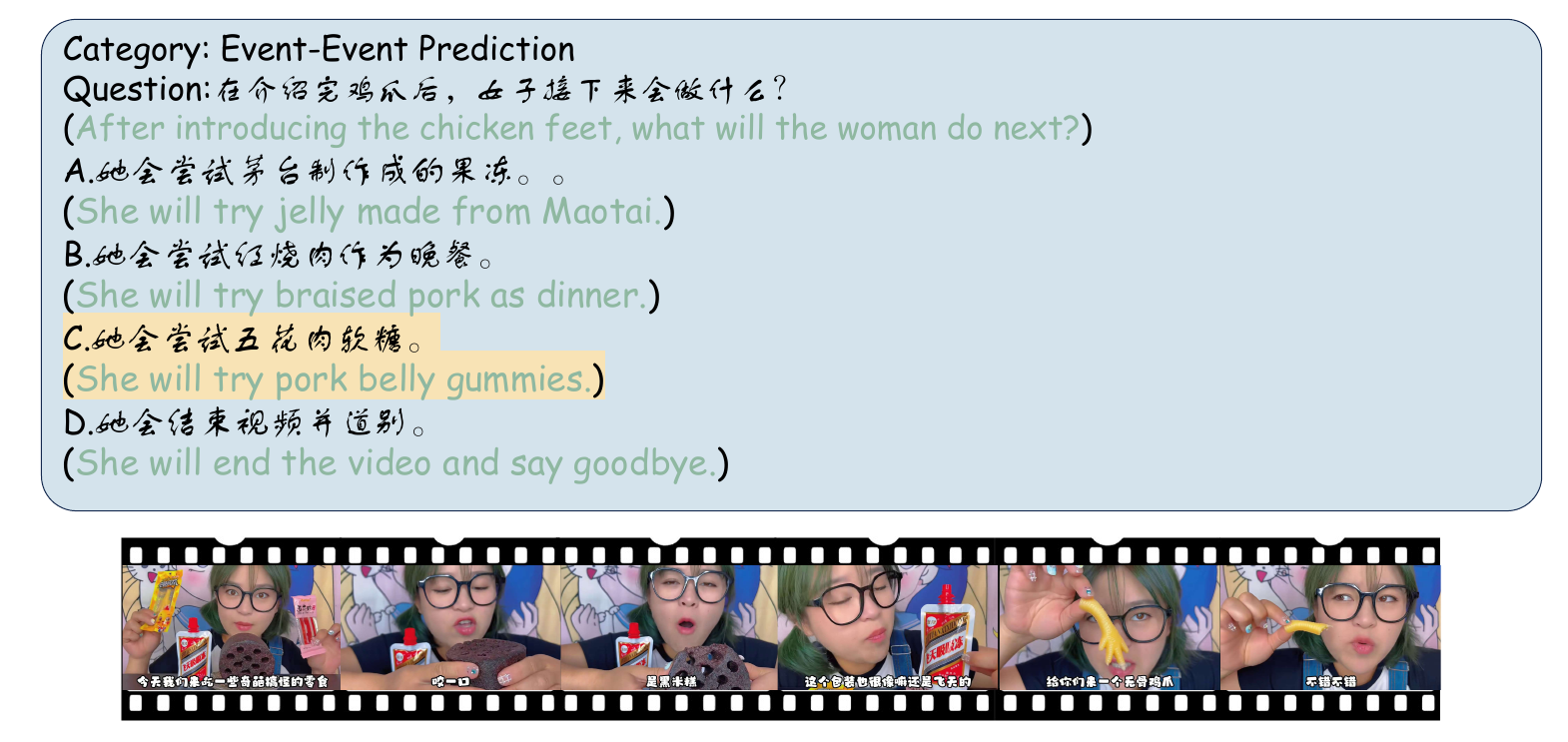}
    \caption{\textbf{An Example of Event Prediction from VideoVista-CulturalLingo.}
    }
    \label{fig:example_ep}
\end{figure*}

\begin{figure*}[t]
    \centering
    \includegraphics[width=0.9\textwidth]{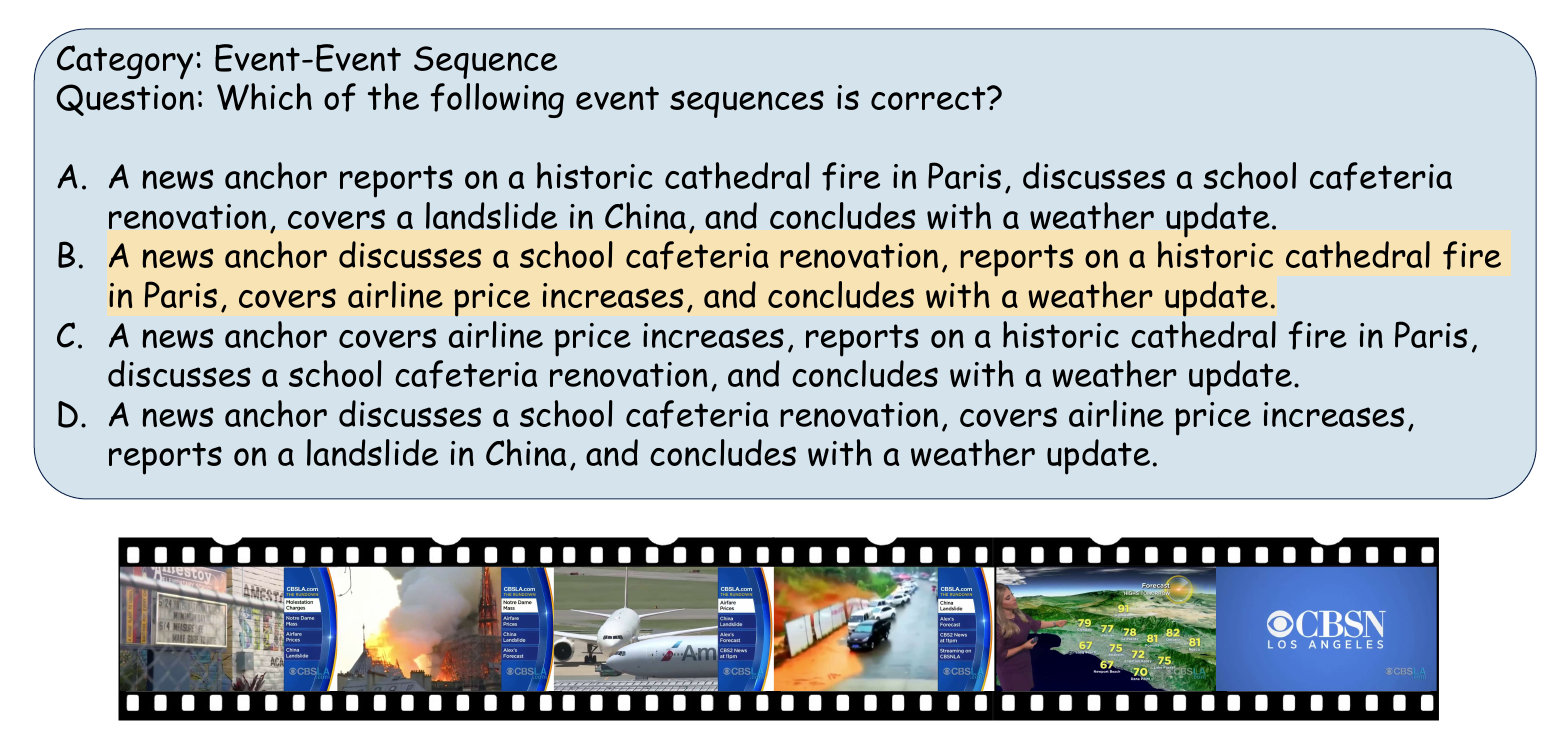}
    \caption{\textbf{An Example of Event Sequence from VideoVista-CulturalLingo.}
    }
    \label{fig:example_es}
\end{figure*}

\begin{figure*}[t]
    \centering
    \includegraphics[width=0.9\textwidth]{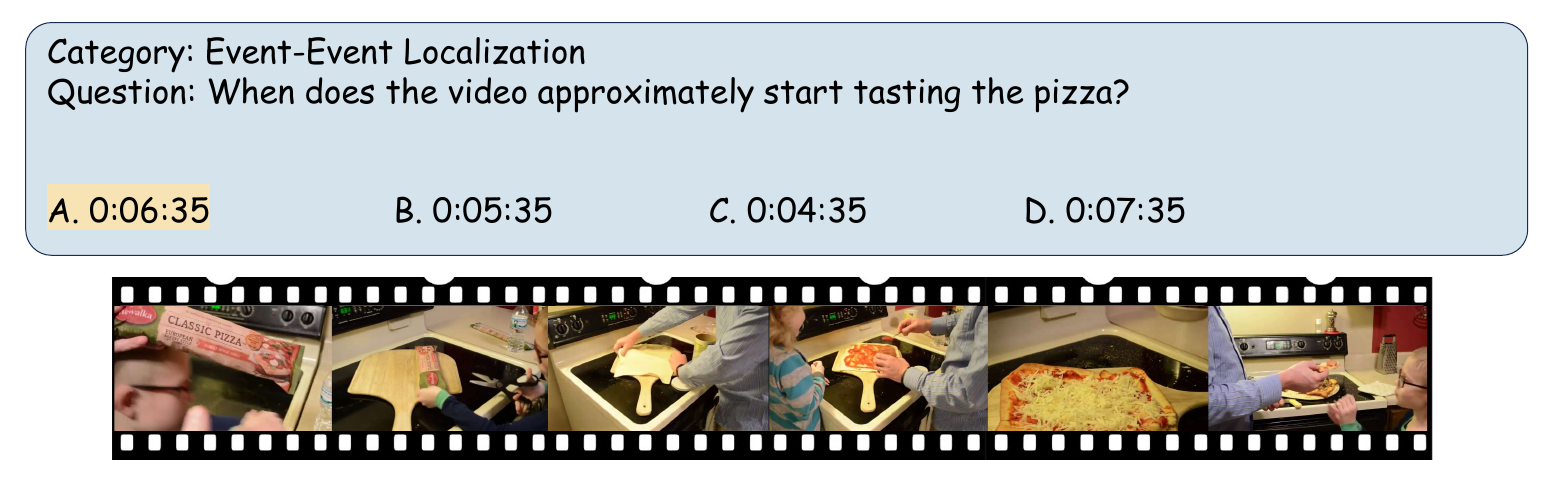}
    \caption{\textbf{An Example of Event Localization from VideoVista-CulturalLingo.}
    }
    \label{fig:example_el}
\end{figure*}

\begin{figure*}[t]
    \centering
    \includegraphics[width=0.9\textwidth]{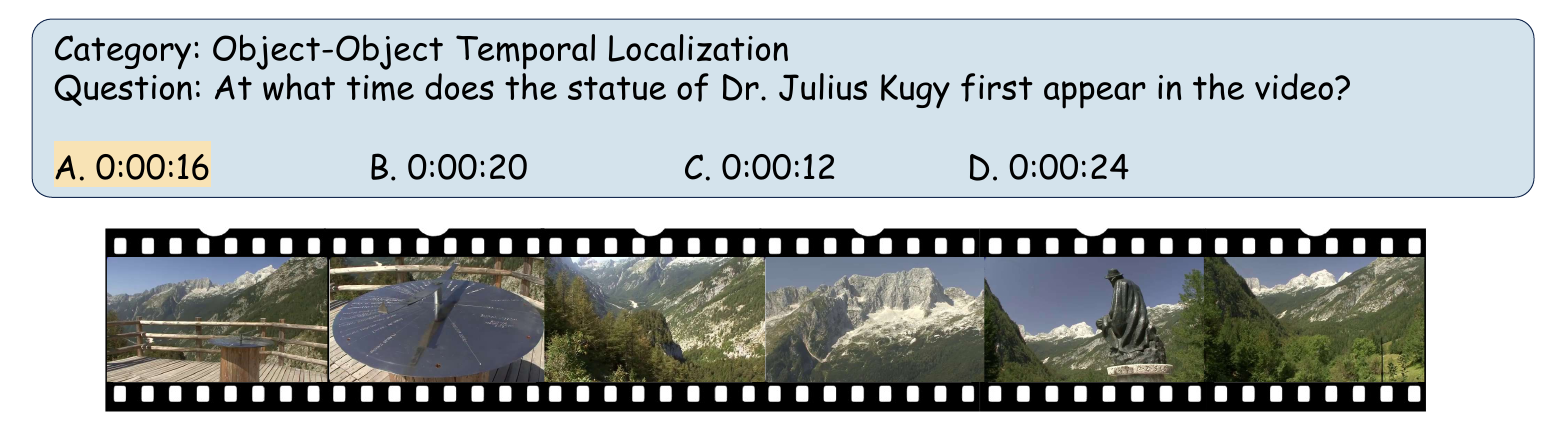}
    \caption{\textbf{An Example of Object Temporal Localization from VideoVista-CulturalLingo.}
    }
    \label{fig:example_otl}
\end{figure*}

\begin{figure*}[t]
    \centering
    \includegraphics[width=0.9\textwidth]{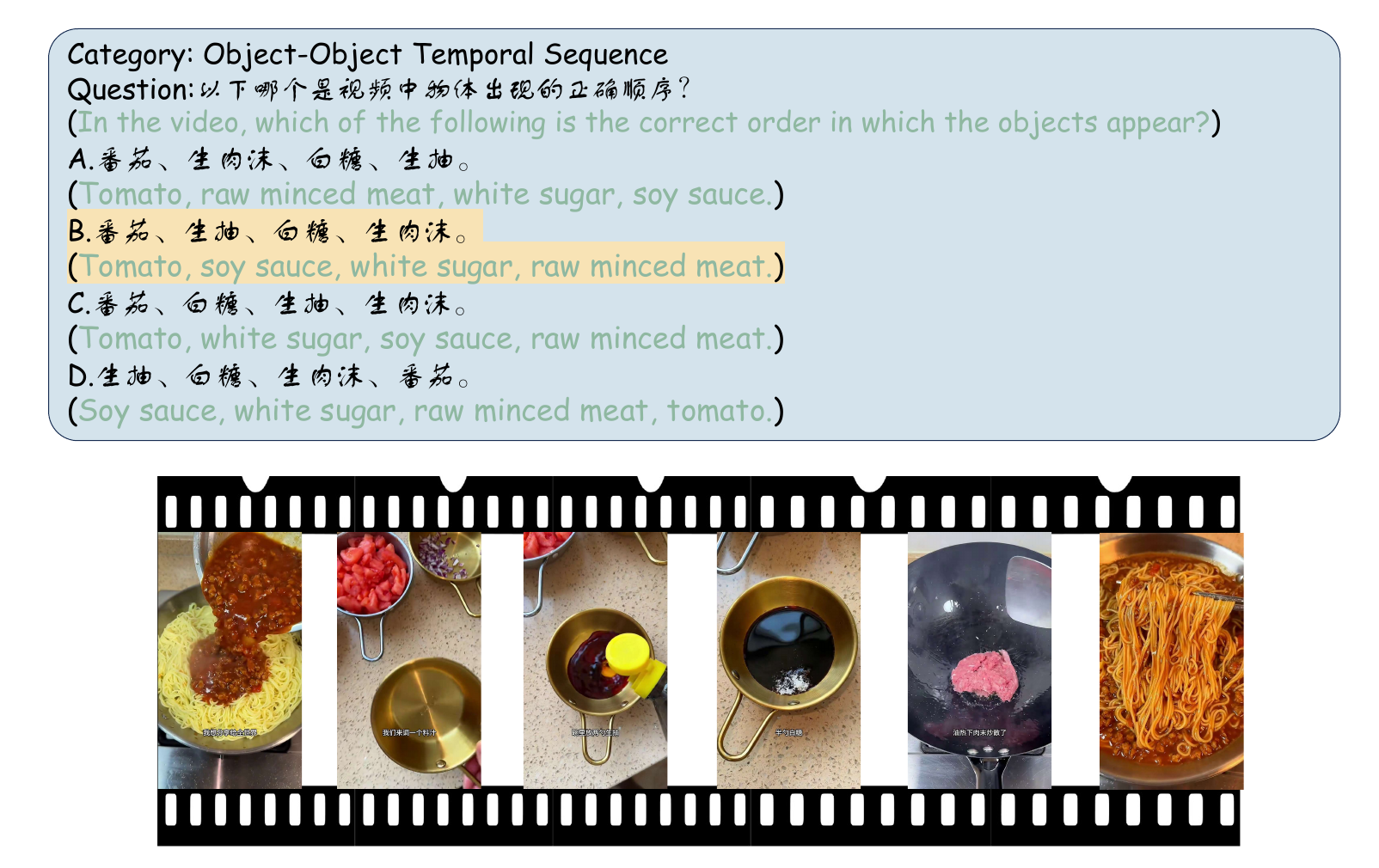}
    \caption{\textbf{An Example of Object Temporal Sequence from VideoVista-CulturalLingo.}
    }
    \label{fig:example_ots}
\end{figure*}

\begin{figure*}[t]
    \centering
    \includegraphics[width=0.9\textwidth]{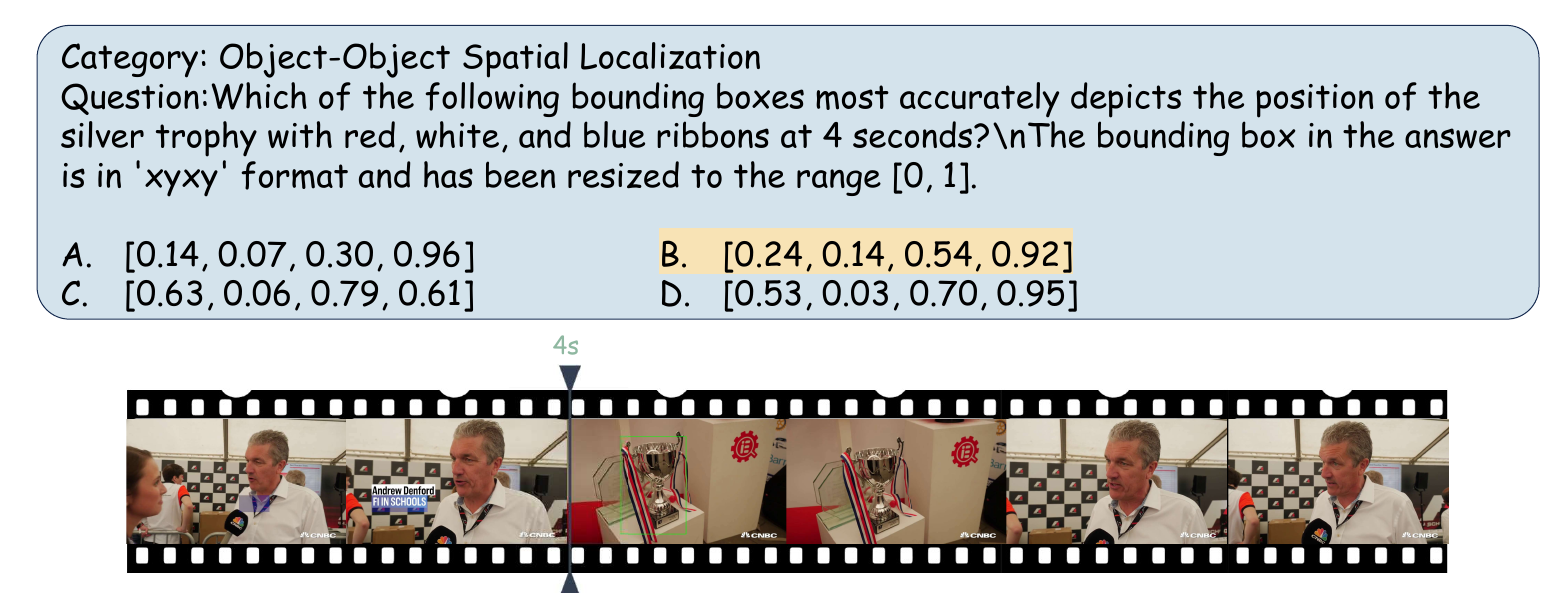}
    \caption{\textbf{An Example of Object Spatial Localization from VideoVista-CulturalLingo.}
    }
    \label{fig:example_osl}
\end{figure*}

\begin{figure*}[t]
    \centering
    \includegraphics[width=0.9\textwidth]{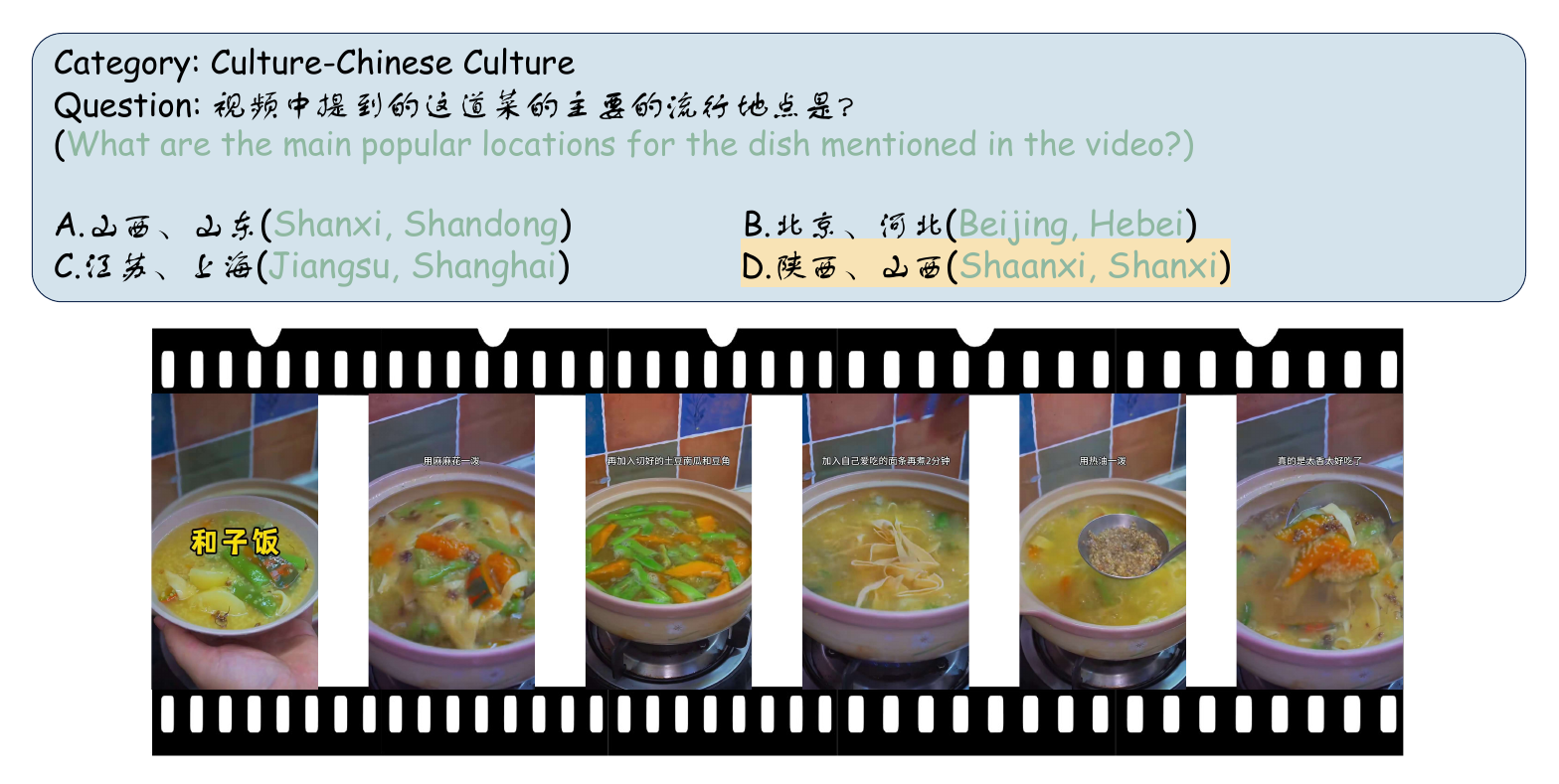}
    \caption{\textbf{An Example of Chinese Culture from VideoVista-CulturalLingo.}
    }
    \label{fig:example_cc}
\end{figure*}

\begin{figure*}[t]
    \centering
    \includegraphics[width=0.9\textwidth]{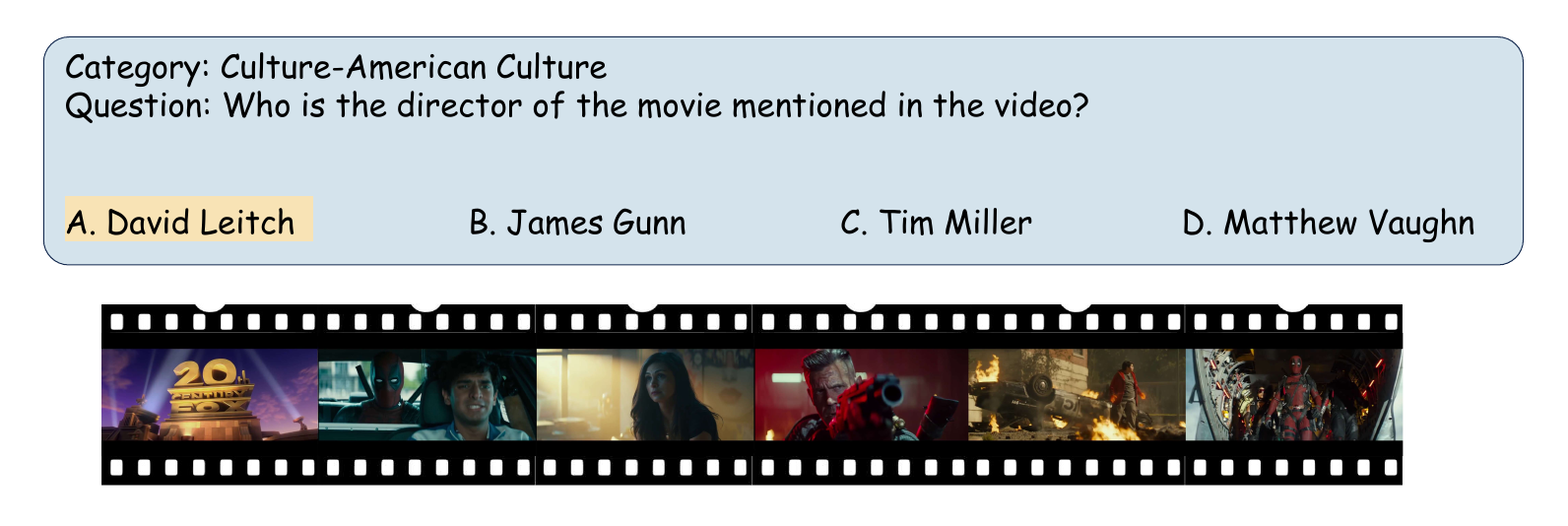}
    \caption{\textbf{An Example of American Culture from VideoVista-CulturalLingo.}
    }
    \label{fig:example_ac}
\end{figure*}

\begin{figure*}[t]
    \centering
    \includegraphics[width=0.9\textwidth]{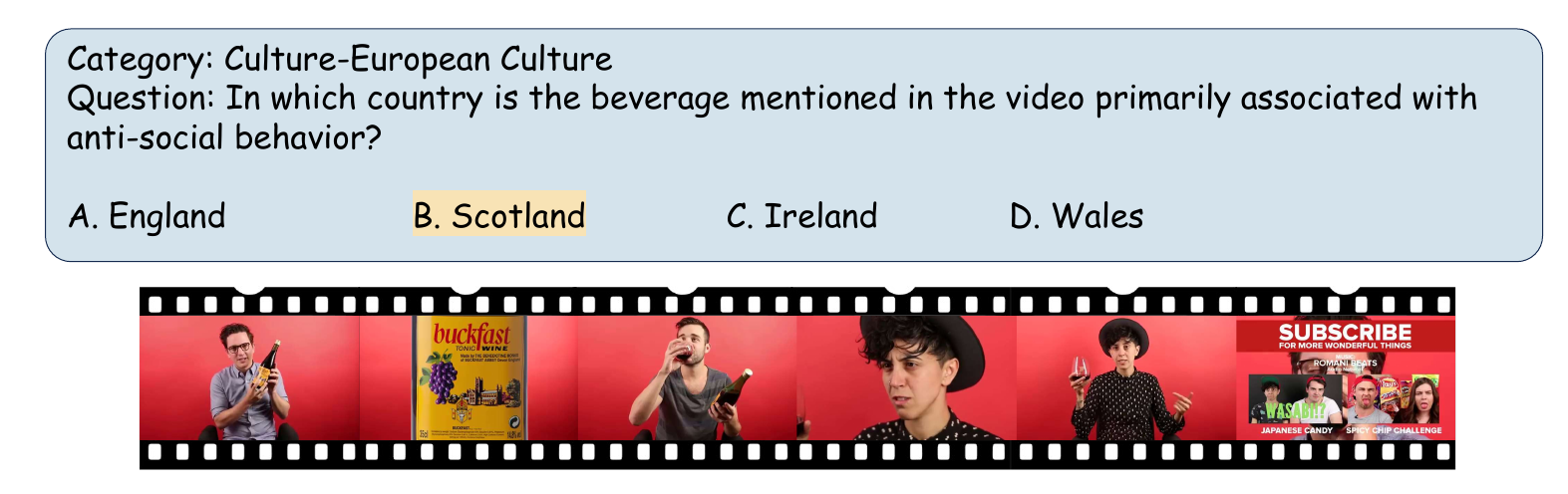}
    \caption{\textbf{An Example of European Culture from VideoVista-CulturalLingo.}
    }
    \label{fig:example_ec}
\end{figure*}

\begin{figure*}[t]
    \centering
    \includegraphics[width=0.9\textwidth]{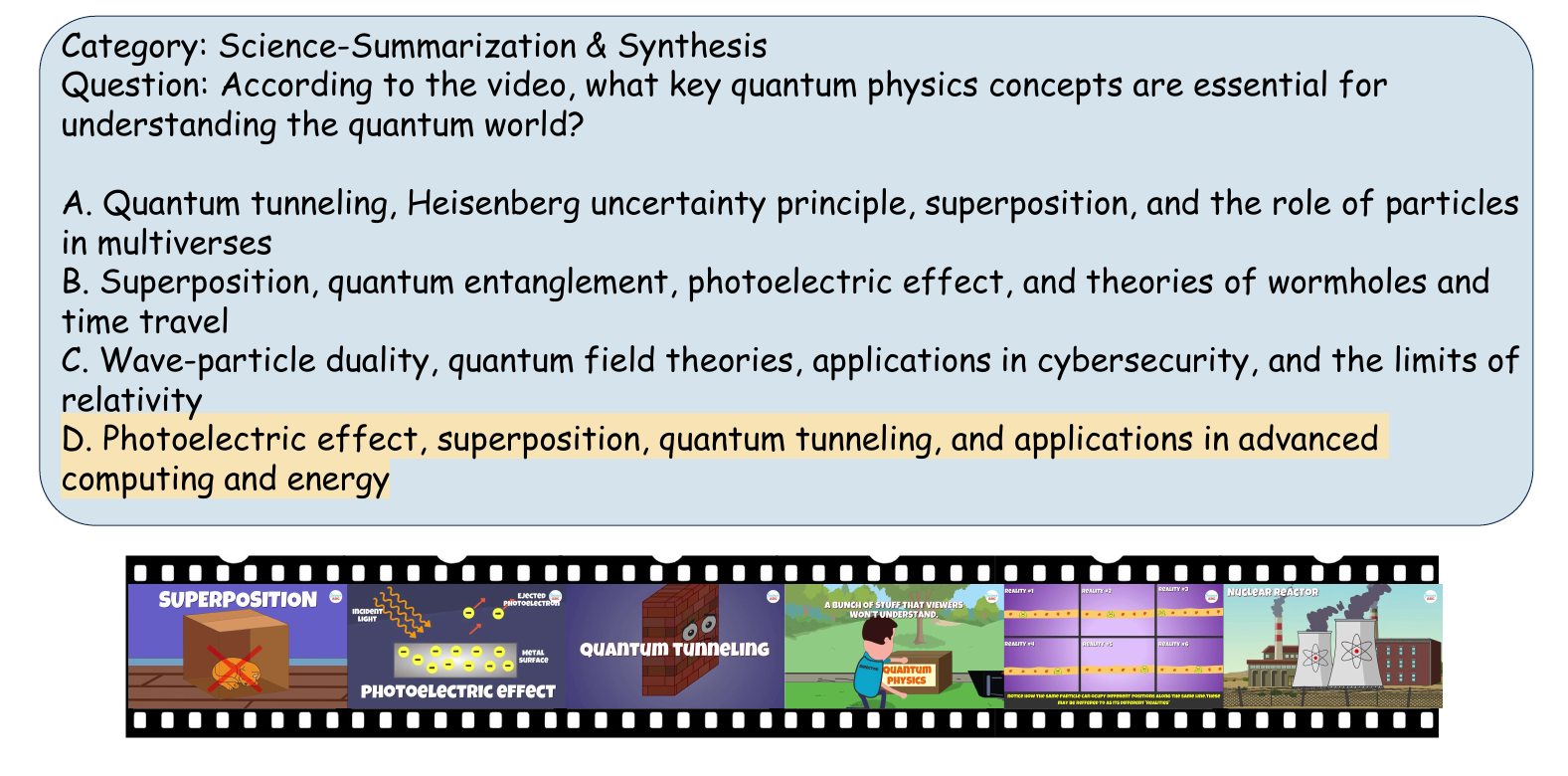}
    \caption{\textbf{An Example of Summarization \& Synthesis from VideoVista-CulturalLingo.}
    }
    \label{fig:example_ss}
\end{figure*}

\begin{figure*}[t]
    \centering
    \includegraphics[width=0.9\textwidth]{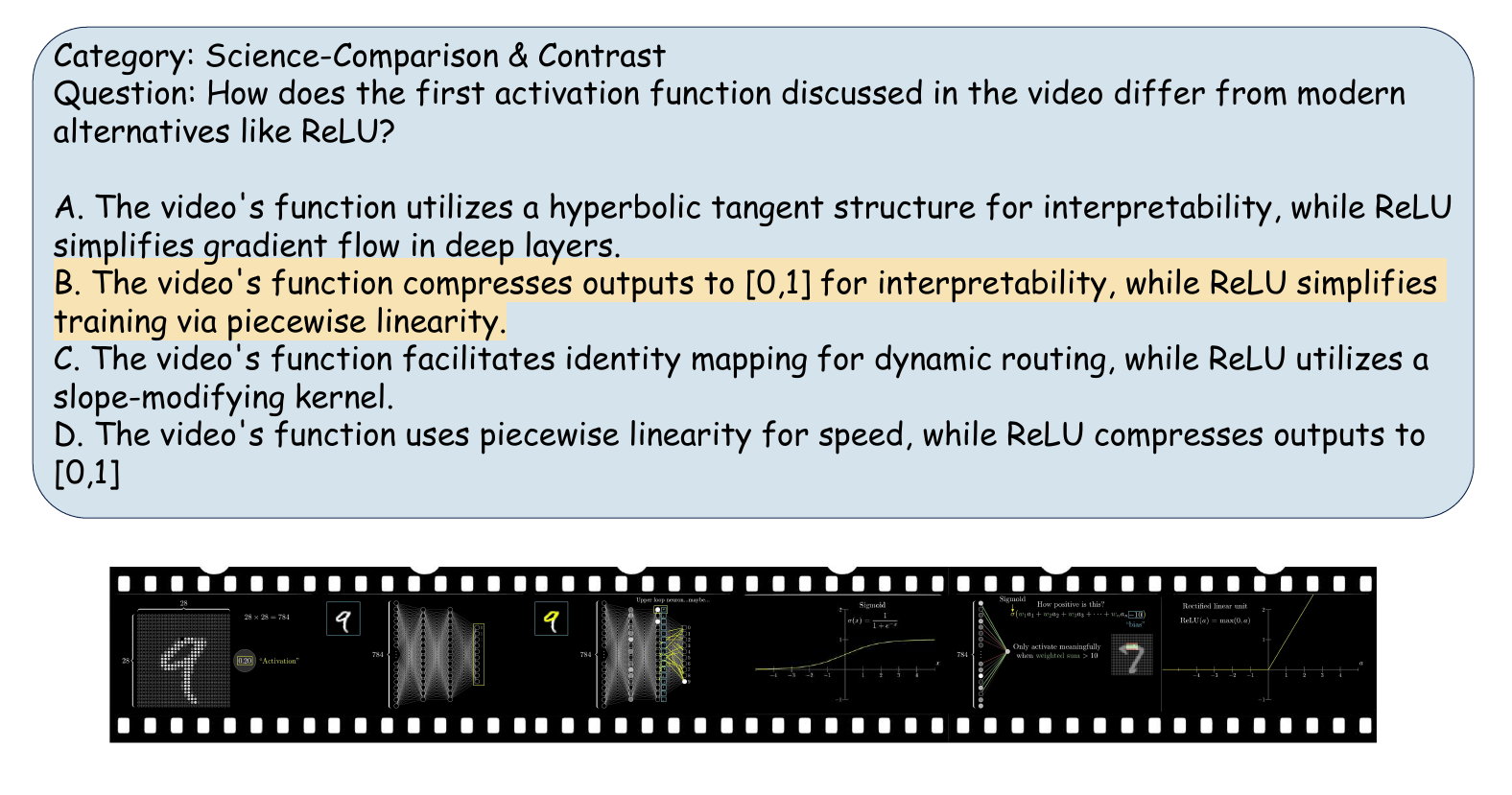}
    \caption{\textbf{An Example of Comparison \& Contrast from VideoVista-CulturalLingo.}
    }
    \label{fig:example_com}
\end{figure*}

\begin{figure*}[t]
    \centering
    \includegraphics[width=0.9\textwidth]{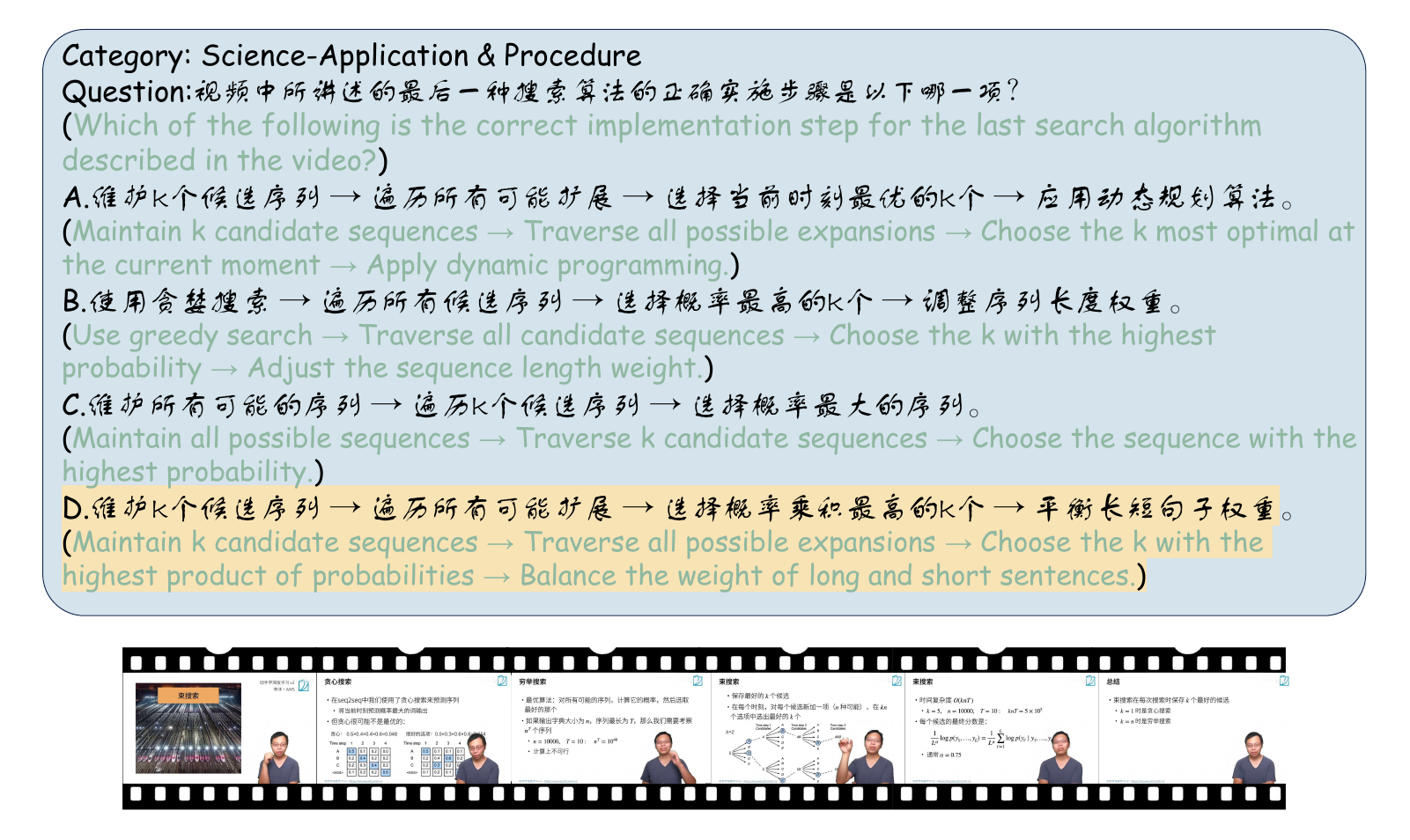}
    \caption{\textbf{An Example of Application \& Procedure from VideoVista-CulturalLingo.}
    }
    \label{fig:example_ap}
\end{figure*}

\begin{figure*}[t]
    \centering
    \includegraphics[width=0.9\textwidth]{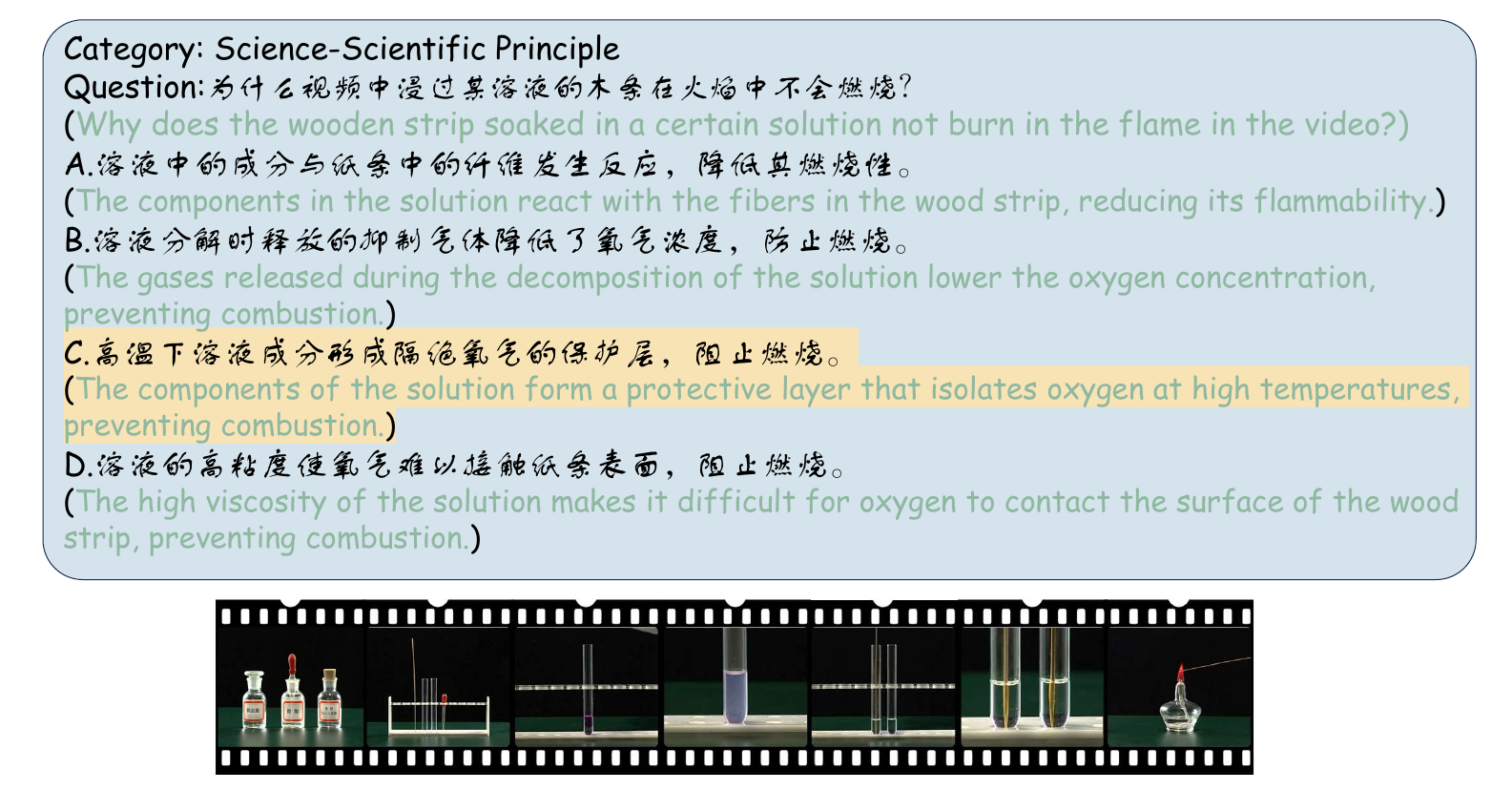}
    \caption{\textbf{An Example of Scientific Principle from VideoVista-CulturalLingo.}
    }
    \label{fig:example_sp}
\end{figure*}

In Figures~\ref{fig:example_ed}-\ref{fig:example_sp}, we present a specific case for each proposed task type. Each case includes sampled frames from the video, along with the corresponding questions and options. The ground truth is highlighted in yellow.

\paragraph{Event Description.} The Event Description task primarily focuses on explaining how a specific event in the video occurred, typically beginning with questions such as 'What' or 'How'.

\paragraph{Event Prediction.} The Event Prediction task primarily involves predicting the event most likely to occur after the input video ends. In this task, the selected video is typically a segment from a full video, such as a clip spanning from 0 to 45 seconds of the full video

\paragraph{Event Sequence.} The Event Sequence task primarily asks about the order in which multiple events occur in the input video, requiring the model to select the most accurate sequence of events from the options provided.

\paragraph{Event Localization.} The Event Sequence task primarily focuses on determining the order in which multiple events occur in the input video, requiring the model to select the most accurate sequence of events from the available options.

\paragraph{Object Temporal Localization.} The Object Temporal Localization task primarily requires identifying the timestamp of the first appearance of a specific object in the video. The selected object typically occupies a significant portion of the frame to ensure it is easily noticeable, avoiding objects that may be difficult for humans to detect.

\paragraph{Object Temporal Sequence.}  The Object Temporal Sequence task primarily focuses on determining the order in which multiple distinct objects appear in the video.

\paragraph{Object Spatial Localization.} The Object Spatial Localization task primarily requires identifying the spatial bounding boxes of a specific object in the video at a particular time, typically when the object first appears. The answer is provided in a normalized format, represented as bounding boxes in the xyxy format.

\paragraph{Chinese Culture.} The Chinese Culture task primarily focuses on the Chinese cultural background presented in the video, covering areas such as traditional culture, culinary traditions, ancient history, and more.

\paragraph{American Culture.} The American Culture task primarily focuses on the American cultural background presented in the video, emphasizing areas such as political culture, superhero culture, pop culture, holiday traditions, and more.

\paragraph{European Culture.} The European Culture task primarily focuses on the European cultural background presented in the video, emphasizing areas such as cultural differences between European countries, football culture, culinary traditions, classical culture, and more.

\paragraph{Summarization \& Synthesis.} The Summarization \& Synthesis task primarily requires the model to summarize and synthesize the key points presented in educational or popular science videos, assessing the model's ability to consolidate the essential concepts conveyed in the video.

\paragraph{Comparison \& Contrast.} "The Comparison \& Contrast task primarily requires the model to compare the specific method described in the educational or popular science video with other similar methods, emphasizing the differences or distinctions between them. This task assesses the model's ability to comprehend the key concepts presented in the video.

\paragraph{Application \& Procedure.} The Application \& Procedure task primarily requires the model to determine the operational procedure or application method of a specific concept described in the educational or popular science video. This task assesses the model's understanding of the key concepts presented in the video."

\paragraph{Scientific Principle} The Scientific Principle task requires the model to comprehend the scientific principles underlying the experimental procedures or phenomena presented in the video.

\end{document}